\documentclass[10pt,twocolumn,letterpaper]{article}

\usepackage[pagenumbers]{cvpr} %

\usepackage{graphicx}
\usepackage{amsmath}
\usepackage{amssymb}
\usepackage{booktabs}
\usepackage{xcolor}
\usepackage{amssymb}%
\usepackage{pifont}%
\usepackage{tocloft}
\newcommand{\cmark}{\ding{51}}%
\newcommand{\xmark}{\ding{55}}%

\usepackage[pagebackref,breaklinks,colorlinks]{hyperref}

\usepackage[capitalize]{cleveref}
\crefname{section}{Sec.}{Secs.}
\Crefname{section}{Sec.}{Sections}
\Crefname{table}{Table}{Tables}
\crefname{table}{Tab.}{Tabs.}
\Crefname{figure}{Fig.}{Figs.}
\Crefname{equation}{Eq.}{Eqs.}

\usepackage{amsmath,amsfonts,bm,mathtools}

\def\eqref#1{equation~\ref{#1}}

\def\1{\bm{1}}

\def\rvepsilon{{\boldsymbol{\epsilon}}}

\def\rvc{{\mathbf{c}}}

\def\rvf{{\mathbf{f}}}
\def\rvg{{\mathbf{g}}}

\def\rvm{{\mathbf{m}}}

\def\rvv{{\mathbf{v}}}

\def\rvx{{\mathbf{x}}}
\def\rvy{{\mathbf{y}}}
\def\rvz{{\mathbf{z}}}

\def\mI{{\bm{I}}}

\DeclareMathAlphabet{\mathsfit}{\encodingdefault}{\sfdefault}{m}{sl}
\SetMathAlphabet{\mathsfit}{bold}{\encodingdefault}{\sfdefault}{bx}{n}

\def\gD{{\mathcal{D}}}
\def\gE{{\mathcal{E}}}

\def\gN{{\mathcal{N}}}

\newcommand{\E}{\mathbb{E}}

\newcommand{\R}{\mathbb{R}}

\DeclareMathOperator*{\argmin}{arg\,min}

\def\cvprPaperID{8978} %
\def\confName{CVPR}
\def\confYear{2023}

\begin{document}

\title{\vspace{-2.7em} Align your Latents: \\ High-Resolution Video Synthesis with Latent Diffusion Models}

\author{
Andreas Blattmann\textsuperscript{1\;\,*,\dag}
\quad\quad
Robin Rombach\textsuperscript{1\;\,*,\dag}
\quad\quad
Huan Ling\textsuperscript{2,3,4\;\,*}
\quad\quad
Tim Dockhorn\textsuperscript{2,3,5\;\,*,\dag}
\vspace{0.15cm}
\\
Seung Wook Kim\textsuperscript{2,3,4}
\quad\quad
Sanja Fidler\textsuperscript{2,3,4}
\quad\quad
Karsten Kreis\textsuperscript{2}
\vspace{0.25cm}
\\
{\small\textsuperscript{1}LMU Munich \quad \textsuperscript{2}NVIDIA \quad \textsuperscript{3}Vector Institute \quad \textsuperscript{4}University of Toronto \quad \textsuperscript{5}University of Waterloo \vspace{0pt}}
\vspace{0.15cm}
\\
\small \textit{Project page:} \url{https://research.nvidia.com/labs/toronto-ai/VideoLDM/}
}

\twocolumn[{
    \renewcommand\twocolumn[1][]{#1}
    \maketitle
    \begin{center}
            \vspace{-15pt}
        \centering
        \includegraphics[width=\linewidth]{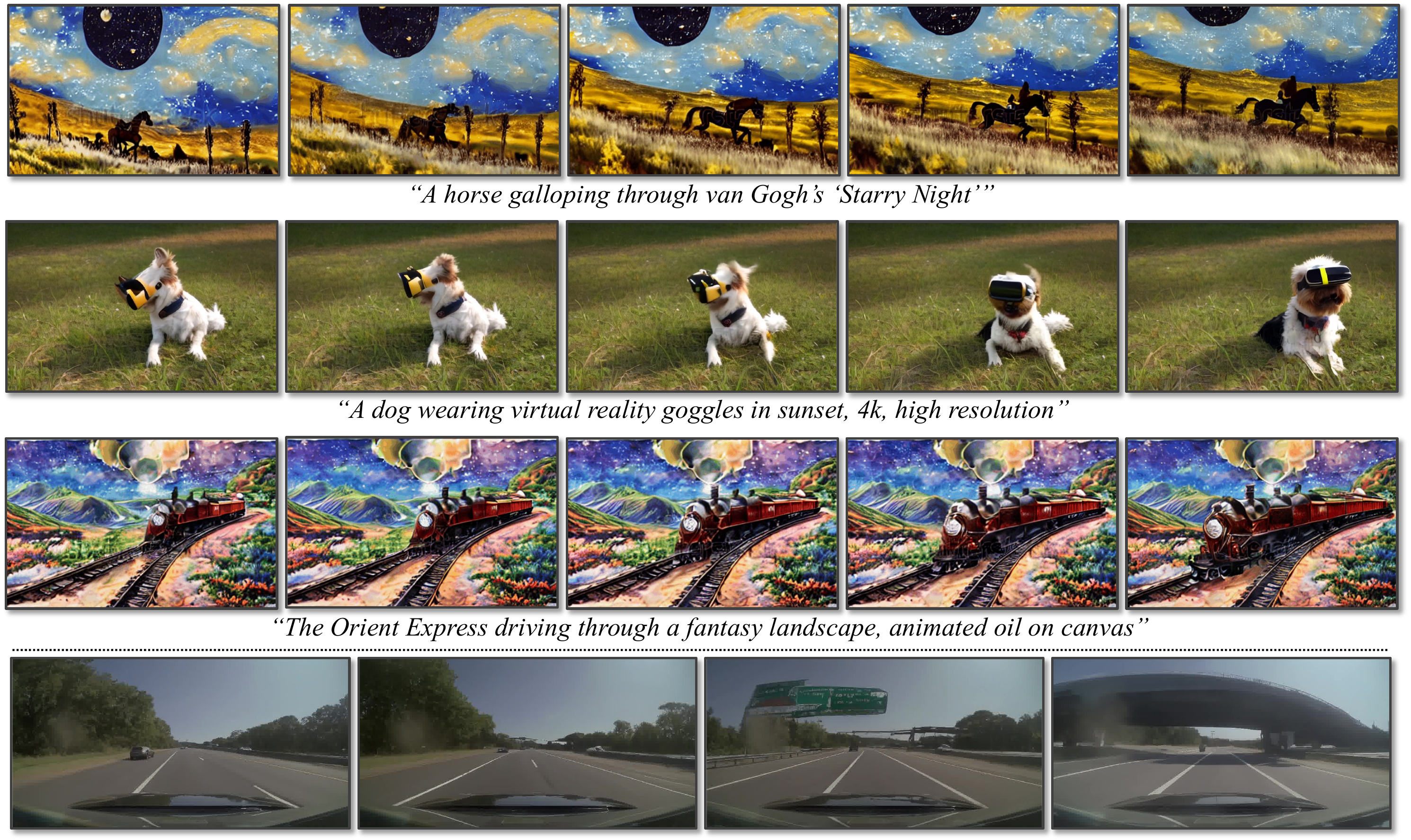}
    \vspace{-0.7cm}
    \captionof{figure}{
        \small \textbf{Video LDM samples}. \emph{Top:} Text-to-Video generation. \emph{Bottom:} $512\times 1024$ resolution real driving scene video generation.
    }
    \label{fig:teaser}

    \end{center}
}]

\begin{figure*}[t!]
  \begin{minipage}[c]{0.69\textwidth}
  \vspace{-7mm}
    \includegraphics[width=1.0\textwidth]{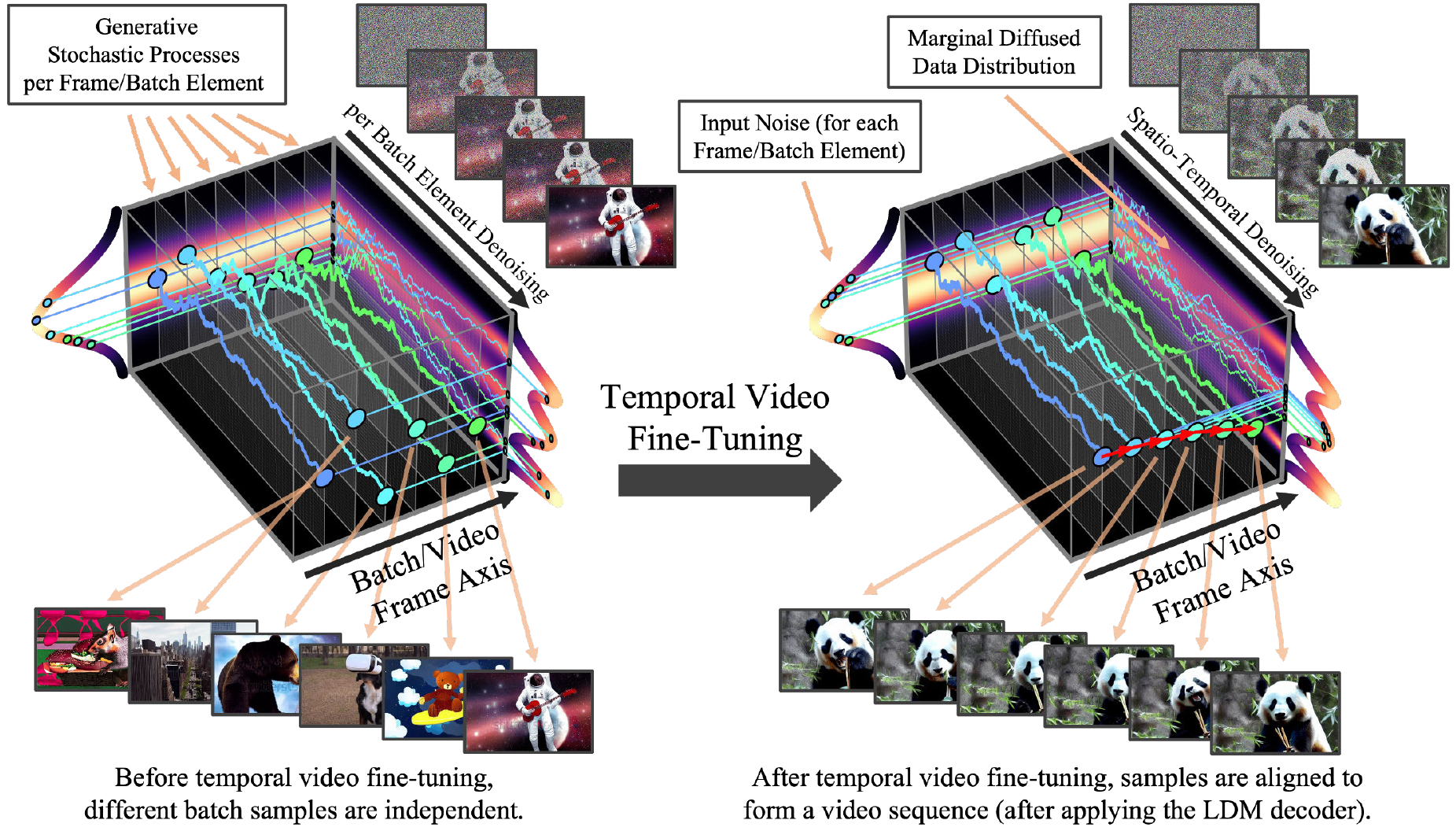}
  \end{minipage}\hfill
  \begin{minipage}[c]{0.31\textwidth}
  \vspace{-1mm}
    \caption{\small \textbf{Temporal Video Fine-Tuning.} We turn pre-trained image diffusion models into temporally consistent video generators. Initially, different samples of a batch synthesized by the model are independent. After temporal video fine-tuning, the samples are temporally aligned and form coherent videos. The stochastic generation process before and after fine-tuning is visualised for a diffusion model of a one-dim. toy distribution. For clarity, the figure corresponds to alignment in pixel space. In practice, we perform alignment in LDM's latent space and obtain videos after applying LDM's decoder (see \Cref{fig:ldm_figure}). We also video fine-tune diffusion model upsamplers in pixel or latent space (\Cref{sec:method_upsampler}).}
    \label{fig:pipeline}
  \end{minipage}
  \vspace{-7mm}
\end{figure*}
\begin{abstract}
\vspace{-0.2cm}
Latent Diffusion Models (LDMs) enable high-quality image synthesis while avoiding excessive compute demands by training a diffusion model in a compressed lower-dimensional latent space. Here, we apply the LDM paradigm to high-resolution video generation, a particularly resource-intensive task. 
We first pre-train an LDM on images only; then, we turn the image generator into a video generator by introducing a temporal dimension to the latent space diffusion model and fine-tuning on encoded image sequences, i.e., videos. Similarly, we temporally align diffusion model upsamplers, turning them into temporally consistent video super resolution models. 
We focus on two relevant real-world applications: Simulation of in-the-wild driving data and creative content creation with text-to-video modeling. In particular, we validate our \textbf{Video LDM} on real driving videos 
of resolution $512\times1024$, achieving state-of-the-art performance. %
Furthermore, our approach can easily leverage off-the-shelf pre-trained image LDMs, as we only need to train a temporal alignment model in that case.  Doing so, we turn the publicly available, state-of-the-art text-to-image LDM Stable Diffusion into an efficient and expressive text-to-video model with resolution up to $1280\times2048$.
We show that the temporal layers trained in this way generalize to different fine-tuned text-to-image LDMs. Utilizing this property, we show the first results for personalized text-to-video generation, opening exciting directions for future content creation.
{\let\thefootnote\relax\footnote{{\textsuperscript{*}Equal contribution.}}}
{\let\thefootnote\relax\footnote{{\textsuperscript{\dag}Andreas, Robin and Tim did the work during internships at NVIDIA.}}}
\end{abstract}

\vspace{-3mm}
\section{Introduction}\label{sec:intro}
\vspace{-1mm}

Generative models of images have received unprecedented attention, owing to recent breakthroughs in the underlying modeling methodology. The most powerful models today are built on generative adversarial networks~\cite{goodfellow2014generative,karras2019style,karras2020analyzing,karras2021aliasfree,sauer2021styleganxl}, autoregressive transformers~\cite{esser2020taming,ramesh2021dalle,yu2022parti}, and most recently diffusion models~\cite{sohl2015deep,ho2020ddpm,song2020score,nichol2021improved,dhariwal2021diffusion,ho2021cascaded,nichol2021glide,rombach2021highresolution,ramesh2022dalle2,saharia2022imagen}.
Diffusion models (DMs) in particular have desirable advantages; they offer a robust and scalable training objective and are typically less parameter intensive than their transformer-based counterparts. 
However, while the image domain has seen great progress, \emph{video} modeling has lagged behind---mainly due to the significant computational cost associated with training on video data, and the lack of large-scale, general, and publicly available video datasets. While there is a rich literature on video synthesis~\cite{babaeizadeh2018stochastic,svg,lee2018savp,hvrnn,lsvg,Weissenborn2020Scaling,yan2021videogpt,hong2022cogvideo,wu2021godiva,wu2022nuwa,ge2022longvideo,Gupta_2022_CVPR,scene_dyn,yu2022generating,tian2021a, villegas17mcnet,Luc2020TransformationbasedAV,TGAN2020,brooks2022generating,Skorokhodov_2022_CVPR,kahembwe2020lower,hong2022cogvideo,mittal2017sync,Pan2017ToCW,marwah2017attentive,li2017video,gupta2018imagine}, most works, including previous video DMs~\cite{yang2022video,ho2022video,hoeppe2022diffusion,voleti2022mcvd,harvey2022flexible}, only generate relatively low-resolution, often short, videos.
Here, we apply video models to real-world problems and generate high-resolution, long videos. 
Specifically, we focus on two relevant real-world video generation problems: (i) video synthesis of high-resolution real-word driving data, which has great potential as a simulation engine in the context of autonomous driving, and (ii) text-guided video synthesis for creative content generation; see \cref{fig:teaser}.

To this end, we build on latent diffusion models (LDMs), which can reduce the heavy computational burden when training on high-resolution images~\cite{rombach2021highresolution}. 
We propose \emph{Video LDMs} and extend LDMs to high-resolution \emph{video} generation, a particularly compute-intensive task. In contrast to previous work on DMs for video generation~\cite{yang2022video,ho2022video,hoeppe2022diffusion,voleti2022mcvd,harvey2022flexible}, we first pre-train our Video LDMs on images only (or use available pre-trained image LDMs), thereby allowing us to leverage large-scale image datasets. 
We then transform the LDM image generator into a video generator by introducing a temporal dimension into the latent space DM and training only these temporal layers on encoded image sequences, \ie, videos (\Cref{fig:pipeline}), while fixing the pre-trained spatial layers.
We similarly fine-tune LDM's decoder to achieve temporal consistency in pixel space (\Cref{fig:ldm_figure}).
To further enhance the spatial resolution, we also temporally align pixel-space and latent DM upsamplers~\cite{ho2021cascaded}, which are widely used for image super resolution~\cite{saharia2021image,li2022srdiff,saharia2022imagen,rombach2021highresolution}, turning them into temporally consistent video super resolution models.
Building on LDMs, our method can generate globally coherent and long videos in a computationally and memory efficient manner. For synthesis at very high resolutions, the video upsampler only needs to operate locally, keeping training and computational requirements low.
We ablate our method and test on $512\times1024$ real driving scene videos, achieving state-of-the-art video quality, and synthesize videos of several minutes length. 
We also video fine-tune a powerful, publicly available text-to-image LDM, \textit{Stable Diffusion}~\cite{rombach2021highresolution}, and turn it into an efficient and powerful text-to-video generator with resolution up to $1280\times2048$. 
Since we only need to train the temporal alignment layers in that case, we can use a relatively small training set of captioned videos. 
By transferring the trained temporal layers to differently fine-tuned text-to-image LDMs, we demonstrate personalized text-to-video generation for the first time.
We hope our work opens new avenues for efficient digital content creation and autonomous driving simulation.

\looseness=-1

\textbf{Contributions.} \textit{(i)} We present an efficient approach for training high-resolution, long-term consistent video generation models based on LDMs. Our key insight is to leverage pre-trained image DMs and turn them into video generators by inserting temporal layers that learn to align images in a temporally consistent manner (\Cref{fig:pipeline,fig:ldm_figure}). \textit{(ii)} We further temporally fine-tune super resolution DMs, which are ubiquitous in the literature. \textit{(iii)} We achieve state-of-the-art high-resolution video synthesis performance on real driving scene videos, and we can generate multiple minute long videos. \textit{(iv)} We transform the publicly available \emph{Stable Diffusion} text-to-image LDM into a powerful and expressive text-to-video LDM, and \textit{(v)} show that the learned temporal layers can be combined with different image model checkpoints (\eg, \emph{DreamBooth}~\cite{ruiz2022dreambooth}).

\vspace{-2mm}
\section{Background}\label{sec:background}
\vspace{-1mm}
\begin{figure}[t!]
  \vspace{-0.2cm}
    \includegraphics[width=0.49\textwidth]{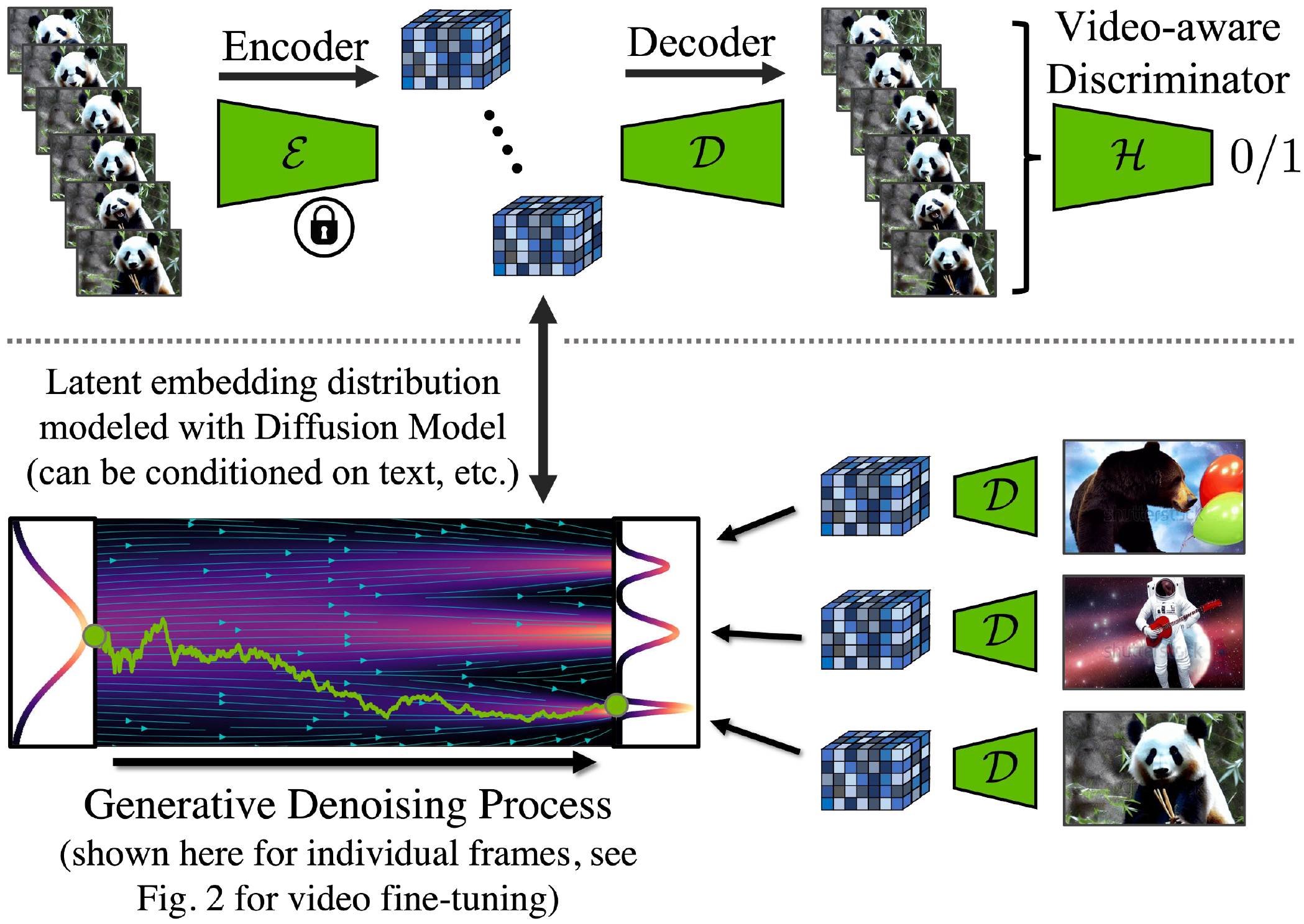}
    \caption{\small \textit{Top:} During temporal decoder fine-tuning, we process video sequences with a frozen encoder, which processes frames independently, and enforce temporally coherent reconstructions across frames. We additionally employ a video-aware discriminator. \textit{Bottom:} in LDMs, a diffusion model is trained in latent space. It synthesizes latent features, which are then transformed through the decoder into images. Note that the bottom visualization is for individual frames; see \Cref{fig:pipeline} for the video fine-tuning framework that generates temporally consistent frame sequences.} 
    \label{fig:ldm_figure}
  \vspace{-4mm}
\end{figure}
DMs~\cite{sohl2015deep,ho2020ddpm,song2020score} learn to model a data distribution $p_{\text{data}}(\rvx)$ via \emph{iterative denoising} and are trained with \textit{denoising score matching}~\cite{hyvarinen2005scorematching,lyu2009scorematching,vincent2011,sohl2015deep,song2019generative,ho2020ddpm,song2020score}: Given samples $\rvx \sim p_{\text{data}}$, \emph{diffused} inputs $\rvx_\tau = \alpha_\tau \rvx + \sigma_\tau \rvepsilon, \; \rvepsilon \sim \gN(\mathbf{0}, \mI)$ are constructed; $\alpha_{\tau}$ and $\sigma_\tau$ define a \emph{noise schedule}, parameterized via 
a diffusion-time $\tau$, such that the logarithmic signal-to-noise ratio $\lambda_\tau = \log(\alpha_{\tau}^2/\sigma_\tau^2)$ monotonically decreases. A denoiser model $\rvf_\theta$ (parameterized with learnable parameters $\theta$) receives the diffused $\rvx_\tau$ as input and is
optimized minimizing the denoising score matching objective
\begin{align}
\E_{\rvx \sim p_{\text{data}}, \tau \sim p_{\tau}, \rvepsilon \sim \gN(\mathbf{0}, \mI)} \left[\Vert \rvy - \rvf_\theta(\rvx_\tau; \rvc, \tau) \Vert_2^2 \right],
\label{eq:diffusionobjective}
\end{align}
where $\rvc$ is optional conditioning information, such as a text prompt, and the target vector $\rvy$ is either the random noise $\rvepsilon$ or $\rvv = \alpha_\tau \rvepsilon - \sigma_\tau \rvx$.
The latter objective (often referred to as \emph{$\rvv$-prediction}) has been introduced in the context of progressive distillation~\cite{salimans2022progressive} and empirically often yields faster convergence of the model (here, we use both objectives). Furthermore,
$p_\tau$ is a uniform distribution over the diffusion time $\tau$. The forward diffusion as well as the reverse generation process in diffusion models can be described via stochastic differential equations in a continuous-time framework~\cite{song2020score} (see~\Cref{fig:pipeline,fig:ldm_figure}), but in practice a fixed discretization can be used~\cite{ho2020ddpm}.
The maximum diffusion time is generally chosen such that the input data is entirely perturbed into Gaussian random noise and an iterative generative denoising process that employs the learned denoiser $\rvf_\theta$ can be initialized from such Gaussian noise to synthesize novel data.
Here, we use $p_\tau\sim\mathcal{U}\{0,1000\}$ and rely on a \emph{variance-preserving} noise schedule~\cite{song2020score}, for which $\sigma_\tau^2 = 1 - \alpha_\tau^2$ (see \Cref{app:arch,app:exp_details} for details). 

\textbf{Latent Diffusion Models (LDMs)}~\cite{rombach2021highresolution} improve in computational and memory efficiency over pixel-space DMs by first training a compression model to transform input images $\rvx{\sim}p_{\text{data}}$ into a spatially lower-dimensional latent space of reduced complexity, from which the original data can be reconstructed at high fidelity.
In practice, this approach is implemented with a regularized autoencoder, which reconstructs inputs $\rvx$ via an encoder module $\gE$ and a decoder $\gD$, such that the reconstruction $\hat{\rvx}{=}\gD(\gE(\rvx)){\approx}\rvx$ (\Cref{fig:ldm_figure}). To ensure photorealistic reconstructions, an adversarial objective can be added to the autoencoder training~\cite{rombach2021highresolution}, which is implemented using a patch-based discriminator~\cite{isola2017image}. 
A DM can then be trained in the compressed latent space and $\rvx$ in \Cref{eq:diffusionobjective} is replaced by its latent representation $\rvz{=}\gE(\rvx)$. This latent space DM can be typically smaller in terms of parameter count and memory consumption compared to corresponding pixel-space DMs of similar performance.

\vspace{-1.5mm}
\section{Latent Video Diffusion Models}\label{sec:video_ldm}
Here we describe how we \emph{video fine-tune} pre-trained image LDMs (and DM upsamplers) for high-resolution video synthesis. We assume access to a dataset $p_{\text{data}}$ of videos, such that $\rvx \in \R^{T \times 3 \times \tilde{H} \times \tilde{W}}, \; \rvx \sim p_{\text{data}}$ is a sequence of $T$ RGB frames, with height and width $\tilde{H}$ and $\tilde{W}$.
\begin{figure}[t!]
  \vspace{-0.3cm}
    \includegraphics[width=0.49\textwidth]{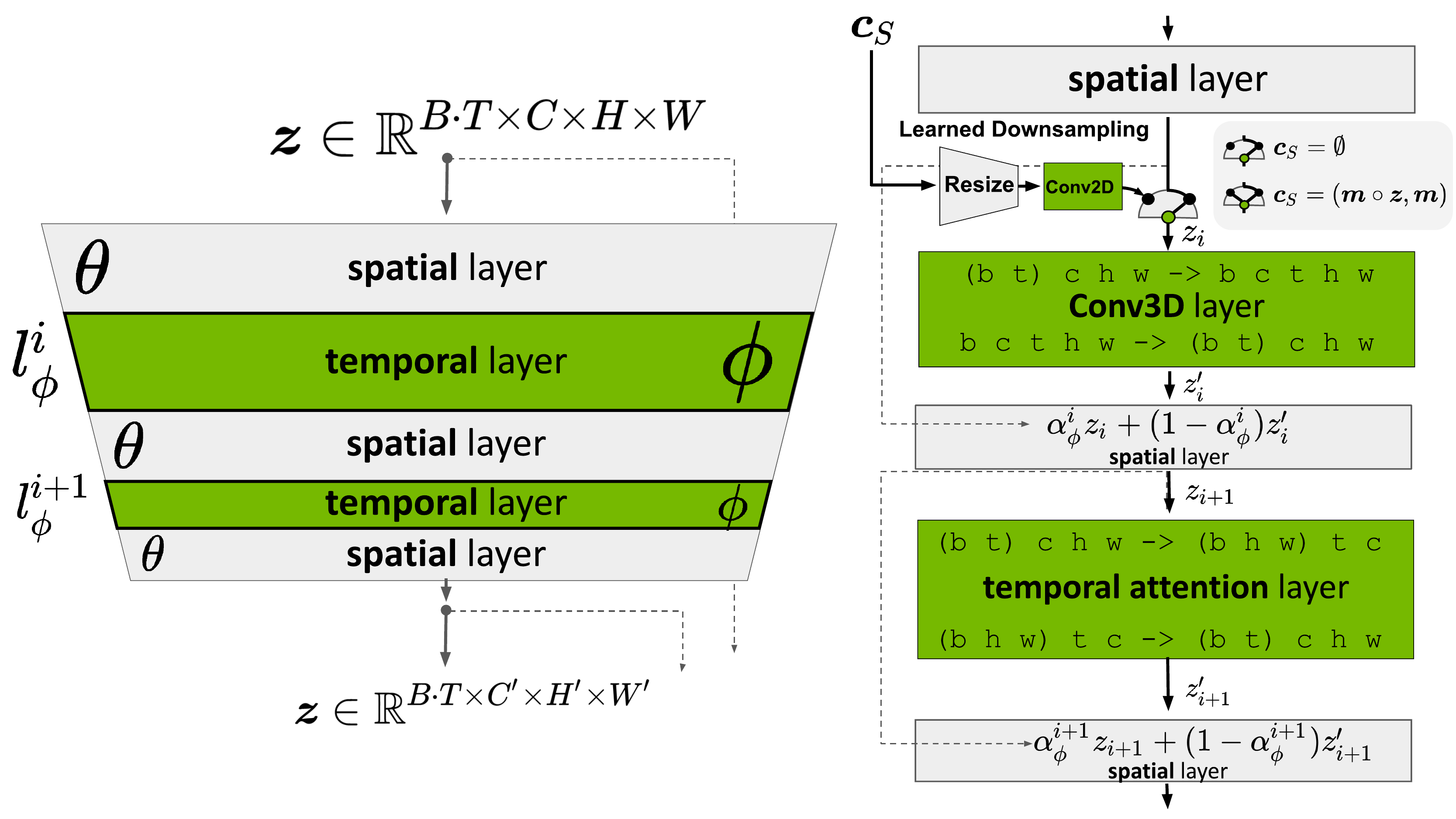}
    \caption{\small \textit{Left:} We turn a pre-trained LDM into a video generator by inserting \emph{temporal} layers that learn to align frames into temporally consistent sequences. During optimization, the image backbone $\theta$ remains fixed and only the parameters $\phi$ of the temporal layers $l_\phi^i$ are trained, \cf \Cref{eq:videoobjective}. \textit{Right:}
During training, the base model $\theta$ interprets the input sequence of length $T$ as a batch of images. For the temporal layers $\l_\phi^i$, these batches are reshaped into video format. Their output $\rvz'$ is combined with the spatial output $\rvz$, using a learned merge parameter $\alpha$.
During inference, skipping the temporal layers ($\alpha_\phi^i{=}1$)  yields the original image model.
    For illustration purposes, only a single U-Net Block is shown.
    $B$ denotes batch size, $T$ sequence length, $C$ input channels and $H$ and $W$ the spatial dimensions of the input. $\rvc_S$ is optional context frame conditioning, when training prediction models (\Cref{sec:prediction}).
    } 
    \label{fig:architecture}

   \vspace{-5.5mm}
\end{figure}
\subsection{Turning Latent Image into Video Generators}
\label{sec:turning}
Our key insight for efficiently training a video generation model is to re-use a pre-trained, fixed image generation model; an LDM parameterized by parameters $\theta$. Formally, let us denote the neural network layers that comprise the image LDM and process inputs over the pixel dimensions as \emph{spatial} layers $l_\theta^i$, with layer index $i$.
However, although such a model is able to synthesize individual frames at high quality, using it directly to render a video of $T$ consecutive frames will fail, as the model has no temporal awareness. 
We thus introduce additional \emph{temporal} neural network layers $l_\phi^i$, which are interleaved with the existing \emph{spatial} layers $l_\theta^i$ and learn to align individual frames in a temporally consistent manner. These $L$ additional temporal layers $\{l_\phi^i\}_{i=1}^L$ define the \emph{video-aware} temporal backbone of our model, and the full model $\rvf_{\theta, \phi}$ is thus the combination of the spatial and temporal layers; see~\Cref{fig:architecture} for a visualization.

We start from a frame-wise encoded input video $\mathcal{E}(\rvx)=\rvz \in \R^{T \times C \times H \times W}$, 
where $C$ is the number of latent channels and $H$ and $W$ are the spatial latent dimensions.
The spatial layers interpret the video as a batch of independent images (by shifting the temporal axis into the batch dimension), and for each \emph{temporal mixing layer} $l_\phi^i$, we reshape back to video dimensions as follows (using \texttt{einops}~\cite{rogozhnikov2022einops} notation):
\begin{align*}
\rvz' &\leftarrow \texttt{rearrange}(\rvz, \; \texttt{(b t) c h w} \rightarrow \texttt{b c t h w}) \\
\rvz' &\leftarrow l_\phi^i(\rvz', \rvc) \\
\rvz' &\leftarrow \texttt{rearrange}(\rvz', \; \texttt{b c t h w} \rightarrow \texttt{(b t) c h w}) ,
\end{align*}
where we added the batch dimension $\texttt{b}$ for clarity. %
In other words, the spatial layers treat all $B{\cdot}T$ encoded video frames independently in the batch dimension $\texttt{b}$, while the temporal layers $l_\phi^i(\rvz', \rvc)$ process entire videos in a new temporal dimension $\texttt{t}$. Furthermore, $\rvc$ is (optional) conditioning information such as a text prompt.
After each temporal layer, the output $\rvz'$ is combined with $\rvz$ as
$\alpha_\phi^i \rvz + (1 - \alpha_\phi^i) \rvz'$; 
$\alpha_\phi^i\in[0,1]$ denotes a (learnable) parameter (also \Cref{app:convolutional_ldm}).

In practice, we implement two different kinds of temporal mixing layers: (i) temporal attention and (ii) residual blocks based on 3D convolutions, \cf \Cref{fig:architecture}. We use sinusoidal embeddings~\cite{vaswani2017attention,ho2020ddpm} to provide the model with a positional encoding for time. 

Our video-aware temporal backbone is then trained using the same noise schedule as the underlying image model, and, importantly, we fix the spatial layers $l_\theta^i$ and \emph{only} optimize the temporal layers $l_\phi^i$ via
{\small\begin{align}
\argmin_{\phi} \E_{\rvx\sim p_{\text{data}}, \tau \sim p_{\tau}, \rvepsilon \sim \gN(\mathbf{0}, \mI)} \left[\Vert \rvy - \rvf_{\theta, \phi}(\rvz_\tau; \rvc, \tau) \Vert_2^2 \right],
\label{eq:videoobjective}
\end{align}}
where $\rvz_\tau$ denotes diffused encodings $\rvz=\mathcal{E}(\rvx)$.
This way, we retain the native image generation capabilities by simply skipping the temporal blocks, \eg by setting $\alpha_\phi^i = 1$ for each layer. A crucial advantage of our strategy is that huge image datasets can be used to pre-train the spatial layers, while the video data, which is often less widely available, can be utilized for focused training of the temporal layers.

\subsubsection{Temporal Autoencoder Finetuning}
\label{sec:dec_finetuna}
\begin{figure}[t!]
  \vspace{-0.4cm}
    \includegraphics[width=0.49\textwidth]{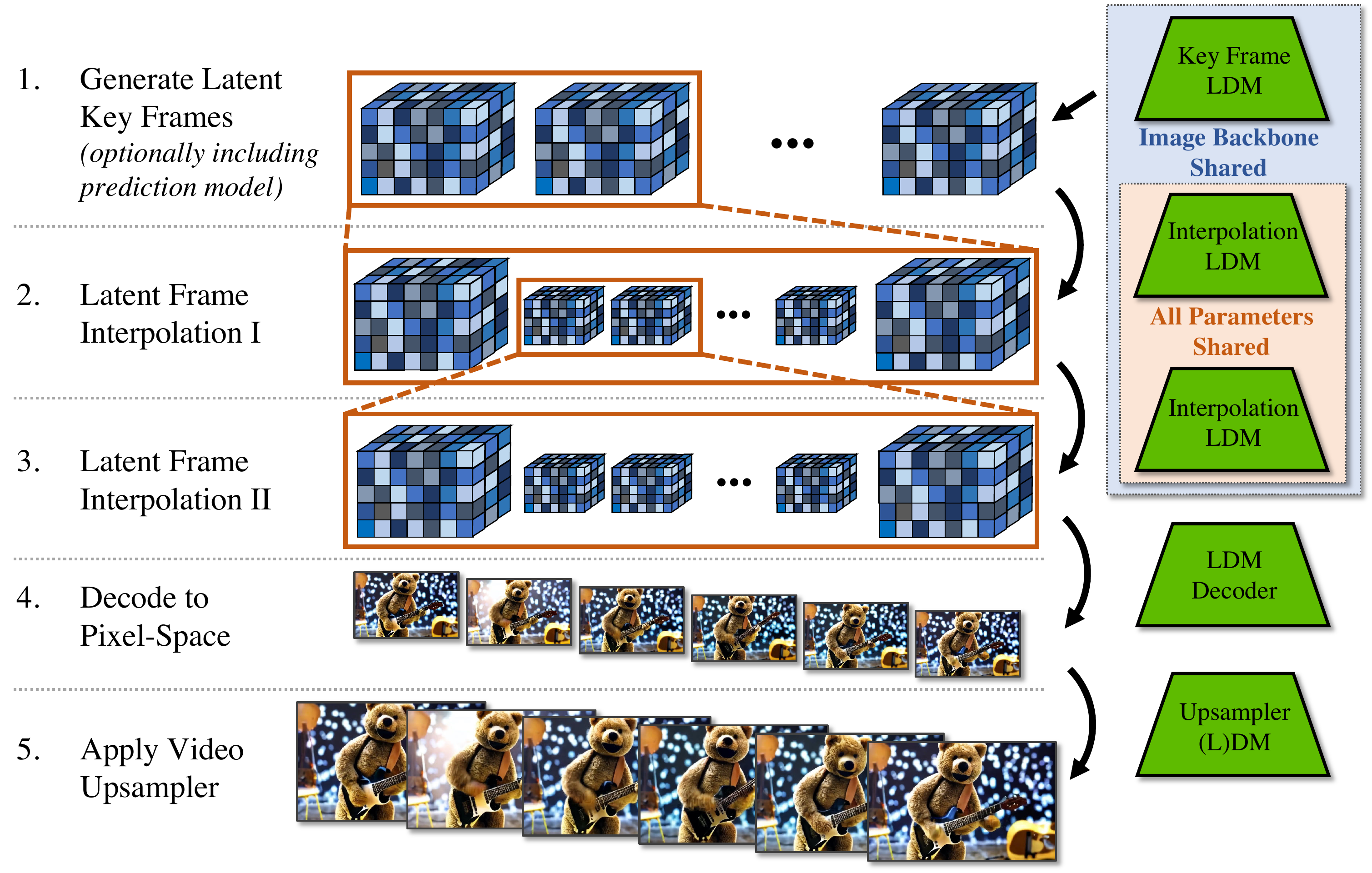}
    \caption{\small \textbf{Video LDM Stack.} We first generate sparse key frames. Then we temporally interpolate in two steps with the same interpolation model to achieve high frame rates. These operations are all based on latent diffusion models (LDMs) that share the same image backbone. Finally, the latent video is decoded to pixel space and optionally a video upsampler diffusion model is applied.} 
    \label{fig:stack_figure}
  \vspace{-4mm}
\end{figure}
Our video models build on pre-trained image LDMs. While this increases efficiency, the autoencoder of the LDM is trained on images only, causing flickering artifacts when encoding and decoding a temporally coherent sequence of images. %
To counteract this, we introduce additional temporal layers for the autoencoder's decoder, which we finetune on video data with a (patch-wise) temporal discriminator built from 3D convolutions, \cf ~\Cref{fig:ldm_figure}. Note that the encoder remains unchanged from image training such that the image DM that operates in latent space on encoded video frames can be re-used.
As demonstrated by computing reconstruction FVD~\cite{unterthiner2018towards} scores in~\Cref{tab:av_upsampler}, this step is critical for achieving good results.

\begin{figure*}[t!]
  \vspace{-0.8cm}
    \includegraphics[width=\textwidth]{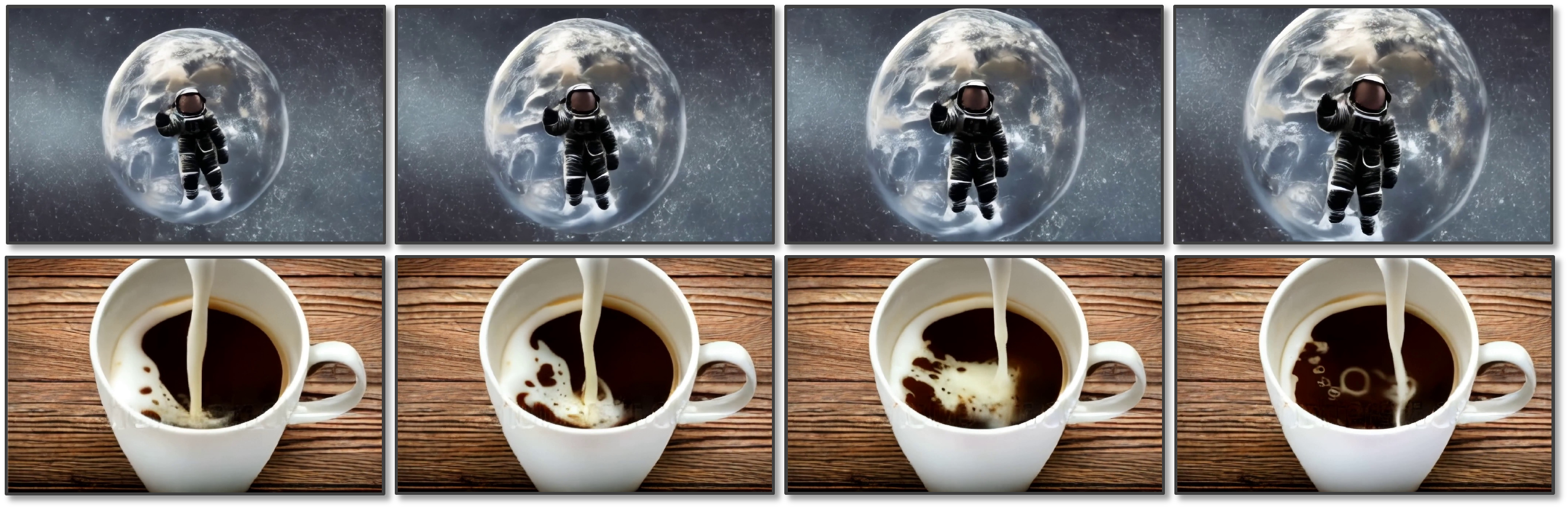}
    \vspace{-8mm}
    \caption{\small $1280\times 2048$ resolution samples from our Stable Diffusion-based text-to-video LDM, including video fine-tuned upsampler. Prompts: ``An astronaut flying in space, 4k, high resolution''  and ``Milk dripping into a cup of coffee, high definition, 4k''.} \vspace{-0.5em}
    \label{fig:text2image_samples}
  \vspace{-4mm}
\end{figure*}
\subsection{Prediction Models for Long-Term Generation}
\label{sec:prediction}
Although the approach described in \Cref{sec:turning} is efficient for generating short video sequences, it reaches its limits when it comes to synthesizing very long videos. Therefore, we also train models as \emph{prediction models} given a number of (first) $S$ context frames. We implement this by introducing a temporal binary mask $\rvm_S$ 
which masks the $T-S$ frames the model has to predict, where $T$ is the total sequence length as in \Cref{sec:turning}. We feed this mask and the masked encoded video frames into the model for conditioning.
Specifically, the frames are encoded with LDM's image encoder $\gE$, multiplied by the mask, and then fed (channel-wise concatenated with the masks) into the temporal layers $l_\phi^i$ after being processed with a learned downsampling operation, see~\Cref{fig:architecture}. Let $\rvc_S = (\rvm_S \circ \rvz, \rvm_S)$ denote the concatenated spatial conditioning of masks and masked (encoded) images. Then, the objective from \Cref{eq:videoobjective} reads
\begin{align}
\E_{\rvx \sim p_{\text{data}}, \rvm_S\sim p_S, \tau \sim p_{\tau}, \rvepsilon} \left[\Vert \rvy - \rvf_{\theta, \phi}(\rvz_\tau; \rvc_S, \rvc, \tau) \Vert_2^2 \right],
\label{eq:predictionobjective}
\end{align}
where $p_S$ represents the (categorical) mask sampling distribution.
In practice, we learn prediction models that condition either on 0, 1 or 2 context frames, allowing for classifier-free guidance as discussed below.

During inference, for generating long videos, we can apply the sampling process iteratively, re-using the latest predictions as new context. The first initial sequence is generated by synthesizing a single context frame from the base image model and generating a sequence based on that; afterwards, we condition on two context frames to encode movement (details in Appendix). 
To stabilize this process, we found it beneficial to use \emph{classifier-free diffusion guidance}~\cite{ho2021classifierfree}, where we guide the model during sampling via
{\small\begin{equation}
\rvf_{\theta, \phi}'(\rvz_\tau; \rvc_S) = \rvf_{\theta, \phi}(\rvz_\tau) + s \cdot \left(\rvf_{\theta, \phi}(\rvz_\tau; \rvc_S) - \rvf_{\theta, \phi}(\rvz_\tau) \right)
\end{equation}}
where $s{\geq}1$ denotes the guidance scale and we dropped the explicit conditioning on $\tau$ and other information $\rvc$ for readability. 
We refer to this guidance as \emph{context guidance}.

\subsection{Temporal Interpolation for High Frame Rates}
High-resolution video is characterized not only by high spatial resolution, but also by high temporal resolution, \ie, a high frame rate. 
To achieve this, we divide the synthesis process for high-resolution video into two parts: The first is the process described in \Cref{sec:turning} and \Cref{sec:prediction}, which can generate \emph{key frames} with large semantic changes, but (due to memory constraints) only at a relatively low frame rate. 
For the second part, we introduce an additional model whose task is to interpolate between given key frames. 
To implement this, we use the masking-conditioning mechanism introduced in \Cref{sec:prediction}. However, unlike the prediction task, we now mask the frames to be interpolated---otherwise, the mechanism remains the same, \ie, the image model is refined into a video interpolation model. In our experiments, we predict three frames between two given key frames, thereby training a $T \rightarrow 4T$ interpolation model. To achieve even larger frame rates, we train the model simultaneously in the $T \rightarrow 4T$ and $4T \rightarrow 16T$ regimes (using videos with different fps), specified by binary conditioning. 

Our training approach for prediction and interpolation models is inspired by recent works~\cite{voleti2022mcvd,harvey2022flexible,hoeppe2022diffusion} that use similar masking techniques (also see \Cref{sec:related_extended}).

\begin{figure*}[t!]
  \vspace{-0.8cm}
    \includegraphics[width=\textwidth]{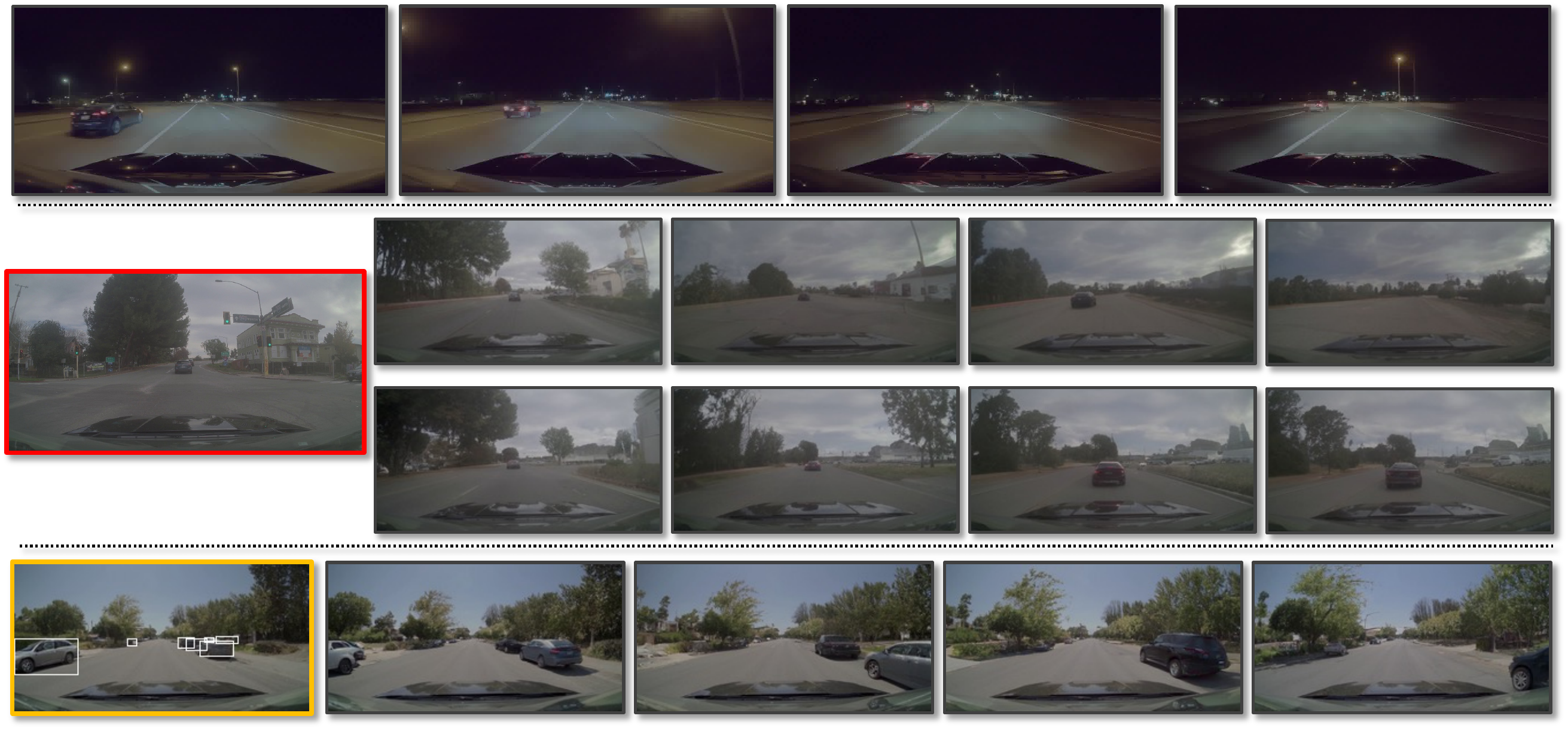}
    \vspace{-7mm}
    \caption{\small $512\times1024$ resolution video modeling of real-world driving scenes with our Video LDM and video upsampler. \emph{Top:} (Night time) \textbf{Driving Video Generation}. \emph{Middle:} \textbf{Multimodal Driving Scenario Prediction}: We simulate two different scenarios given the same initial frame (red). \emph{Bottom:} \textbf{Specific Driving Scenario Simulation}: We synthesize a scenario based on a manually designed, initial scene generated with a bounding box-conditioned Image LDM (yellow). More examples in the \Cref{app:extended_driving}.  \vspace{-0.5em}}
    \label{fig:av_samples}
  \vspace{-4mm}
\end{figure*}
\subsection{Temporal Fine-tuning of SR Models} \label{sec:method_upsampler}
Although the LDM mechanism already provides a good native resolution we aim to push this towards the megapixel range. We take inspiration from cascaded DMs~\cite{ho2021cascaded} and use a DM to further scale up the Video LDM outputs by $4\times$.
For our driving video synthesis experiments, we use a pixel-space DM~\cite{ho2021cascaded} (\Cref{sec:driving_exp}) and scale to $512\times1024$; for our text-to-video models, we use an LDM upsampler~\cite{rombach2021highresolution} (\Cref{sec:text-to-video}) and scale to $1280\times2048$.
We use noise augmentation with noise level conditioning~\cite{ho2021cascaded,saharia2022imagen} and train the super resolution (SR) model $\rvg_{\theta, \phi}$ (on images or latents) via
{\small\begin{align}
\E_{\rvx \sim p_{\text{data}}, (\tau,\tau_{\gamma}) \sim p_{\tau}, \rvepsilon \sim \gN(\mathbf{0}, \mI)} \left[\Vert \rvy - \rvg_{\theta, \phi}(\rvx_\tau; \rvc_{\tau_{\gamma}}, \tau_{\gamma}, \tau) \Vert_2^2 \right]
\label{eq:upscaleobjective}
\end{align}}
where $\rvc_{\tau_{\gamma}} = \alpha_{\tau_\gamma} \rvx + \sigma_{\tau_\gamma} \rvepsilon, \; \rvepsilon \sim \gN(\mathbf{0}, \mI)$, denotes a noisy low-resolution image given to the model via concatenation, and $\tau_\gamma$ the amount of noise added to the low-resolution image following the noise schedule $\alpha_\tau$, $\sigma_\tau$.

Since upsampling video frames independently would result in poor temporal consistency, we also
make this SR model video-aware. We follow the mechanism introduced in~\Cref{sec:turning} with spatial layers $l_\theta^i$ and temporal layers $l_\phi^i$ and similarly video fine-tune the upscaler, conditioning on a low-resolution sequence of length $T$ and concatenating low-resolution video images frame-by-frame. 
Since the upscaler operates locally, we conduct all upscaler training efficiently on patches only and later apply the model convolutionally.

Overall, we believe that the combination of an LDM with an upsampler DM is ideal for efficient high-resolution video synthesis. On the one hand, the main LDM component of our Video LDM leverages a computationally efficient, compressed latent space to perform all video modeling. This allows us to use large batch sizes and jointly encode more video frames, which benefits long-term video modeling, without excessive memory demands, as all video predictions and interpolations are carried out in latent space. On the other hand, the upsampler can be trained in an efficient patch-wise manner, therefore similarly saving computational resources and reducing memory consumption, and it also does not need to capture long-term temporal correlations due to the low-resolution conditioning. Therefore, no prediction and interpolation framework is required for this component. A model overview, bringing together all components from \Cref{sec:turning} to \Cref{sec:method_upsampler}, is depicted in~\Cref{fig:stack_figure}.

\textit{A discussion of related work can be found in \Cref{sec:related_extended}.}

\vspace{-1mm}
\section{Experiments}\label{sec:experiments}
\vspace{-1mm}
\textbf{Datasets.} Since we focus on driving scene video generation as well as text-to-video, we use two corresponding datasets/models: \textit{(i)} An in-house dataset of real driving scene (RDS) videos. The dataset consists of 683,060 videos of 8 seconds each at resolution $512\times1024$ ($H\times W$) and frame rate up to 30 fps. Furthermore, the videos have binary night/day labels, annotations for the number of cars in a scene (``crowdedness''), and a subset of the data also has car bounding boxes. \textit{(ii)} We use the  WebVid-10M~\cite{bain21frozen} dataset to turn the publicly available \textit{Stable Diffusion} Image LDM~\cite{rombach2021highresolution} into a Video LDM. WebVid-10M consists of 10.7M video-caption pairs with a total of 52K video hours. We resize the videos into resolution $320\times512$. 
\textit{(iii)} Moreover, in \Cref{app:mountain_bike}, we show experiments on the Mountain Biking dataset by Brooks et al.~\cite{brooks2022generating}. %

\textbf{Evaluation Metrics.} To evaluate our models, we use frame-wise Fr\'echet Inception Distance (FID)~\cite{heusel2017gans} as well as Fr\'echet Video Distance (FVD)~\cite{unterthiner2018towards}. Since FVD can be unreliable (discussed, for instance, by Brooks et al.~\cite{brooks2022generating}), we additionally perform human evaluation. For our text-to-video experiments, we also evaluate CLIP similarity (CLIPSIM)~\cite{wu2021godiva} and (video) inception score (IS) (\Cref{app:quant_eval}).

\textbf{Model Architectures and Sampling.} Our Image LDMs are based on Rombach et al.~\cite{rombach2021highresolution}. They use convolutional encoders and decoders, and their latent space DM architecture build on the U-Net by Dhariwal et al.\cite{dhariwal2021diffusion}. Our pixel-space upsampler DMs use the same Image DM backbone~\cite{dhariwal2021diffusion}. DM sampling is performed using DDIM~\cite{song2021denoising} in all experiments.\looseness=-1

Further architecture, training, evaluation, sampling and dataset details can be found in the Appendix.

\subsection{High-Resolution Driving Video Synthesis} \label{sec:driving_exp}
We train our Video LDM pipeline, including a $4\times$ pixel-space video upsampler, on the real driving scene (RDS) data. We condition on day/night labels and crowdedness, and randomly drop these labels during training to allow for classifier-free guidance and unconditional synthesis (we do not condition on bounding boxes here). Following the proposed training strategy above, we first train the image backbone LDM (spatial layers) on video frames independently, before we then train the temporal layers on videos. We also train Long Video GAN (LVG)~\cite{brooks2022generating}, the previous state-of-the-art in long-term high-resolution video synthesis, on the RDS data to serve as main baseline. \Cref{tab:av_main} (left) shows our main results for the Video LDM at $128\times256$ resolution, without upsampler. We show both performance of our model with and without conditioning on crowdedness and day/night. Our Video LDM generally outperforms LVG and adding conditioning further reduces FVD. \Cref{tab:av_user_study} shows our human evaluation: Our samples are generally preferred over LVG in terms of realism, and samples from our conditional model are also preferred over unconditional samples.\looseness=-1

Next, we compare our video fine-tuned pixel-space upsampler with independent frame-wise image upsampling (\Cref{tab:av_upsampler}), using $128\times256$ 30 fps ground truth videos for conditioning. We find that temporal alignment of the upsampler is crucial for high performance. FVD degrades significantly, if the video frames are upsampled independently, indicating loss of temporal consistency. As expected, FID is essentially unaffected, because the individual frames are still of high quality when upsampled independently.

In~\Cref{fig:teaser} (bottom) and \Cref{fig:av_samples} (top), we show conditional samples from the combined Video LDM and video upsampler model. We observe high-quality videos. Moreover, using our prediction approach, we find that we can generate very long, temporally coherent high-resolution driving videos of multiple minutes. We validated this for up to 5 minutes; see Appendix and supplementary video for results.

\begin{table}
    \vspace{-0.1cm}
    \centering
    \caption{\emph{Left:} Comparison with LVG on RDS; \emph{Right:}~Ablations.\vspace{-1em}}
    \label{tab:av_main}
    \resizebox{.48\linewidth}{!}{%
    \begin{tabular}{l c c}
        \toprule
        \textbf{Method} & FVD & FID \\
        \midrule
        LVG~\cite{brooks2022generating} & 478 & 53.5 \\
        \emph{Ours} & 389 & \textbf{31.6} \\
        \emph{Ours} (cond.) & \textbf{356} & 51.9 \\
        \bottomrule
    \end{tabular}
    }
    \hfill
    \resizebox{.48\linewidth}{!}{%
    \begin{tabular}{l c c}
        \toprule
        \textbf{Method} & FVD & FID \\
        \midrule
        Pixel-baseline & 639,56 & 59.70 \\
        End-to-end LDM & 1155.10 & 71.26 \\
        Attention-only & 704.41 & 50.01\\
        \midrule
        \emph{Ours} & 534.17 & \textbf{48.26} \\
        \emph{Ours} (context-guided) & \textbf{508.82} & 54.16 \\
        \bottomrule
    \end{tabular}
    }
    \vspace{-0.3em}
\end{table}

\begin{table}
    \centering
        \caption{User study on Driving Video Synthesis on RDS. \vspace{-1em}}
    \label{tab:av_user_study}
    
    \resizebox{ \linewidth}{!}{%
    \begin{tabular}{l  c c c}
        \toprule
        \textbf{Method} &   Pref. A &  Pref. B & Equal  \\
        \midrule
        \emph{Ours} (cond.) v.s \emph{Ours} (uncond.) & \textbf{49.33} & 42.67 &8.0  \\

        \emph{Ours} (uncond.) v.s LVG   & \textbf{54.02} & 40.23 & 5.74 \\
        \emph{Ours} (cond.) v.s LVG   & \textbf{62.03} & 31.65 & 6.33 \\
        \bottomrule
        \vspace{-1.5em}
    \end{tabular}
    }
\end{table}

\vspace{-1em}

\begin{table}[]
    \centering
    \caption{\emph{Left:} Evaluating temporal fine-tuning for diffusion upsamplers on RDS data; \emph{Right:} Video fine-tuning of the first stage decoder network leads to significantly improved consistency.~\vspace{-1.em}}
    \label{tab:av_upsampler}
    \hfill
    \resizebox{.53\linewidth}{!}{%
    \begin{tabular}{l c c}
        \toprule
        \textbf{Method} & FVD & FID \\
        \midrule
        \emph{Ours} Image Upsampler & 165.98 & \textbf{19.71}  \\
        \emph{Ours} Video Upsampler & \textbf{45.39} & 19.85 \\
        \bottomrule
    \end{tabular}
}
    \hfill
    \resizebox{.45\linewidth}{!}{%
    \begin{tabular}{l c c }
        \toprule
         \textbf{Decoder} & \footnotesize\emph{image-only}  & \footnotesize\emph{finetuned} \\
        \midrule
        FVD & 390.88 & \textbf{32.94} \\
        FID & \textbf{7.61} & 9.17 \\
        \bottomrule
    \end{tabular}
    }
    \vspace{-1.3em}
\end{table}
\begin{figure*}[t!]
  \vspace{-0.8cm}
    \includegraphics[width=\textwidth]{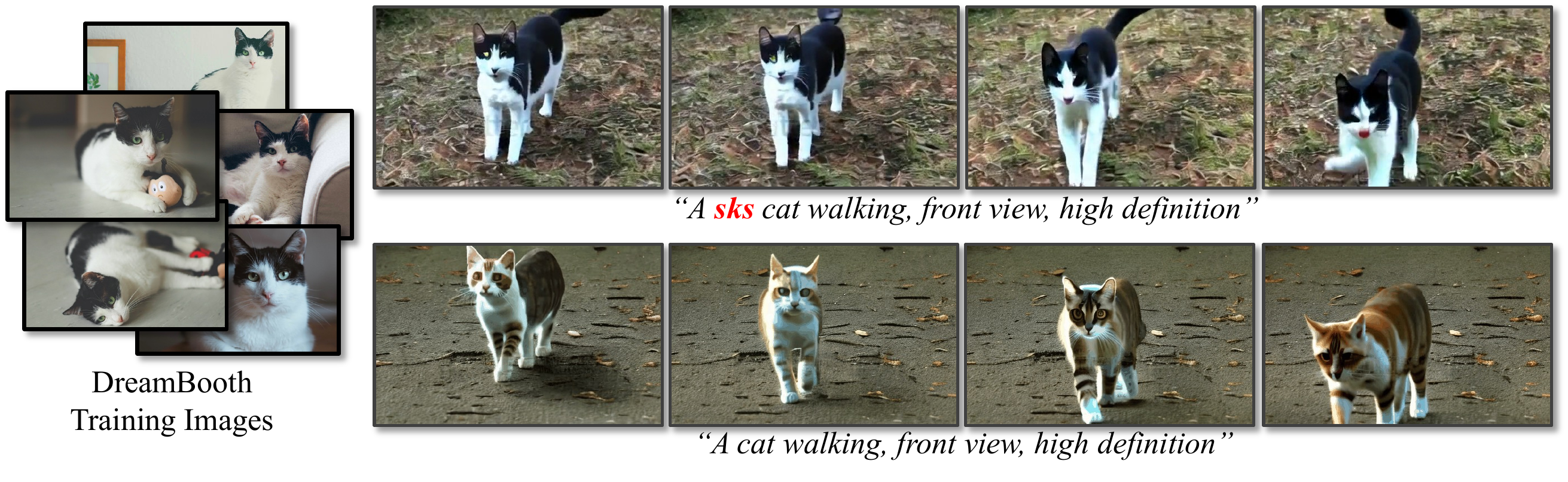}
    \vspace{-8.5mm}
    \caption{\small \emph{Left:} DreamBooth Training Images. \emph{Top row:} Video generated by our Video LDM with DreamBooth Image LDM backbone. \emph{Bottom row:} Video generated without DreamBooth Image backbone. We see that the DreamBooth model preserves subject identity well. \vspace{-0.5em}}
    \label{fig:dreambooth}
  \vspace{-2.5mm}
\end{figure*}
\subsubsection{Ablation Studies}\label{sec:ablations}
To show the efficacy of our design choices (\Cref{sec:video_ldm}), we compare a smaller version of our Video LDM with various baselines on the RDS dataset and present the results in~\Cref{tab:av_main} (right) (for evaluation details, see \Cref{app:quant_eval}). First, using the exact same architecture as for our Video LDM, we apply our temporal finetuning strategy to a pre-trained pixel-space image diffusion model, which is clearly outperformed by ours. Further, we train an End-to-End LDM, whose entire set of parameters $\{\theta, \phi\}$ is learned on RDS videos without image pre-training of $\theta$, leading to heavy degradations both in FID and FVD, when compared with our Video LDM. Another important architectural choice is the introduction of 3D convolutional temporal layers, since they allow us to feed the context frames $\boldsymbol{c}_{S}$ to the network spatially. This model achieves both lower FVD and FID scores than an attention-only temporal model, which uses the same set of spatial layers $\theta$ and has the same number of trainable parameters. Finally, we see that we can further lower FVD scores by applying \emph{context guidance} while sacrificing a bit of visual quality indicated by increased FID scores.

Moreover, we provide an analysis on the effects of video fine-tuning the decoder of the compression model (\cf~\Cref{sec:dec_finetuna}) which encompasses the LDM framework~\cite{rombach2021highresolution}. We apply our fine-tuning strategy to decoders of these compression models on the RDS
dataset and compare both the obtained FVD/FID scores of reconstructed videos/image frames with those of their non-video-finetuned counterparts. 
Video fine-tuning leads to improvements by orders of magnitudes, as can be seen in Table~\ref{tab:av_upsampler}.

\vspace{-2mm}
\subsubsection{Driving Scenario Simulation}\label{sec:driving_simulation}
\vspace{-1mm}
A high-resolution video generator trained on in-the-wild driving scenes can potentially serve as a powerful simulation engine. We qualitatively explore this in~\Cref{fig:av_samples}. Given an initial frame, our video model can generate several different plausible future predictions. Furthermore, we also trained a separate, bounding box-conditioned image LDM on our data (only for image synthesis). A user can now manually create a scene composition of interest by specifying the bounding boxes of different cars, generate a corresponding image, and then use this image as initialization for our Video LDM, which can then predict different scenarios in a multimodal fashion (bottom in~\Cref{fig:av_samples}).

\subsection{Text-to-Video with Stable Diffusion}\label{sec:text-to-video}
\vspace{-1mm}
Instead of first training our own Image LDM backbone, our Video LDM approach can also leverage existing Image LDMs and turn them into video generators. To demonstrate this, we turn the publicly available text-to-image LDM \emph{Stable Diffusion} into a text-to-video generator. Specifically, using the WebVid-10M text-captioned video dataset, we train a temporally aligned version of Stable Diffusion for text-conditioned video synthesis. We briefly fine-tune Stable Diffusion's spatial layers on frames from WebVid, and then insert the temporal alignment layers and train them (at resolution $320\times512$). We also add text-conditioning in those alignment layers. Moreover, we further video fine-tune the publicly available latent \emph{Stable Diffusion upsampler}, which enables $4\times$ upscaling and allows us to generate videos at resolution $1280\times 2048$. We generate videos consisting of 113 frames, which we can render, for instance, into clips of 4.7 seconds length at 24 fps or into clips of 3.8 seconds length at 30 fps.
Samples from the trained models are shown in~\Cref{fig:teaser,fig:text2image_samples}. While WebVid-10M consists of photo-quality real-life videos, we are able to generate highly expressive and artistic videos beyond the video training data. This demonstrates that the general image generation capabilities of the Image LDM backbone readily translate to video generation, even though the video dataset we trained on is much smaller and limited in diversity and style. The Video LDM effectively combines the styles and expressions from the image model with the movements and temporal consistency learnt from the WebVid videos.

We evaluate zero-shot text-to-video generation on UCF-101~\cite{soomro2012ucf101} and MSR-VTT~\cite{xu2016msr-vtt} (Tabs.~\ref{tab:ucf} \&~\ref{tab:msrvtt}).
Evaluation details in \Cref{app:quant_eval}.
We significantly outperform all baselines except Make-A-Video~\cite{singer2022make}, which we still surpass in IS on UCF-101. However, Make-A-Video is concurrent work, focuses entirely on text-to-video and trains with more video data than we do. We use only WebVid-10M; Make-A-Video also uses HD-VILA-100M~\cite{xue2022hdvila}.

\begin{table}[t]
    \centering
        \caption{\small UCF-101 text-to-video generation.\vspace{-1em}}
    \label{tab:ucf}
    
    \resizebox{ 0.77\linewidth}{!}{%
    \begin{tabular}{l  c c c}
        \toprule
        \textbf{Method} &  Zero-Shot & IS ($\uparrow$) & FVD ($\downarrow$)  \\
        \midrule
        CogVideo (Chinese)~\cite{hong2022cogvideo} & Yes  & 23.55 & 751.34 \\
        CogVideo (English)~\cite{hong2022cogvideo} & Yes  & 25.27 & 701.59 \\
        MagicVideo~\cite{zhou2022magicvideo} & Yes  & - & 699.00 \\
        Make-A-Video~\cite{singer2022make} & Yes  & 33.00 & 367.23 \\
        \midrule
        Video LDM \emph{(Ours)} & Yes & 33.45 & 550.61 \\
        \bottomrule
        \vspace{-2.2em}
    \end{tabular}
    }
\end{table}
\begin{table}[t]
    \centering
        \caption{\small MSR-VTT text-to-video generation performance.\vspace{-1em}}
    \label{tab:msrvtt}
    
    \resizebox{ 0.77\linewidth}{!}{%
    \begin{tabular}{l  c c}
        \toprule
        \textbf{Method} &  Zero-Shot & CLIPSIM ($\uparrow$)  \\
        \midrule
        GODIVA~\cite{wu2021godiva} & No  & 0.2402 \\
        N\"{U}WA~\cite{wu2022nuwa} & No & 0.2439 \\
        CogVideo (Chinese)~\cite{hong2022cogvideo} & Yes  & 0.2614 \\
        CogVideo (English)~\cite{hong2022cogvideo} & Yes  & 0.2631 \\
        Make-A-Video~\cite{singer2022make} & Yes  & 0.3049\\
        \midrule
        Video LDM \emph{(Ours)} & Yes & 0.2929 \\
        \bottomrule
        \vspace{-2.9em}
    \end{tabular}
    }
\end{table}

In \Cref{app:convolutional_ldm}, we show how we can apply our model ``convolutional in time'' and ``convolutional in space'', enabling longer and spatially-extended generation without upsampler and prediction models. More video samples shown in \Cref{app:extended_text2video}. Experiment details in \Cref{app:exp_details_text_to_video}.

\vspace{-3mm}
\subsubsection{Personalized Text-to-Video with Dreambooth} \label{sec:dreambooth}
\vspace{-1mm}
Since we have separate spatial and temporal layers in our Video LDM, the question arises whether the temporal layers trained on one Image LDM backbone transfer to other model checkpoints (\eg fine-tuned). 
We test this for personalized text-to-video generation: Using DreamBooth~\cite{ruiz2022dreambooth}, we fine-tune our Stable Diffusion spatial backbone on small sets of images of certain objects, tying their identity to a rare text token (``\emph{sks}''). We then insert the temporal layers from the previously video-tuned Stable Diffusion (without DreamBooth) into the new DreamBooth version of the original Stable Diffusion model and generate videos using the token tied to the training images for DreamBooth (see Fig.~\ref{fig:dreambooth} and examples in \Cref{app:more_dreambooth}). We find that we can generate personalized coherent videos that correctly capture the identity of the Dreambooth training images. This validates that our temporal layers generalize to other Image LDMs. To the best of our knowledge, we are the first to demonstrate personalized text-to-video generation.\looseness=-1

\textit{Additional results and experiments in \Cref{app:additional_results}.}

\vspace{-1.7mm}
\section{Conclusions}\label{sec:conclusions}
 \vspace{-1.7mm}
We presented \textit{Video Latent Diffusion Models} for efficient high-resolution video generation. Our key design choice is to build on pre-trained image diffusion models and to turn them into video generators by temporally video fine-tuning them with temporal alignment layers. To maintain computational efficiency, we leverage LDMs, optionally combined with a super resolution DM, which we also temporally align. Our Video LDM can synthesize high-resolution and temporally coherent driving scene videos of many minutes. We also turn the publicly available \textit{Stable Diffusion} text-to-image LDM into an efficient text-to-video LDM and show that the learned temporal layers transfer to different model checkpoints. We leverage this for personalized text-to-video generation. We hope that our work can benefit simulators in the context of autonomous driving research and 
help democratize high quality video content creation (see \Cref{app:impact} for broader impact and limitations).

{\small
\bibliographystyle{ieee_fullname}
\addcontentsline{toc}{section}{References}
\bibliography{egbib}

\begin{thebibliography}{100}\itemsep=-1pt

\bibitem{babaeizadeh2018stochastic}
Mohammad Babaeizadeh, Chelsea Finn, Dumitru Erhan, Roy~H. Campbell, and Sergey
  Levine.
\newblock Stochastic variational video prediction.
\newblock In {\em International Conference on Learning Representations}, 2018.

\bibitem{bain21frozen}
Max Bain, Arsha Nagrani, G{\"u}l Varol, and Andrew Zisserman.
\newblock Frozen in time: A joint video and image encoder for end-to-end
  retrieval.
\newblock In {\em IEEE International Conference on Computer Vision}, 2021.

\bibitem{balaji2022eDiffi}
Yogesh Balaji, Seungjun Nah, Xun Huang, Arash Vahdat, Jiaming Song, Karsten
  Kreis, Miika Aittala, Timo Aila, Samuli Laine, Bryan Catanzaro, Tero Karras,
  and Ming-Yu Liu.
\newblock ediff-i: Text-to-image diffusion models with ensemble of expert
  denoisers.
\newblock {\em arXiv preprint arXiv:2211.01324}, 2022.

\bibitem{bao2022analyticdpm}
Fan Bao, Chongxuan Li, Jun Zhu, and Bo Zhang.
\newblock Analytic-dpm: an analytic estimate of the optimal reverse variance in
  diffusion probabilistic models.
\newblock In {\em International Conference on Learning Representations}, 2022.

\bibitem{ipoke}
Andreas Blattmann, Timo Milbich, Michael Dorkenwald, and Bj{\"{o}}rn Ommer.
\newblock ipoke: Poking a still image for controlled stochastic video
  synthesis.
\newblock In {\em 2021 {IEEE/CVF} International Conference on Computer Vision,
  {ICCV} 2021, Montreal, QC, Canada, October 10-17, 2021}, 2021.

\bibitem{brooks2022generating}
Tim Brooks, Janne Hellsten, Miika Aittala, Ting-Chun Wang, Timo Aila, Jaakko
  Lehtinen, Ming-Yu Liu, Alexei~A Efros, and Tero Karras.
\newblock Generating long videos of dynamic scenes.
\newblock {\em arXiv:2206.03429}, 2022.

\bibitem{carreira2017quo}
Joao Carreira and Andrew Zisserman.
\newblock Quo vadis, action recognition? a new model and the kinetics dataset.
\newblock In {\em proceedings of the IEEE Conference on Computer Vision and
  Pattern Recognition}, pages 6299--6308, 2017.

\bibitem{hvrnn}
Lluis Castrejon, Nicolas Ballas, and Aaron Courville.
\newblock Improved conditional vrnns for video prediction.
\newblock In {\em The IEEE International Conference on Computer Vision (ICCV)},
  October 2019.

\bibitem{svg}
Emily Denton and Rob Fergus.
\newblock Stochastic video generation with a learned prior.
\newblock In {\em Proceedings of the 35th International Conference on Machine
  Learning, {ICML} 2018, Stockholmsm{\"{a}}ssan, Stockholm, Sweden, July 10-15,
  2018}, 2018.

\bibitem{dhariwal2021diffusion}
Prafulla Dhariwal and Alexander~Quinn Nichol.
\newblock Diffusion models beat {GAN}s on image synthesis.
\newblock In {\em Advances in Neural Information Processing Systems}, 2021.

\bibitem{dockhorn2022genie}
Tim Dockhorn, Arash Vahdat, and Karsten Kreis.
\newblock Genie: Higher-order denoising diffusion solvers.
\newblock In {\em Advances in Neural Information Processing Systems}, 2022.

\bibitem{dockhorn2022score}
Tim Dockhorn, Arash Vahdat, and Karsten Kreis.
\newblock Score-based generative modeling with critically-damped langevin
  diffusion.
\newblock In {\em International Conference on Learning Representations (ICLR)},
  2022.

\bibitem{si2v}
Michael Dorkenwald, Timo Milbich, Andreas Blattmann, Robin Rombach,
  Konstantinos~G. Derpanis, and Bj{\"{o}}rn Ommer.
\newblock Stochastic image-to-video synthesis using cinns.
\newblock In {\em {IEEE} Conference on Computer Vision and Pattern Recognition,
  {CVPR} 2021, virtual, June 19-25, 2021}, 2021.

\bibitem{esser2023structure}
Patrick Esser, Johnathan Chiu, Parmida Atighehchian, Jonathan Granskog, and
  Anastasis Germanidis.
\newblock Structure and content-guided video synthesis with diffusion models.
\newblock {\em arXiv preprint arXiv:2302.03011}, 2023.

\bibitem{esser2020taming}
Patrick Esser, Robin Rombach, and Björn Ommer.
\newblock Taming transformers for high-resolution image synthesis.
\newblock {\em arXiv preprint arXiv:2012.09841}, 2020.

\bibitem{fox2021stylevideogan}
Gereon Fox, Ayush Tewari, Mohamed Elgharib, and Christian Theobalt.
\newblock Stylevideogan: A temporal generative model using a pretrained
  stylegan.
\newblock In {\em British Machine Vision Conference (BMVC)}, 2021.

\bibitem{lsvg}
Jean-Yves Franceschi, Edouard Delasalles, Micka\"{e}l Chen, Sylvain Lamprier,
  and Patrick Gallinari.
\newblock Stochastic latent residual video prediction.
\newblock In {\em Proceedings of the 37th International Conference on Machine
  Learning}, 2020.

\bibitem{gal2022animage}
Rinon Gal, Yuval Alaluf, Yuval Atzmon, Or Patashnik, Amit~H. Bermano, Gal
  Chechik, and Daniel Cohen-Or.
\newblock An image is worth one word: Personalizing text-to-image generation
  using textual inversion.
\newblock {\em arXiv preprint arXiv:2208.01618}, 2022.

\bibitem{ge2022longvideo}
Songwei Ge, Thomas Hayes, Harry Yang, Xi Yin, Guan Pang, David Jacobs, Jia-Bin
  Huang, and Devi Parikh.
\newblock Long video generation with time-agnostic vqgan and time-sensitive
  transformer.
\newblock In Shai Avidan, Gabriel Brostow, Moustapha Ciss{\'e}, Giovanni~Maria
  Farinella, and Tal Hassner, editors, {\em Computer Vision -- ECCV 2022},
  pages 102--118, Cham, 2022. Springer Nature Switzerland.

\bibitem{ge2022long}
Songwei Ge, Thomas Hayes, Harry Yang, Xi Yin, Guan Pang, David Jacobs, Jia-Bin
  Huang, and Devi Parikh.
\newblock Long video generation with time-agnostic vqgan and time-sensitive
  transformer.
\newblock {\em arXiv preprint arXiv:2204.03638}, 2022.

\bibitem{goodfellow2014generative}
Ian Goodfellow, Jean Pouget-Abadie, Mehdi Mirza, Bing Xu, David Warde-Farley,
  Sherjil Ozair, Aaron Courville, and Yoshua Bengio.
\newblock Generative adversarial nets.
\newblock {\em Advances in neural information processing systems}, 27, 2014.

\bibitem{Gupta_2022_CVPR}
Sonam Gupta, Arti Keshari, and Sukhendu Das.
\newblock Rv-gan: Recurrent gan for unconditional video generation.
\newblock In {\em Proceedings of the IEEE/CVF Conference on Computer Vision and
  Pattern Recognition (CVPR) Workshops}, pages 2024--2033, June 2022.

\bibitem{gupta2018imagine}
Tanmay Gupta, Dustin Schwenk, Ali Farhadi, Derek Hoiem, and Aniruddha Kembhavi.
\newblock Imagine this! scripts to compositions to videos.
\newblock In Vittorio Ferrari, Martial Hebert, Cristian Sminchisescu, and Yair
  Weiss, editors, {\em Computer Vision -- ECCV 2018}, pages 610--626, Cham,
  2018. Springer International Publishing.

\bibitem{harvey2022flexible}
William Harvey, Saeid Naderiparizi, Vaden Masrani, Christian Weilbach, and
  Frank Wood.
\newblock Flexible diffusion modeling of long videos.
\newblock {\em arXiv preprint arXiv:2205.11495}, 2022.

\bibitem{hertz2022prompt}
Amir Hertz, Ron Mokady, Jay Tenenbaum, Kfir Aberman, Yael Pritch, and Daniel
  Cohen-Or.
\newblock Prompt-to-prompt image editing with cross attention control.
\newblock {\em arXiv preprint arXiv:2208.01626}, 2022.

\bibitem{heusel2017gans}
Martin Heusel, Hubert Ramsauer, Thomas Unterthiner, Bernhard Nessler, and Sepp
  Hochreiter.
\newblock Gans trained by a two time-scale update rule converge to a local nash
  equilibrium.
\newblock In I. Guyon, U.~V. Luxburg, S. Bengio, H. Wallach, R. Fergus, S.
  Vishwanathan, and R. Garnett, editors, {\em Advances in Neural Information
  Processing Systems}, volume~30. Curran Associates, Inc., 2017.

\bibitem{ho2022imagenvideo}
Jonathan Ho, William Chan, Chitwan Saharia, Jay Whang, Ruiqi Gao, Alexey
  Gritsenko, Diederik~P. Kingma, Ben Poole, Mohammad Norouzi, David~J. Fleet,
  and Tim Salimans.
\newblock Imagen video: High definition video generation with diffusion models.
\newblock {\em arXiv preprint arXiv:2210.02303}, 2022.

\bibitem{ho2020ddpm}
Jonathan Ho, Ajay Jain, and Pieter Abbeel.
\newblock Denoising diffusion probabilistic models.
\newblock In {\em Advances in Neural Information Processing Systems}, 2020.

\bibitem{ho2021cascaded}
Jonathan Ho, Chitwan Saharia, William Chan, David~J Fleet, Mohammad Norouzi,
  and Tim Salimans.
\newblock Cascaded diffusion models for high fidelity image generation.
\newblock {\em arXiv preprint arXiv:2106.15282}, 2021.

\bibitem{ho2021classifierfree}
Jonathan Ho and Tim Salimans.
\newblock Classifier-free diffusion guidance.
\newblock In {\em NeurIPS 2021 Workshop on Deep Generative Models and
  Downstream Applications}, 2021.

\bibitem{ho2022video}
Jonathan Ho, Tim Salimans, Alexey Gritsenko, William Chan, Mohammad Norouzi,
  and David~J. Fleet.
\newblock Video diffusion models.
\newblock {\em arXiv preprint arXiv:2204.03458}, 2022.

\bibitem{hong2022cogvideo}
Wenyi Hong, Ming Ding, Wendi Zheng, Xinghan Liu, and Jie Tang.
\newblock Cogvideo: Large-scale pretraining for text-to-video generation via
  transformers.
\newblock {\em arXiv:2205.15868}, 2022.

\bibitem{hoeppe2022diffusion}
Tobias H{\"o}ppe, Arash Mehrjou, Stefan Bauer, Didrik Nielsen, and Andrea
  Dittadi.
\newblock Diffusion models for video prediction and infilling.
\newblock {\em arXiv preprint arXiv:2206.07696}, 2022.

\bibitem{hyvarinen2005scorematching}
Aapo Hyv\"{a}rinen.
\newblock Estimation of non-normalized statistical models by score matching.
\newblock {\em Journal of Machine Learning Research}, 6:695–709, 2005.

\bibitem{isola2017image}
Phillip Isola, Jun-Yan Zhu, Tinghui Zhou, and Alexei~A Efros.
\newblock Image-to-image translation with conditional adversarial networks.
\newblock In {\em Proceedings of the IEEE conference on computer vision and
  pattern recognition}, pages 1125--1134, 2017.

\bibitem{jolicoeur2021gotta}
Alexia Jolicoeur-Martineau, Ke Li, R{\'e}mi Pich{\'e}-Taillefer, Tal Kachman,
  and Ioannis Mitliagkas.
\newblock Gotta go fast when generating data with score-based models.
\newblock {\em arXiv:2105.14080}, 2021.

\bibitem{kahembwe2020lower}
Emmanuel Kahembwe and Subramanian Ramamoorthy.
\newblock Lower dimensional kernels for video discriminators.
\newblock {\em Neural Networks}, 132:506--520, 2020.

\bibitem{karras2021aliasfree}
Tero Karras, Miika Aittala, Samuli Laine, Erik H\"ark\"onen, Janne Hellsten,
  Jaakko Lehtinen, and Timo Aila.
\newblock Alias-free generative adversarial networks.
\newblock In {\em Proc. NeurIPS}, 2021.

\bibitem{karras2019style}
Tero Karras, Samuli Laine, and Timo Aila.
\newblock A style-based generator architecture for generative adversarial
  networks.
\newblock In {\em Proceedings of the IEEE/CVF Conference on Computer Vision and
  Pattern Recognition}, pages 4401--4410, 2019.

\bibitem{karras2020analyzing}
Tero Karras, Samuli Laine, Miika Aittala, Janne Hellsten, Jaakko Lehtinen, and
  Timo Aila.
\newblock Analyzing and improving the image quality of stylegan.
\newblock In {\em Proceedings of the IEEE/CVF Conference on Computer Vision and
  Pattern Recognition}, pages 8110--8119, 2020.

\bibitem{kawar2022restoration}
Bahjat Kawar, Michael Elad, Stefano Ermon, and Jiaming Song.
\newblock Denoising diffusion restoration models.
\newblock {\em arXiv preprint arXiv:2201.11793}, 2022.

\bibitem{lee2018savp}
Alex~X. Lee, Richard Zhang, Frederik Ebert, Pieter Abbeel, Chelsea Finn, and
  Sergey Levine.
\newblock Stochastic adversarial video prediction.
\newblock {\em arXiv preprint arXiv:1804.01523}, 2018.

\bibitem{li2022srdiff}
Haoying Li, Yifan Yang, Meng Chang, Shiqi Chen, Huajun Feng, Zhihai Xu, Qi Li,
  and Yueting Chen.
\newblock Srdiff: Single image super-resolution with diffusion probabilistic
  models.
\newblock {\em Neurocomputing}, 479:47--59, 2022.

\bibitem{li2017video}
Yitong Li, Martin~Renqiang Min, Dinghan Shen, David Carlson, and Lawrence
  Carin.
\newblock Video generation from text.
\newblock {\em arXiv preprint arXiv:1710.00421}, 2017.

\bibitem{liu2022pseudo}
Luping Liu, Yi Ren, Zhijie Lin, and Zhou Zhao.
\newblock Pseudo numerical methods for diffusion models on manifolds.
\newblock In {\em International Conference on Learning Representations}, 2022.

\bibitem{lu2022dpm}
Cheng Lu, Yuhao Zhou, Fan Bao, Jianfei Chen, Chongxuan Li, and Jun Zhu.
\newblock Dpm-solver: A fast ode solver for diffusion probabilistic model
  sampling in around 10 steps.
\newblock {\em arXiv:2206.00927}, 2022.

\bibitem{Luc2020TransformationbasedAV}
Pauline Luc, Aidan Clark, Sander Dieleman, Diego de Las~Casas, Yotam Doron,
  Albin Cassirer, and Karen Simonyan.
\newblock Transformation-based adversarial video prediction on large-scale
  data.
\newblock {\em ArXiv}, 2020.

\bibitem{lugmayr2022repaint}
Andreas Lugmayr, Martin Danelljan, Andres Romero, Fisher Yu, Radu Timofte, and
  Luc~Van Gool.
\newblock Repaint: Inpainting using denoising diffusion probabilistic models.
\newblock {\em arXiv preprint arXiv:2201.09865}, 2022.

\bibitem{luhman2021knowledge}
Eric Luhman and Troy Luhman.
\newblock Knowledge distillation in iterative generative models for improved
  sampling speed.
\newblock {\em arXiv preprint arXiv:2101.02388}, 2021.

\bibitem{lyu2009scorematching}
Siwei Lyu.
\newblock Interpretation and generalization of score matching.
\newblock In {\em Proceedings of the Twenty-Fifth Conference on Uncertainty in
  Artificial Intelligence}, UAI ’09, page 359–366, Arlington, Virginia,
  USA, 2009. AUAI Press.

\bibitem{marwah2017attentive}
Tanya Marwah, Gaurav Mittal, and Vineeth~N. Balasubramanian.
\newblock Attentive semantic video generation using captions.
\newblock In {\em 2017 IEEE International Conference on Computer Vision
  (ICCV)}, pages 1435--1443, 2017.

\bibitem{meng2022distillation}
Chenlin Meng, Ruiqi Gao, Diederik~P. Kingma, Stefano Ermon, Jonathan Ho, and
  Tim Salimans.
\newblock On distillation of guided diffusion models.
\newblock {\em arXiv preprint arXiv:2210.03142}, 2022.

\bibitem{meng2022sdedit}
Chenlin Meng, Yutong He, Yang Song, Jiaming Song, Jiajun Wu, Jun-Yan Zhu, and
  Stefano Ermon.
\newblock {SDE}dit: Guided image synthesis and editing with stochastic
  differential equations.
\newblock In {\em International Conference on Learning Representations}, 2022.

\bibitem{Mescheder2018ICML}
Lars Mescheder, Sebastian Nowozin, and Andreas Geiger.
\newblock Which training methods for gans do actually converge?
\newblock In {\em International Conference on Machine Learning (ICML)}, 2018.

\bibitem{mittal2017sync}
Gaurav Mittal, Tanya Marwah, and Vineeth~N. Balasubramanian.
\newblock Sync-draw: Automatic video generation using deep recurrent attentive
  architectures.
\newblock In {\em Proceedings of the 25th ACM International Conference on
  Multimedia}, MM '17, page 1096–1104, New York, NY, USA, 2017. Association
  for Computing Machinery.

\bibitem{molad2023dreamix}
Eyal Molad, Eliahu Horwitz, Dani Valevski, Alex~Rav Acha, Yossi Matias, Yael
  Pritch, Yaniv Leviathan, and Yedid Hoshen.
\newblock Dreamix: Video diffusion models are general video editors.
\newblock {\em arXiv preprint arXiv:2302.01329}, 2023.

\bibitem{nichol2021glide}
Alex Nichol, Prafulla Dhariwal, Aditya Ramesh, Pranav Shyam, Pamela Mishkin,
  Bob McGrew, Ilya Sutskever, and Mark Chen.
\newblock Glide: Towards photorealistic image generation and editing with
  text-guided diffusion models.
\newblock {\em arXiv preprint arXiv:2112.10741}, 2021.

\bibitem{nichol2021improved}
Alexander~Quinn Nichol and Prafulla Dhariwal.
\newblock Improved denoising diffusion probabilistic models.
\newblock In {\em International Conference on Machine Learning}, 2021.

\bibitem{Pan2017ToCW}
Yingwei Pan, Zhaofan Qiu, Ting Yao, Houqiang Li, and Tao Mei.
\newblock To create what you tell: Generating videos from captions.
\newblock {\em Proceedings of the 25th ACM international conference on
  Multimedia}, 2017.

\bibitem{preechakul2021diffusion}
Konpat Preechakul, Nattanat Chatthee, Suttisak Wizadwongsa, and Supasorn
  Suwajanakorn.
\newblock Diffusion autoencoders: Toward a meaningful and decodable
  representation.
\newblock In {\em IEEE Conference on Computer Vision and Pattern Recognition
  (CVPR)}, 2022.

\bibitem{radford2021learning}
Alec Radford, Jong~Wook Kim, Chris Hallacy, Aditya Ramesh, Gabriel Goh,
  Sandhini Agarwal, Girish Sastry, Amanda Askell, Pamela Mishkin, Jack Clark,
  et~al.
\newblock Learning transferable visual models from natural language
  supervision.
\newblock In {\em International Conference on Machine Learning}, pages
  8748--8763. PMLR, 2021.

\bibitem{ramesh2022dalle2}
Aditya Ramesh, Prafulla Dhariwal, Alex Nichol, Casey Chu, and Mark Chen.
\newblock Hierarchical text-conditional image generation with clip latents.
\newblock {\em arXiv preprint arXiv:2204.06125}, 2022.

\bibitem{ramesh2021dalle}
Aditya Ramesh, Mikhail Pavlov, Gabriel Goh, Scott Gray, Chelsea Voss, Alec
  Radford, Mark Chen, and Ilya Sutskever.
\newblock Zero-shot text-to-image generation.
\newblock In Marina Meila and Tong Zhang, editors, {\em Proceedings of the 38th
  International Conference on Machine Learning}, volume 139 of {\em Proceedings
  of Machine Learning Research}, pages 8821--8831. PMLR, 18--24 Jul 2021.

\bibitem{rogozhnikov2022einops}
Alex Rogozhnikov.
\newblock Einops: Clear and reliable tensor manipulations with einstein-like
  notation.
\newblock In {\em International Conference on Learning Representations}, 2022.

\bibitem{rombach2021highresolution}
Robin Rombach, Andreas Blattmann, Dominik Lorenz, Patrick Esser, and Bj{\"o}rn
  Ommer.
\newblock High-resolution image synthesis with latent diffusion models.
\newblock {\em arXiv preprint arXiv:2112.10752}, 2021.

\bibitem{ruiz2022dreambooth}
Nataniel Ruiz, Yuanzhen Li, Varun Jampani, Yael Pritch, Michael Rubinstein, and
  Kfir Aberman.
\newblock Dreambooth: Fine tuning text-to-image dissusion models for
  subject-driven generation.
\newblock {\em arXiv preprint arXiv:2208.12242}, 2022.

\bibitem{saharia2021palette}
Chitwan Saharia, William Chan, Huiwen Chang, Chris~A. Lee, Jonathan Ho, Tim
  Salimans, David~J. Fleet, and Mohammad Norouzi.
\newblock Palette: Image-to-image diffusion models.
\newblock {\em arXiv preprint arXiv:2111.05826}, 2021.

\bibitem{saharia2022imagen}
Chitwan Saharia, William Chan, Saurabh Saxena, Lala Li, Jay Whang, Emily
  Denton, Seyed Kamyar~Seyed Ghasemipour, Burcu~Karagol Ayan, S.~Sara Mahdavi,
  Rapha~Gontijo Lopes, Tim Salimans, Jonathan Ho, David~J Fleet, and Mohammad
  Norouzi.
\newblock Photorealistic text-to-image diffusion models with deep language
  understanding.
\newblock {\em arXiv preprint arXiv:2205.11487}, 2022.

\bibitem{saharia2021image}
Chitwan Saharia, Jonathan Ho, William Chan, Tim Salimans, David~J Fleet, and
  Mohammad Norouzi.
\newblock Image super-resolution via iterative refinement.
\newblock {\em arXiv preprint arXiv:2104.07636}, 2021.

\bibitem{TGAN2017}
Masaki Saito, Eiichi Matsumoto, and Shunta Saito.
\newblock Temporal generative adversarial nets with singular value clipping.
\newblock In {\em ICCV}, 2017.

\bibitem{TGAN2020}
Masaki Saito, Shunta Saito, Masanori Koyama, and Sosuke Kobayashi.
\newblock Train sparsely, generate densely: Memory-efficient unsupervised
  training of high-resolution temporal gan.
\newblock {\em International Journal of Computer Vision}, May 2020.

\bibitem{salimans2016inceptionscore}
Tim Salimans, Ian Goodfellow, Wojciech Zaremba, Vicki Cheung, Alec Radford, Xi
  Chen, and Xi Chen.
\newblock Improved techniques for training gans.
\newblock In {\em Advances in Neural Information Processing Systems}, 2016.

\bibitem{salimans2022progressive}
Tim Salimans and Jonathan Ho.
\newblock Progressive distillation for fast sampling of diffusion models.
\newblock In {\em International Conference on Learning Representations (ICLR)},
  2022.

\bibitem{sasaki2021unitddpm}
Hiroshi Sasaki, Chris~G. Willcocks, and Toby~P. Breckon.
\newblock {UNIT-DDPM}: Unpaired image translation with denoising diffusion
  probabilistic models.
\newblock {\em arXiv preprint arXiv:2104.05358}, 2021.

\bibitem{sauer2021styleganxl}
Axel Sauer, Katja Schwarz, and Andreas Geiger.
\newblock Stylegan-xl: Scaling stylegan to large diverse datasets.
\newblock In {\em ACM SIGGRAPH 2022 Conference Proceedings}, pages 1--10, 2022.

\bibitem{singer2022make}
Uriel Singer, Adam Polyak, Thomas Hayes, Xi Yin, Jie An, Songyang Zhang, Qiyuan
  Hu, Harry Yang, Oron Ashual, Oran Gafni, et~al.
\newblock Make-a-video: Text-to-video generation without text-video data.
\newblock {\em arXiv:2209.14792}, 2022.

\bibitem{sinha2021d2c}
Abhishek Sinha, Jiaming Song, Chenlin Meng, and Stefano Ermon.
\newblock D2c: Diffusion-denoising models for few-shot conditional generation.
\newblock In {\em Advances in Neural Information Processing Systems}, 2021.

\bibitem{Skorokhodov_2022_CVPR}
Ivan Skorokhodov, Sergey Tulyakov, and Mohamed Elhoseiny.
\newblock Stylegan-v: A continuous video generator with the price, image
  quality and perks of stylegan2.
\newblock In {\em Proceedings of the IEEE/CVF Conference on Computer Vision and
  Pattern Recognition (CVPR)}, pages 3626--3636, June 2022.

\bibitem{sohl2015deep}
Jascha Sohl-Dickstein, Eric Weiss, Niru Maheswaranathan, and Surya Ganguli.
\newblock Deep unsupervised learning using nonequilibrium thermodynamics.
\newblock In {\em International Conference on Machine Learning}, 2015.

\bibitem{song2021denoising}
Jiaming Song, Chenlin Meng, and Stefano Ermon.
\newblock Denoising diffusion implicit models.
\newblock In {\em International Conference on Learning Representations}, 2021.

\bibitem{song2019generative}
Yang Song and Stefano Ermon.
\newblock Generative modeling by estimating gradients of the data distribution.
\newblock In {\em Proceedings of the 33rd Annual Conference on Neural
  Information Processing Systems}, 2019.

\bibitem{song2020score}
Yang Song, Jascha Sohl-Dickstein, Diederik~P Kingma, Abhishek Kumar, Stefano
  Ermon, and Ben Poole.
\newblock Score-based generative modeling through stochastic differential
  equations.
\newblock In {\em International Conference on Learning Representations}, 2021.

\bibitem{soomro2012ucf101}
Khurram Soomro, Amir~Roshan Zamir, and Mubarak Shah.
\newblock Ucf101: A dataset of 101 human actions classes from videos in the
  wild.
\newblock {\em arXiv preprint arXiv:1212.0402}, 2012.

\bibitem{su2022dual}
Xuan Su, Jiaming Song, Chenlin Meng, and Stefano Ermon.
\newblock Dual diffusion implicit bridges for image-to-image translation.
\newblock {\em arXiv preprint arXiv:2203.08382}, 2022.

\bibitem{tian2021a}
Yu Tian, Jian Ren, Menglei Chai, Kyle Olszewski, Xi Peng, Dimitris~N. Metaxas,
  and Sergey Tulyakov.
\newblock A good image generator is what you need for high-resolution video
  synthesis.
\newblock In {\em International Conference on Learning Representations}, 2021.

\bibitem{tran2015c3d}
D. Tran, L. Bourdev, R. Fergus, L. Torresani, and M. Paluri.
\newblock Learning spatiotemporal features with 3d convolutional networks.
\newblock In {\em 2015 IEEE International Conference on Computer Vision
  (ICCV)}, 2015.

\bibitem{unterthiner2018towards}
Thomas Unterthiner, Sjoerd van Steenkiste, Karol Kurach, Raphael Marinier,
  Marcin Michalski, and Sylvain Gelly.
\newblock Towards accurate generative models of video: A new metric \&
  challenges.
\newblock {\em arXiv:1812.01717}, 2018.

\bibitem{vahdat2021score}
Arash Vahdat, Karsten Kreis, and Jan Kautz.
\newblock Score-based generative modeling in latent space.
\newblock In {\em Advances in Neural Information Processing Systems}, 2021.

\bibitem{vaswani2017attention}
Ashish Vaswani, Noam Shazeer, Niki Parmar, Jakob Uszkoreit, Llion Jones,
  Aidan~N Gomez, \L~ukasz Kaiser, and Illia Polosukhin.
\newblock Attention is all you need.
\newblock In {\em Advances in Neural Information Processing Systems}, 2017.

\bibitem{villegas2022phenaki}
Ruben Villegas, Mohammad Babaeizadeh, Pieter-Jan Kindermans, Hernan Moraldo,
  Han Zhang, Mohammad~Taghi Saffar, Santiago Castro, Julius Kunze, and Dumitru
  Erhan.
\newblock Phenaki: Variable length video generation from open domain textual
  description.
\newblock {\em arXiv:2210.02399}, 2022.

\bibitem{villegas17mcnet}
Ruben Villegas, Jimei Yang, Seunghoon Hong, Xunyu Lin, and Honglak Lee.
\newblock Decomposing motion and content for natural video sequence prediction.
\newblock {\em ICLR}, 2017.

\bibitem{vincent2011}
Pascal Vincent.
\newblock A connection between score matching and denoising autoencoders.
\newblock {\em Neural Computation}, 23(7):1661--1674, 2011.

\bibitem{voleti2022mcvd}
Vikram Voleti, Alexia Jolicoeur-Martineau, and Christopher Pal.
\newblock Mcvd: Masked conditional video diffusion for prediction, generation,
  and interpolation.
\newblock {\em arXiv preprint arXiv:2205.09853}, 2022.

\bibitem{scene_dyn}
Carl Vondrick, Hamed Pirsiavash, and Antonio Torralba.
\newblock Generating videos with scene dynamics.
\newblock In {\em Proceedings of the 30th International Conference on Neural
  Information Processing Systems}, 2016.

\bibitem{Wang_2020_CVPR}
Yaohui Wang, Piotr Bilinski, Francois Bremond, and Antitza Dantcheva.
\newblock G3an: Disentangling appearance and motion for video generation.
\newblock In {\em IEEE/CVF Conference on Computer Vision and Pattern
  Recognition (CVPR)}, 2020.

\bibitem{watson2022learning}
Daniel Watson, William Chan, Jonathan Ho, and Mohammad Norouzi.
\newblock Learning fast samplers for diffusion models by differentiating
  through sample quality.
\newblock In {\em International Conference on Learning Representations}, 2022.

\bibitem{Weissenborn2020Scaling}
Dirk Weissenborn, Oscar Täckström, and Jakob Uszkoreit.
\newblock Scaling autoregressive video models.
\newblock In {\em International Conference on Learning Representations}, 2020.

\bibitem{wu2021godiva}
Chenfei Wu, Lun Huang, Qianxi Zhang, Binyang Li, Lei Ji, Fan Yang, Guillermo
  Sapiro, and Nan Duan.
\newblock Godiva: Generating open-domain videos from natural descriptions.
\newblock {\em arXiv:2104.14806}, 2021.

\bibitem{wu2022nuwa}
Chenfei Wu, Jian Liang, Lei Ji, Fan Yang, Yuejian Fang, Daxin Jiang, and Nan
  Duan.
\newblock N{\"u}wa: Visual synthesis pre-training for neural visual world
  creation.
\newblock In {\em European Conference on Computer Vision}, pages 720--736.
  Springer, 2022.

\bibitem{xiao2022DDGAN}
Zhisheng Xiao, Karsten Kreis, and Arash Vahdat.
\newblock Tackling the generative learning trilemma with denoising diffusion
  {GAN}s.
\newblock In {\em International Conference on Learning Representations (ICLR)},
  2022.

\bibitem{xu2016msr-vtt}
Jun Xu, Tao Mei, Ting Yao, and Yong Rui.
\newblock Msr-vtt: A large video description dataset for bridging video and
  language.
\newblock In {\em International Conference on Computer Vision and Pattern
  Recognition (CVPR)}, 2016.

\bibitem{xue2022hdvila}
Hongwei Xue, Tiankai Hang, Yanhong Zeng, Yuchong Sun, Bei Liu, Huan Yang,
  Jianlong Fu, and Baining Guo.
\newblock Advancing high-resolution video-language representation with
  large-scale video transcriptions.
\newblock In {\em International Conference on Computer Vision and Pattern
  Recognition (CVPR)}, 2022.

\bibitem{yan2021videogpt}
Wilson Yan, Yunzhi Zhang, Pieter Abbeel, and Aravind Srinivas.
\newblock Videogpt: Video generation using vq-vae and transformers, 2021.

\bibitem{yang2022video}
Ruihan Yang, Prakhar Srivastava, and Stephan Mandt.
\newblock Diffusion probabilistic modeling for video generation.
\newblock {\em arXiv preprint arXiv:2203.09481}, 2022.

\bibitem{yu2022parti}
Jiahui Yu, Yuanzhong Xu, Jing~Yu Koh, Thang Luong, Gunjan Baid, Zirui Wang,
  Vijay Vasudevan, Alexander Ku, Yinfei Yang, Burcu~Karagol Ayan, Ben
  Hutchinson, Wei Han, Zarana Parekh, Xin Li, Han Zhang, Jason Baldridge, and
  Yonghui Wu.
\newblock Scaling autoregressive models for content-rich text-to-image
  generation.
\newblock {\em arXiv preprint arXiv:2206.10789}, 2022.

\bibitem{yu2022generating}
Sihyun Yu, Jihoon Tack, Sangwoo Mo, Hyunsu Kim, Junho Kim, Jung-Woo Ha, and
  Jinwoo Shin.
\newblock Generating videos with dynamics-aware implicit generative adversarial
  networks.
\newblock In {\em International Conference on Learning Representations}, 2022.

\bibitem{zeng2022lion}
Xiaohui Zeng, Arash Vahdat, Francis Williams, Zan Gojcic, Or Litany, Sanja
  Fidler, and Karsten Kreis.
\newblock Lion: Latent point diffusion models for 3d shape generation.
\newblock In {\em Advances in Neural Information Processing Systems (NeurIPS)},
  2022.

\bibitem{zhang2022Fast}
Qinsheng Zhang and Yongxin Chen.
\newblock Fast sampling of diffusion models with exponential integrator.
\newblock {\em arXiv:2204.13902}, 2022.

\bibitem{zhou2022magicvideo}
Daquan Zhou, Weimin Wang, Hanshu Yan, Weiwei Lv, Yizhe Zhu, and Jiashi Feng.
\newblock Magicvideo: Efficient video generation with latent diffusion models.
\newblock {\em arXiv preprint arXiv:2211.11018}, 2022.

\end{thebibliography}
}

\newpage
\appendix
\onecolumn

\clearpage
\begingroup
\hypersetup{linkcolor=black}
\tableofcontents
\endgroup
\clearpage

\section{Linked Videos} \label{app:supp_videos}
For fully rendered videos, we primarily refer the reader to our project page, \url{https://research.nvidia.com/labs/toronto-ai/VideoLDM/}.

Moreover, we include 5 videos in the following google drive folder: \url{https://drive.google.com/drive/folders/1ENd9_9lzN6mI3E_HAjP52_KMWFdd0c8u}:
\begin{itemize}
    \item \texttt{driving.mp4}: This video summarizes different results for our Video LDM trained on the real-world driving data.
    \item \texttt{text2video.mp4}: This video shows different results for our text-to-video LDM based on Stable Diffusion.
    \item \texttt{mountain$\_$biking.mp4}: This video presents further results on an additional mountain biking video dataset.
    \item \texttt{5$\_$minutes$\_$driving.mp4}: This video shows full 5 minute long generated videos for the Video LDM trained on the real-world driving data.
    \item \texttt{5$\_$minutes$\_$biking.mp4}: This video shows full 5 minute long generated videos for the Video LDM trained on the mountain biking data.
\end{itemize}
We also keep a copy of the latest version of the paper in the google drive folder, in case the paper is updated at a later point in time.
\section{Broader Impact and Limitations} \label{app:impact}
Powerful video generative models, like our Video LDM, have the potential to enable important content creation applications in the future and streamline and improve the creative workflow of digital artists, thereby democratizing artistic expression. Moreover, when applied for instance on videos captured from vehicles, as in our main validation experiments, generative models like ours may also serve as simulators in autonomous driving research.

Our synthesized videos are not indistinguishable from real content yet. However, enhanced versions of our model may in the future reach an even higher quality, potentially being able to generate videos that appear to be deceptively real. This has important ethical and safety implications, as state-of-the-art deep generative models can also be used for malicious purposes, and therefore generative models like ours generally need to be applied with an abundance of caution. Moreover, the data sources cited in this paper are for research purposes only and not intended for commercial application or use, and the text-to-image backbone LDMs used in this research project have been trained on large amounts of internet data. Consequently, our model is not suitable for productization.
An important direction for future work is training large-scale generative models with ethically sourced, commercially viable data.

\section{Related Work}\label{sec:related_extended}
Here, we present an extended discussion about related work.

\textbf{Diffusion Models for Image Synthesis.}
Diffusion models (DMs)~\cite{sohl2015deep,ho2020ddpm,song2020score} have proven to be powerful image generators, yielding state-of-the art results in both unconditional and class-conditional synthesis~\cite{nichol2021improved,rombach2021highresolution,dhariwal2021diffusion} as well as text-to-image generation~\cite{nichol2021glide,rombach2021highresolution,ramesh2022dalle2,saharia2022imagen,balaji2022eDiffi}. They have also been successfully used for various image editing and processing tasks~\cite{meng2022sdedit,lugmayr2022repaint,saharia2021image,li2022srdiff,sasaki2021unitddpm,saharia2021palette,su2022dual,kawar2022restoration,hertz2022prompt,ruiz2022dreambooth,gal2022animage}.

However, despite advances in model distillation~\cite{salimans2022progressive,luhman2021knowledge,meng2022distillation} and accelerated sampling~\cite{song2021denoising,jolicoeur2021gotta,dockhorn2022score,liu2022pseudo,xiao2022DDGAN,zhang2022Fast,lu2022dpm,dockhorn2022genie,watson2022learning,bao2022analyticdpm}, DMs generally require repeated evaluations of a computationally demanding large U-Net. 
Thus, DMs are computationally expensive during both training and inference, especially when applied at high resolutions. 
To address this, \emph{cascaded}~\cite{ho2021cascaded} and \emph{latent}~\cite{vahdat2021score,rombach2021highresolution} diffusion models have been introduced. Both approaches divide the synthesis (and training) process into multiple stages and move the resource-intensive training and evaluation to a space of lower computational complexity. Cascaded diffusion models start out as low-resolution models and apply a series of super resolution diffusion models. 
Latent space models first compress the image data using an autoencoder and learn the DMs on the resulting latent space. 
We combine the best of these approaches for video synthesis. 
Our main video generator is a latent diffusion model. Additionally, some of our models use a video upsampler like in cascaded models to further increase the resolution.
Variations of latent diffusion models have also been used for tasks such as controllable and semantic image generation~\cite{sinha2021d2c,preechakul2021diffusion}, and beyond image synthesis, such as 3D shape synthesis~\cite{zeng2022lion}.

\textbf{Video Synthesis.} Video generation has been tackled with recurrent neural networks~\cite{babaeizadeh2018stochastic,svg,lee2018savp,hvrnn,lsvg}, autoregressive transformers~\cite{Weissenborn2020Scaling,yan2021videogpt,hong2022cogvideo,wu2021godiva,wu2022nuwa,ge2022longvideo,Gupta_2022_CVPR}, Normalizing Flows~\cite{si2v,ipoke}, and generative adversarial networks (GANs)~\cite{scene_dyn,yu2022generating,tian2021a, villegas17mcnet,Luc2020TransformationbasedAV,TGAN2020,brooks2022generating,Skorokhodov_2022_CVPR,kahembwe2020lower,TGAN2017,Wang_2020_CVPR,fox2021stylevideogan}. In particular LongVideoGAN~\cite{brooks2022generating} achieves high-resolution video synthesis over relatively long time intervals, combining a low- and a high-resolution model. Moreover, the idea to insert temporal layers into pre-trained generators has been explored by the GAN-based methods MoCoGAN-HD~\cite{tian2021a} and StyleVideoGAN~\cite{fox2021stylevideogan} before, but at a much smaller scale for simple object-centric videos.
Another important work is CogVideo~\cite{hong2022cogvideo}, which video fine-tunes a text-to-image model. 
However, it is a strictly autoregressive architecture building on transformers, trained in a discrete latent space. Our method, in contrast, relies on continuous DMs, is much less parameter intensive and outperforms CogVideo in text-to-video synthesis. Furthermore, our alignment layers are also implemented differently. They do not require an autoregressive masking and, for our text-to-video experiments, condition on the text prompts.
Finally, we also show driving scene generation, where we employ an additional video fine-tuned upsampler module. Many more text-to-video models exist~\cite{mittal2017sync,Pan2017ToCW,marwah2017attentive,li2017video,gupta2018imagine,wu2021godiva,wu2022nuwa,ho2022video}.

Most related to our work are previous DMs for video synthesis:
Ho et al.~\cite{ho2022video} model low-resolution videos with DMs in pixel space and train jointly on image and video data. 
Yang et al.~\cite{yang2022video} parameterize autoregressive video generation using a deterministic frame predictor together with a probabilistic DM. Voleti et al.~\cite{voleti2022mcvd} introduce a general DM framework that simultaneously enables video generation, prediction, and interpolation, similarly to H\"oppe et al.~\cite{hoeppe2022diffusion}. Closely related, Harvey et al.~\cite{harvey2022flexible} synthesize long videos by generating sparse frames first, and then adding the missing frames. However, they model only low-resolution videos from small simulated toy worlds, to which the DM easily overfits, whereas we are training exclusively on diverse high-resolution real-world data. In our Video LDM, we build on these works for our video prediction and interpolation frameworks. However, in contrast to our Video LDM, all previous video DMs work directly in pixel space, synthesize low-resolution videos only, and do not directly leverage pre-trained image DMs.

\textbf{Concurrent Work.} Concurrently with us, Make-A-Video~\cite{singer2022make} leveraged a DALL$\cdot$E 2-like text-to-image DM~\cite{ramesh2022dalle2} and temporally aligned its decoder as well as one super resolution DM for consistent video generation. Furthermore, Imagen Video~\cite{ho2022imagenvideo} trained a large-scale cascaded text-to-video model building on the Imagen~\cite{saharia2022imagen} architecture. It constructs a highly demanding spatial and temporal super resolution pipeline consisting of in total 7 DMs with a total of ${>}11$B parameters (also see model parameters comparisons in \Cref{app:model_params}).
In contrast to our Video LDM, both Imagen Video and Make-A-Video operate entirely in pixel space, which is less efficient, and exclusively target text-to-video, whereas we also demonstrate high-resolution driving scene generation. 
Furthermore, neither the large-scale DALL$\cdot$E 2 and Imagen text-to-image models, nor the corresponding text-to-video models are easily reproducible without substantial GPU resources. In contrast, we are building on \emph{Stable Diffusion} to train our text-to-video model, and construct an overall significantly more efficient pipeline. This makes our approach more user-friendly. 
Phenaki~\cite{villegas2022phenaki} is another strong concurrent text-to-video model that compresses videos into discrete tokens and models the token distribution via bidirectional transformers. MagicVideo~\cite{zhou2022magicvideo} is a concurrent method that also leverages latent diffusion models for video generation and follows a partially related strategy compared to our Video LDM. We are outperforming MagicVideo quantitatively (see \Cref{sec:text-to-video}), enable higher resolution generation, show personalized video generation, and also demonstrate long driving scene and mountain bike video generation; in contrast, MagicVideo tackles text-to-video only. Other relevant concurrent works include GEN-1~\cite{esser2023structure}, which also leverages a latent diffusion framework and combines it with depth conditioning for structure- and content-aware video editing and stylization, as well as Dreamix~\cite{molad2023dreamix}, a method for video editing that demonstrates personalized video generation.

\section{Using Video LDM ``Convolutional in Time'' and ``Convolutional in Space''} \label{app:convolutional_ldm}
An intriguing property of image LDMs is their ability to generalize to spatial resolutions much larger than the ones they are trained on. This is realized by increasing the spatial size of the sampled noise and leveraging the convolutional nature of the U-Net backbone as presented in~\cite{rombach2021highresolution}. Since we use the Stable Diffusion LDM as fixed generative image backbone for our text-to-video model, our approach naturally preserves this property, which enables us to increase the spatial resolution at inference time without significant loss of image quality. We show examples at spatial resolution $512\times512$ generated by applying this convolutional sampling "in space" in our videos and in \Cref{fig:space_conv1,fig:space_conv2}. Note that our model was trained on resolution $320\times512$. This convolutional application of our model for spatially extended generation essentially comes for free. We are not using the additional upsampler diffusion model in those experiments.

Furthermore, to extend convolutional sampling to the temporal dimension and, thus, to be able to generate videos much longer than those our model has been trained on, we make the following design choices regarding our temporal layers. First, we use relative sinusoidal positional encodings for our temporal attention layers similar to those used to encode the timesteps in our U-Net backbone~\cite{ho2020ddpm}. Second, we parameterize the learned mixing factors $\alpha_{\phi}^{i}$, \cf~ \Cref{sec:turning}, with scalars for our text-to-video model (in the other models, $\alpha_{\phi}^{i}$ varies along the temporal dimension, which would prevent convolutional-in-time processing. $\alpha_{\phi}^{i}$ is always constant in the spatial and channel dimensions for all models). These choices ensure that our model can generate longer sequences by simply increasing the number of frames for the model to render. Note that when applying our model convolutionally in time, we mask the temporal attention layers such that we only attend over a maximum frame distance of 8 frames, similar to training.

We find that our Video LDM generalizes to longer sequences as shown in~\Cref{fig:time_conv1}, although some degradation in quality can be observed. Furthermore, we can combine convolutional sampling in space and time leading to high-resolution videos of lengths up to 30 seconds as presented in~\Cref{fig:longteddy,fig:hotdog} and in our videos, although our model has been trained only on sequences of 4 seconds. That said, we find that convolutional-in-time generation can be fragile, in particular when targeting long videos. Hence, we advocate training of prediction models for more robust long-term generation, as we do in our experiments on real driving scene synthesis. Also synthesizing long text-to-video samples with prediction models is left to future work.

Nevertheless, to the best of our knowledge, our approach is among the first to simultaneously generate long, high resolution \emph{and} high (up to 30) fps videos while keeping training cost tolerable. 
\section{Datasets}

\subsection{Real Driving Scene (RDS) Video} \label{sec:rds_details}
Our RDS dataset consists of 683,060 real driving videos of 8 seconds length each at resolution $512\times1024$ ($H\times W$).
85,841 of these video have a frame rate of 30 fps, the rest 10 fps. Furthermore, all videos have binary night/day labels (1 for night, 0 for day) and annotations for the number of cars in a scene (``crowdedness''). Note that most of the driving videos are relatively empty highway scenes with low crowdedness.

Moreover, the data comes with an additional 100k, independent, frames that have car bounding box annotations (we only used that data for training a bounding box-conditioned image LDM to initialize video synthesis in the qualitative experiments in \Cref{sec:driving_simulation} and \Cref{fig:av_samples}). 

\subsection{WebVid-10M}
When turning Stable Diffusion into a text-to-video generator, we rely on WebVid-10M~\cite{bain21frozen}. WebVid-10M is a large-scale dataset of short videos with textual descriptions sourced from stock footage sites. The videos are diverse and rich in their content. There are 10.7M video-caption pairs with a total of 52k video hours.

\subsection{Mountain Bike} \label{app:mountain_bike_dataset}
In \Cref{app:mountain_bike}, we report additional results on a first-person mountain biking video dataset, originally introduced by Brooks et al. in Long Video GAN~\cite{brooks2022generating}. The dataset consists of 1,202 clips of varying, but at least 5 seconds length. The videos have a frame rate of 30 fps. The dataset is available in different resolutions, with a maximum resolution of $576\times1024$. More details about the dataset can be found in Brooks et al.~\cite{brooks2022generating} and at \url{https://github.com/NVlabs/long-video-gan}.
\section{Architecture, Training and Sampling Details}\label{app:arch}
Our Image LDMs are based on Rombach et al.~\cite{rombach2021highresolution}. They use convolutional encoders and decoders, and their latent space DM architecture builds on the U-Net by Dhariwal et al.\cite{dhariwal2021diffusion}. Our pixel-space upsampler DMs use the same Image DM backbone~\cite{dhariwal2021diffusion}. We generally work with discretized diffusion time steps $t\in\{0,1000\}$, and use a linear noise schedule, following the formulation of Ho et al.~\cite{ho2020ddpm}. Also see Appendix B of Rombach et al.~\cite{rombach2021highresolution}, which recapitulates the diffusion process setup as it is used in LDMs.

All architecture details, diffusion process details, as well as training hyperparameters are provided in \Cref{table:hyperparameters_gen,table:hyperparameters_ablation,table:hyperparameters_ae}.

To sample from our models, we generally use the sampler from \textit{Denoising Diffusion Implicit Models} (DDIM)~\cite{song2021denoising}. The number of sampling steps, the stochasticity $\eta$, and the guidance scale vary and are also shown in \Cref{table:hyperparameters_gen,table:hyperparameters_ablation}.

\begin{table*}
\setlength{\tabcolsep}{4pt}
\caption{Hyperparameters for our diffusion models. ${}^\dag$: For the Text-to-Video LDMs, we do not train our own Image LDM, but use Stable Diffusion, as discussed. \textit{Training} details correspond to Stable Diffusion 2.1-based Video LDM that was also used in quantitative evaluations in Tabs.~\ref{tab:ucf} and~\ref{tab:msrvtt}. ${}^*$: We trained with 80 $\times$ 80 patches and run at full image resolution during inference (see \Cref{app:exp_details_text_to_video}).}
\centering
\resizebox*{1.0\textwidth}{!}{
    \begin{tabular}{lcccccccc}
    \toprule
    \textbf{Hyperparameter} & \multicolumn{2}{c}{\textbf{Driving}} & \textbf{Driving} & \multicolumn{2}{c}{\textbf{Text-to-Video}} &  \multicolumn{1}{c}{\textbf{Text-to-Video}} & \multicolumn{2}{c}{\textbf{Mountain Biking}}  \\
      & \multicolumn{2}{c}{\textbf{(Video) LDM}} & \textbf{(Video) Upsampler} & \multicolumn{2}{c}{\textbf{(Video) LDM} ${}^\dag$}  & \multicolumn{1}{c}{\textbf{(Video) LDM Upsampler} ${}^\dag$} & \multicolumn{2}{c}{\textbf{(Video) LDM}} \\
    \midrule\midrule
    \textbf{Image (L)DM}  &  &  & \\
    \textit{Architecture} \\
    LDM & \multicolumn{2}{c}{\cmark} &\xmark& \multicolumn{2}{c}{\cmark}& \multicolumn{1}{c}{\cmark}&  \multicolumn{2}{c}{\cmark}\\
    $f$ & \multicolumn{2}{c}{8} & - & \multicolumn{2}{c}{8} &  \multicolumn{1}{c}{4} & \multicolumn{2}{c}{8} \\
    $z$-shape & \multicolumn{2}{c}{$16 \times 32 \times 4$} & - & \multicolumn{2}{c}{$40 \times 64 \times 4$} &  \multicolumn{1}{c}{$80 \times 80 \times 4$ ${}^*$}  &  \multicolumn{2}{c}{$16 \times 32 \times 4$} \\
    Channels & \multicolumn{2}{c}{224} & 160 & \multicolumn{2}{c}{320} &   \multicolumn{1}{c}{320} & \multicolumn{2}{c}{256}\\
    Depth & \multicolumn{2}{c}{4} & 2 & \multicolumn{2}{c}{2}& \multicolumn{1}{c}{2}&\multicolumn{2}{c}{2} \\
    Channel multiplier & \multicolumn{2}{c}{1,2,2,3,4} & {1,2,4,4} & \multicolumn{2}{c}{1,2,4,4} &  \multicolumn{1}{c}{1,2,4,4} & \multicolumn{2}{c}{1,2,4}\\
    Attention resolutions & \multicolumn{2}{c}{16,8,4} & {8} & \multicolumn{2}{c}{64,32,16}  & \multicolumn{1}{c}{128,64,32}   & \multicolumn{2}{c}{32,16,8}\\
    Head channels & \multicolumn{2}{c}{32} & - & \multicolumn{2}{c}{-} &  \multicolumn{1}{c}{-} & \multicolumn{2}{c}{-} \\
    Number of heads & \multicolumn{2}{c}{-} & 4 & \multicolumn{2}{c}{8}&  \multicolumn{1}{c}{8} & \multicolumn{2}{c}{4} \\
    \midrule
    \emph{CA Conditioning} & & &  \\
    Embedding dimension & \multicolumn{2}{c}{256} & 768 & \multicolumn{2}{c}{768} & \multicolumn{1}{c}{768} & \multicolumn{2}{c}{-}\\
    CA resolutions & \multicolumn{2}{c}{16,8,4} & 8 & \multicolumn{2}{c}{64,32,16} &   \multicolumn{1}{c}{128,64,32} & \multicolumn{2}{c}{-} \\
    CA sequence length & \multicolumn{2}{c}{1} & 1 & \multicolumn{2}{c}{77} &  \multicolumn{1}{c}{77} & \multicolumn{2}{c}{-} \\
    \midrule
    \textit{Training}  &  &  &\\
    Parameterization & \multicolumn{2}{c}{$\rvv$} & $\rvv$ & \multicolumn{2}{c}{$\boldsymbol{\varepsilon}$} &  \multicolumn{1}{c}{$\boldsymbol{\varepsilon}$} &  \multicolumn{2}{c}{$\rvv$}\\
    \# train steps  & \multicolumn{2}{c}{73K}  & 84K & \multicolumn{2}{c}{-} &  \multicolumn{1}{c}{-} & \multicolumn{2}{c}{14K}\\
    Learning rate & \multicolumn{2}{c}{$10^{-4}$} & $10^{-5}$ &\multicolumn{2}{c}{-} &  \multicolumn{1}{c}{-} & \multicolumn{2}{c}{$10^{-4}$}\\
    Batch size per GPU & \multicolumn{2}{c}{40} & 4 &\multicolumn{2}{c}{-} & \multicolumn{1}{c}{-} & \multicolumn{2}{c}{300} \\
    \# GPUs  & \multicolumn{2}{c}{16} & 8 & \multicolumn{2}{c}{-} & \multicolumn{1}{c}{-}   & \multicolumn{2}{c}{1}\\
    GPU-type &\multicolumn{2}{c}{A100-40GB} & A100-80GB & \multicolumn{2}{c}{-} & \multicolumn{1}{c}{-} & \multicolumn{2}{c}{A100-80GB}\\
    $p_{\text{\footnotesize{drop}}}$  & \multicolumn{2}{c}{0.1} & - & \multicolumn{2}{c}{-} & \multicolumn{1}{c}{-} & \multicolumn{2}{c}{-}\\
    \midrule\midrule
    \textbf{Temporal Layers}  &  &  &\\
     & \emph{prediction} & \emph{interpolation} & \emph{upsampler} & \emph{keyframes model} & \emph{interpolation} & \emph{upsampler} & \emph{prediction} & \emph{interpolation} \\
    \midrule
    \textit{Architecture}  &  &  &\\
    Depth & 4 &4 & 2 & 2& 2  & 2& 2 & 2  \\
    Attention resolutions &16,8,4 &16,8,4 &  1, 2, 4, 4  & 64,32,16& 64,32,16 & 128,64,32 & 32,16,8 & 32,16,8\\
    Head channels & 32 &32 &  - & - & - & - & -  & -\\
    Number of heads & - & - & 4 & 8& 8 & 8 & 4 & 4 \\
    Input projection & \cmark &\cmark & \cmark & \cmark& \cmark & \cmark& \cmark  & \cmark\\
    Positional encoding & Sinusoidal& Sinusoidal & - & Sinusoidal& Sinusoidal & Sinusoidal& Sinusoidal& Sinusoidal \\
    $\dim\alpha$ & 10 &5 & 10 & 1& 1 & 1 & 8& 5 \\
    Temporal kernel size & 3,1,1& 3,1,1 & 3,1,1 & 3,1,1& 3,1,1 & 3,1,1 & 3,1,1& 3,1,1\\
    \midrule
    \emph{Concat Conditioning} & & &  \\
    $\dim c_{S}$  & 5 &5 & 3 & - & -   &3 & 5 &5 \\
    Context channels & 128& 128 &128 & - & -  & 128 & 128& 128  \\
    \midrule
    \emph{CA Conditioning} & & & & \\
    Embedding dimension & - & 512 & 768 & 768 & 768  & 768 & - & 512 \\
    CA resolutions &- & 16,4,8 & 8 & 64,32,16 & 64,32,16  & 64,32,16 & - & 32,16,8 \\
    CA sequence length &- & 1& 1 & 77 & 2  & 77 & - & 1 \\
    \midrule
    \emph{Training}  &  &  & \\
    \# train steps  & 146K  & 42K  & 84K & 402K & 95K & 10K & 107K & 33K  \\
    Learning rate & $10^{-4}$ & $10^{-4}$ & $10^{-4}$ & $10^{-4}$ & $10^{-4}$ & $10^{-4}$ & $10^{-4}$ & $10^{-4}$  \\
    Batch size per GPU  & 9 & 16 & 2 & 3 & 8  & 8 & 20 & 20 \\
    \# GPUs  & 40 & 24 & 8 & 256 & 128  & 32 & 8 & 6  \\
    GPU-type  & A100-40GB & A100-40GB &A100-80GB & A100-80GB & A100-80GB & A100-80GB   & A100-40GB & A100-40GB \\
    Sequence length & 10 & 5 & 8 & 8 & 5  & 8 & 8 & 5  \\
    Training data FPS & 2 & 7.5, 30 & 4 & 2 & 7.5, 30  & 2   & 4 & 7.5, 30\\
    \midrule\midrule
    \textbf{Diffusion Setup} \\
    Diffusion steps & \multicolumn{2}{c}{1000} & 1000 & \multicolumn{2}{c}{1000} & \multicolumn{1}{c}{1000} & \multicolumn{2}{c}{1000} \\
    Noise schedule & \multicolumn{2}{c}{Linear} & Linear & \multicolumn{2}{c}{Linear} & \multicolumn{1}{c}{Linear} & \multicolumn{2}{c}{Linear} \\
    $\beta_{0}$ & \multicolumn{2}{c}{0.0015} & 1e-4 & \multicolumn{2}{c}{0.00085} & \multicolumn{1}{c}{0.00085} & \multicolumn{2}{c}{0.0003}  \\
    $\beta_{T}$ & \multicolumn{2}{c}{0.0195} & 2e-2 & \multicolumn{2}{c}{0.0120} & \multicolumn{1}{c}{0.0120} & \multicolumn{2}{c}{0.022}  \\
    \midrule\midrule
    \textbf{Sampling Parameters}  &  &  & \\
    Sampler & DDIM & DDIM  & DDIM  & DDIM  & DDIM  & DDIM  & DDIM & DDIM \\
    Steps & 200 & 50 & 50 & 250 & 100 & 250 & 100 & 50 \\
    $\eta$ & 1.0 & 0.0 & 1.0 & 1.0 & 1.0 & 1.0 & 1.0 & 1.0\\
    Guidance scale & 2.0 & 2.0 & 1.0 & 8.0 & 13.0 & 2.0 & 2.0 & 1.0\\
    \bottomrule
\end{tabular}%
}
\label{table:hyperparameters_gen}
\vspace{-0.4cm}
\end{table*}

\begin{table*}
\setlength{\tabcolsep}{4pt}
\caption{Hyperparameters for our all models presented in the ablation study in~\Cref{tab:av_main}. $^*$: This baseline has twice as many attention layers as the other models, leading to the same number of trainable parameters.}
\centering
\resizebox*{!}{.91\textheight}{
    \begin{tabular}{lcccc}
    \toprule
    \textbf{Hyperparameter} & \textbf{\emph{Ours}} & \textbf{Pixel-baseline} & \textbf{End-to-End LDM} & \textbf{Attention-only}$^*$  \\
       &  &  & &\\
    \midrule\midrule
    \textbf{Image (L)DM}  &  &  & &\\
    \textit{Architecture} \\
    LDM & \cmark & \xmark & \cmark & \cmark \\
    Pretrained &\cmark & \cmark & \xmark & \cmark \\
    $f$ & 8 & - & 8 & 8 \\
    $z$-shape & $16 \times 32 \times 4$ & $128 \times 256 \times 3$ & $16 \times 32 \times 4$ & $16 \times 32 \times 4$ \\
    Channels & 224 & 128 & 224 & 224 \\
    Depth & 4 & 4 & 4 & 4 \\
    Channel multiplier & 1,2,2,3,4 & 1,2,2,3,4 & 1,2,2,3,4 & 1,2,2,3,4\\
    Attention resolutions & 8,4,2 & 16,8 & 8,4,2 & 8,4,2\\
    Head channels & 32 & 32 & 32 & 32 \\
    \midrule
    \emph{CA Conditioning} & & &  \\
    embedding dimension & 256 & 256 & 256 & 256 \\
    CA resolutions & 16,8,4 & 32,16 & 16,8,4& 16,8,4\\
    CA sequence length & 1 & 1 & 1& 1 \\
    \midrule
    \textit{Training}  &  &  &\\
    Parameterization & $\rvv$ & $\rvv$ & $\rvv$ & $\rvv$\\
    \# train steps  & 73K  &  42K & 73K & 73K\\
    Learning rate & $10^{-4}$ & $10^{-4}$ & $10^{-4}$ & $10^{-4}$\\
    Batch size per GPU & 40 & 12 & 40 & 40 \\
    \# GPUs  & 16 & 40 & 16 & 16 \\
    GPU-type  & A100-40GB & A100-40GB & A100-40GB & A100-40GB \\
    $p_{\text{\footnotesize{drop}}}$  & 0.1  & 0.1  & 0.1  & 0.1 \\
    \midrule\midrule
    \textbf{Temporal Layers}  &  &  &\\
    \textit{Architecture}  &  &  &\\
    Depth & 4 & 4 & 4 & 4 \\
    Attention resolutions & 1,2,2,3,4 & 1,2,2,3,4 &1,2,2,3,4 &1,2,2,3,4 \\
    Head channels & 32 & 32 & 32 & 32\\
    Input projection & \cmark & \cmark & \cmark & \cmark \\
    Positional encoding & Sinusoidal & Sinusoidal & Sinusoidal & Sinusoidal \\
    $\dim \alpha$ & 10 & 10 & 10 & 10 \\
    \midrule
    \emph{Concat Conditioning} & & &  \\
    $\dim c_{S}$  & 5 & 4 & 5 & -  \\
    Context channels & 128 & 128 & 128 & -   \\
    Temporal kernel size & 3,1,1 & 3,1,1& 3,1,1 & -\\
    \midrule
    \emph{Training}  &  &  & \\
    \# train steps  & 60K  & 78K & 62K & 61K  \\
    Learning rate & $10^{-4}$ & $10^{-4}$  & $10^{-4}$ & $10^{-4}$ \\
    Batch size per GPU  & 18 & 1 & 16 & 13 \\
    \# GPUs  & 2 & 2 & 2 & 2 \\
    GPU-type  & A100-80GB & A100-80GB & A100-80GB & A100-80GB \\
    Sequence length & 10 & 10 & 10 & 10\\
    Training data FPS & 2 & 2 & 2 & 2 \\
    \midrule\midrule
    \textbf{Diffusion Setup}  &  &  & \\
    Diffusion steps & 1000 & 1000 & 1000 & 1000 \\
    Noise schedule & Linear & Linear & Linear & Linear \\
    $\beta_{0}$ & 0.0015 & $10^{-4}$ & 0.0015 & 0.0015 \\
    $\beta_{T}$ & 0.0195 & 0.02 & 0.0195 & 0.0195\\
    \midrule\midrule
    \textbf{Sampling Parameters}  &  &  & \\
    Sampler  &  DDIM & DDIM   & DDIM & DDIM  \\
    Steps  & 200 & 200  & 200 & 200 \\
    $\eta$  & 1.0 & 1.0  & 1.0  & 1.0 \\
    \bottomrule
\end{tabular}%
}
\label{table:hyperparameters_ablation}
\vspace{-0.4cm}
\end{table*}

\begin{table*}
\setlength{\tabcolsep}{4pt}
\caption{Hyperparameters for our autoencoder models.}
\centering
    \begin{tabular}{lccc}
    \toprule
    \textbf{Hyperparameter} & \textbf{Driving} & \textbf{WebVid} & \textbf{Mountain Biking}  \\
      & \textbf{(Video) VQAE} & \textbf{(Video) AE}& \textbf{(Video) AE} \\
    \midrule\midrule
    \textbf{Image LDM Autoencoder}  &  & &   \\
    \textit{Architecture}  &  & &  \\
    Regularization & \emph{VQ} & \emph{KL} & \emph{KL} \\
    Downsampling factor & 8 & 8 & 8\\ 
    $\dim z$& 4 & 4 & 4 \\
    Codebook size & 16384 & -& -  \\
    Base channels & 128 & 128 & 128 \\
    Channel multiplier & 1,2,2,4 & 1,2,2,4  & 1,2,2,4\\
    Depth & 2 & 2 & 2 \\
    Attention resolutions & 32 & - & - \\
    \midrule\midrule
    \textbf{Temporal Layers}  &  & &  \\
    \textit{Architecture}  &  &    &   \\
    Base channels & 128 & 128 & 128 \\
    Channel multiplier & 1,2,2,4 & 1,2,2,4& 1,2,2,4 \\
    Depth & 2 & 2 & 2 \\
    Temporal kernel size & 3,3,3 & 3,3,3 & 3,3,3 \\
    \midrule
    \textit{Training}  &  &  \\
    \# train steps  & 41K & 35K & 78K \\
    Learning rate & 4.0e-5 & 4.0e-5 &5.0e-5 \\
    Batch size per GPU  & 4  & 1 & 5 \\
    \# GPUs  & 8 & 40 & 8\\
    GPU-type & A100-40GB & A100-40GB & A100-80GB \\
    Sequence length & 6 & 6 & 6 \\
    \bottomrule
\end{tabular}%
\label{table:hyperparameters_ae}
\vspace{-0.4cm}
\end{table*}

\subsection{Description of Hyperparameters}
Our hyperparameter tables (\Cref{table:hyperparameters_gen,table:hyperparameters_ablation,table:hyperparameters_ae}) follow the hyperparameter tables from~\cite{rombach2021highresolution} and should be mostly self-explanatory. Here, we give some additional description for some parameters of the temporal layers:
\begin{itemize}
    \item $\dim \alpha$ (Architecture): Dimension of the learnable skip connection parameter $\alpha$ that is applied after temporal layers; see~\Cref{fig:pipeline} and~\Cref{sec:turning}. 
    \item $\dim c_S$ (Concat Conditioning): Dimension of the concatenated spatial conditioning of masks and masked (encoded) images; see~\Cref{sec:prediction}.
    \item Context channels (Concat Conditioning): Number of output channels of the learned downsampling layers, \cf~\Cref{sec:prediction} and~\Cref{fig:architecture}, which is concatenated to the input of the temporal convolutional layers.
    \item Temporal kernel size (Concat Conditioning): The kernel size of the 3D kernel of our temporal convolutional layers as [\texttt{time}, \texttt{height}, \texttt{width}].
\end{itemize}

\section{Quantitative Evaluation} \label{app:quant_eval}
We perform quantitative evaluations on all datasets. In particular, we compute Fréchet Inception Distance (FID) and Fréchet Video Distance (FVD) metrics. Since FVD can be
unreliable (discussed, for instance, in~\cite{brooks2022generating}), we additionally perform human evaluation. For text-to-video evaluation, we also compute (video) Inception Scores (IS) and CLIP Similarly scores (CLIPSIM).

\textbf{FVD:} The FVD metric measures similarity between real and generated videos~\cite{unterthiner2018towards}. We follow~\cite{brooks2022generating} and generate 2,048 videos (16 frames at 30 fps), except for the FVD score evaluation of our SD 2.1-based text-to-video Video LDM for the UCF-101 benchmark, where we used 10k model samples, following Make-A-Video~\cite{singer2022make}. We then extract features from a pre-trained I3D action classification model\footnote{We use the model provided by \cite{brooks2022generating,carreira2017quo}: \url{https://www.dropbox.com/s/ge9e5ujwgetktms/i3d_torchscript.pt?dl=1'}}. For reference statistics, we take random sequences of videos that contain at least 16 frames from the dataset. For driving scenario and mountain bike video generation, reference statistics are calculated from 2,048 videos; for text-to-video experiments, reference statistics are calculated from 10k videos. Our implementation is a simple adaption from the code provided by \cite{ge2022long}.\footnote{We use the official UCF FVD evaluation code provided by \cite{ge2022long}: \url{https://github.com/SongweiGe/TATS/}}

\textbf{FID:} To compute FID~\cite{heusel2017gans} (for driving video and mountain biking video synthesis), we randomly extract 10k frames (except for mountain biking for which we extract 50k frames) from the 2,048 generated videos as well as from dataset videos. We then extract features from a pre-trained Inception model.\footnote{We use the model provided by~\cite{karras2021aliasfree}: \url{https://api.ngc.nvidia.com/v2/models/nvidia/research/stylegan3/versions/1/files/metrics/inception-2015-12-05.pkl}.}

\begin{figure}
    \centering
    \includegraphics[width=0.8\textwidth]{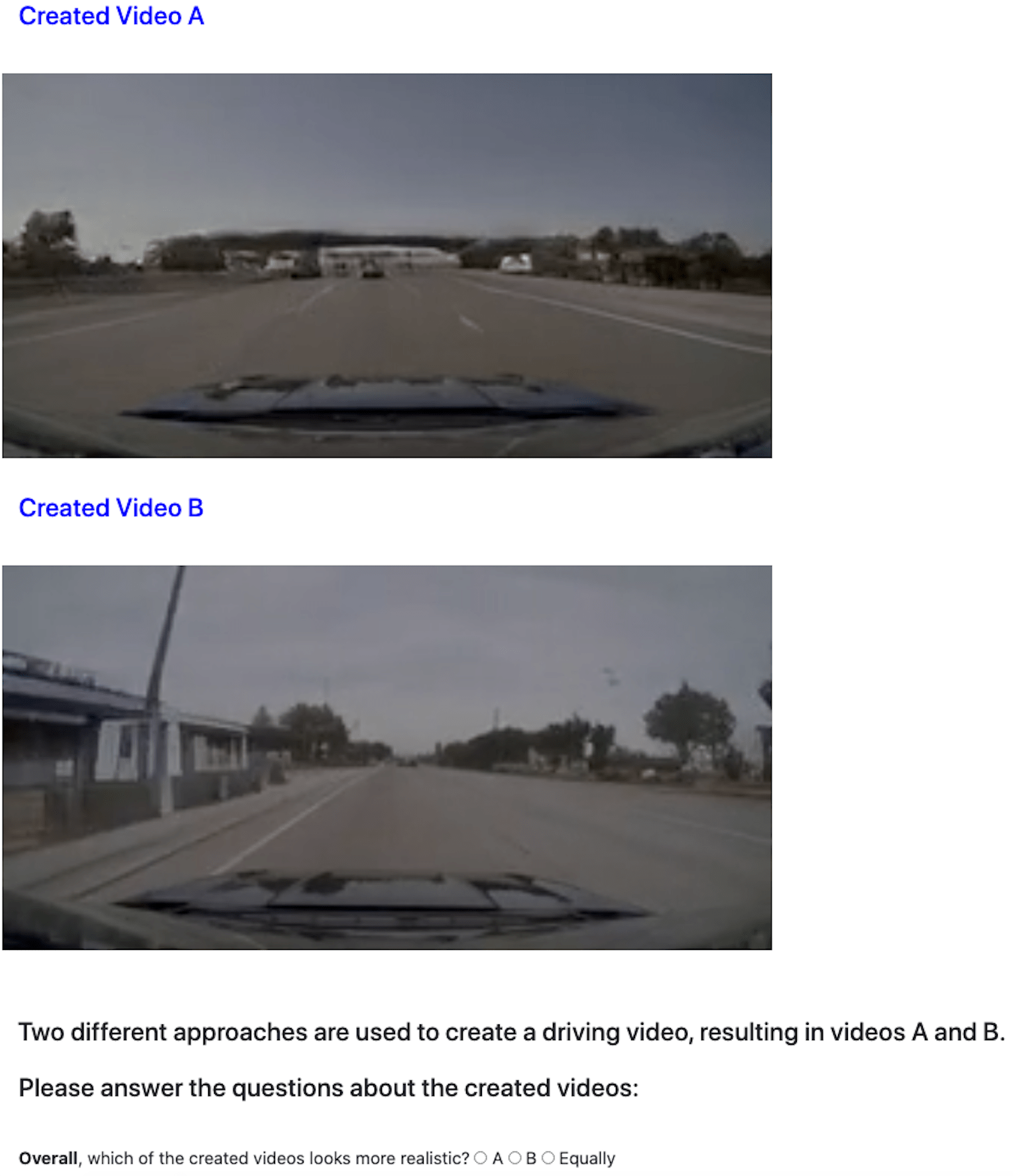}
    \caption{Screenshot of instructions provided to participants for the human evaluation study.}
    \label{fig:human_eval_driving}
\end{figure}

\textbf{Human evaluation:} We conduct human evaluation (user study) on Amazon Mechanical Turk to evaluate the realism of generated videos by our method in comparison to LVG~\cite{brooks2022generating}. For our user study, we create 100 videos, each of length 4 seconds. The user study shows pairs of videos, and each pair has one random video from our method and one from LVG. For each pair, we instruct the participants to select the favorable video in a non-forced-choice response, i.e., the participants can vote for ``equally realistic''. See~\Cref{fig:human_eval_driving} for a screenshot. Note that the A-B order of the pairs is also assigned randomly. 

Each video pair was shown to four participants resulting in 400 responses per dataset. We only select workers from English-speaking countries. There are no human subjects and we do not study the participants themselves; therefore, IRB review is not applicable. We pay \$0.04 for answering a single question.

\textbf{Inception Score:} In our text-to-video experiments on UCF-101, we also evaluated inception scores (IS)~\cite{salimans2016inceptionscore}. 
Following previous work on video synthesis~\cite{hong2022cogvideo,singer2022make}, we used a C3D~\cite{tran2015c3d} model trained on UCF-101 to calculate a video version of the inception score. It is calculated from 10k samples using the official code of TGANv2~\cite{TGAN2020}.\footnote{We use the official UCF IS evaluation code provided by \cite{TGAN2020}: \url{https://github.com/pfnet-research/tgan2}}

\textbf{CLIP Similarity (CLIPSIM):} In our text-to-video experiments on MSR-VTT, we also evaluated CLIP similarity (CLIPSIM)~\cite{wu2021godiva}. The MSR-VTT test set contains 2990 examples and 20 descriptions/prompts per example. We generate 2990 videos (16 frames at 30 fps) by using one random prompt per example. We then average the CLIPSIM score of the 47,840 frames. We use the \emph{ViT-B/32}~\cite{radford2021learning} model to compute the CLIP score. 
\\
\\
Note that we directly use UCF class names as text conditioning, e.g., “Basketball Dunk”, in contrast to Make-A-Video~\cite{singer2022make}, which manually constructs a template sentence for each class.

\section{Experiment Details}\label{app:exp_details}
In addition to the model hyperparameters above in \Cref{table:hyperparameters_gen}, here we provide additional details on the experiments presented in the paper.

\subsection{Details: High-Resolution Driving Video Synthesis---\Cref{sec:driving_exp}}
We initially train our base Image LDM (both autoencoder and latent space diffusion model) on 1 fps videos from the RDS dataset at resolution $128\times256$. We then train the temporal layers for sparse key frame prediction with 2 fps. For that, we extracted 2 fps videos from the 10 fps driving data. As discussed in the main paper, we inform the model about the number of frames given for prediction, which is either 0, 1, or 2.  

The pixel-space $4\times$ upsampler that scales the resolution to $512\times1024$ is trained using the same data, but at a correspondingly higher resolution. The upsampler is trained with noise augmentation and conditioning on the noise level~\cite{ho2021cascaded,saharia2022imagen}. During training we randomly sample $t\in\{0, \dots, 250\}$ and perturb the low-resolution conditioning following our variance-preserving diffusion process (using the same linear noise schedule as for the main upsampling diffusion model). At inference time, we use 150 steps for perturbation.

The LDM decoder fine-tuning is performed on 30 fps videos.

The temporal interpolation model is trained using 30 fps video data, which we process for the different tasks. We train the temporal interpolation model to first scale from 1.875 fps to 7.5 fps, and then to scale from 7.5 fps to 30 fps. We are using one interpolation model with shared parameters for that, providing a conditioning label to indicate to the model which of the two temporal upsampling operations is desired.

\textbf{Video Generation.} For video synthesis, we first generate a single frame using the image LDM, then we run the prediction model, conditioning on the single frame, to generate a sequence of key frames. When extending the video, we again call the prediction model, but condition on two frames (which captures directional information) to produce consistent motion. Next, we optionally perform two steps of the temporal interpolation, going from 1.875 to 7.5 fps and from 7.5 to 30 fps, respectively. Also optionally, the video fine-tuned upsampler is then run over portions of 8 video frames.

\textbf{LVG Baseline.} Our main baseline is the state-of-the-art Long Video GAN (LVG)~\cite{brooks2022generating}, which we trained on the RDS data at 10 fps and resolution $128\times256$ (comparisons to our Video LDM are carried out at this resolution and fps). LVG's low resolution component was trained with a batch size of 64 and two gradient accumulation steps and an R1~\cite{Mescheder2018ICML} penalty of 1.0. For the super resolution module, we used a batch size of 2 without gradient accumulation and the same R1 penalty. Other than that, we used LVG's default hyperparameters.

\textbf{Ablation Studies.} For the ablation studies (\Cref{sec:ablations}), we trained a smaller version of our Video LDM as well as an ``end-to-end'' version of our Video LDM, which simultaneously trains the spatial and temporal layers of the latent diffusion model from scratch. The hyperparameters for trained these models are shown in \Cref{table:hyperparameters_ablation}.

\textbf{Bounding box-conditioned image LDM.}
For the experiments in \Cref{sec:driving_simulation}, we trained a separate bounding-box conditioned image LDM (no videos) on the 100k independent annotated frames (see \Cref{sec:rds_details}). It uses the same hyperparameters as our main model's image LDM backbone for training. The conditioning is implemented using cross-attention~\cite{rombach2021highresolution}: We learn embeddings of the bounding box coordinates, fuse all coordinate embeddings using a transformer, and attend to the resulting fused embeddings.

\subsection{Details: Text-to-Video with Stable Diffusion---\Cref{sec:text-to-video}}\label{app:exp_details_text_to_video}
We ran experiments with three of the publicly available Stable Diffusion (SD) checkpoints as image LDM backbones: 1.4\footnote{\url{https://huggingface.co/CompVis/stable-diffusion-v-1-4-original}}, 2.0\footnote{\url{https://huggingface.co/stabilityai/stable-diffusion-2}}, and 2.1\footnote{\url{https://huggingface.co/stabilityai/stable-diffusion-2-1}}.
Most of the research project was conducted with the SD 1.4-based model and the SD 2.0- and SD 2.1-based Video LDMs were trained primarily for exploration purposes and additional qualitative results later (however, our quantitative evaluation for text-to-video in Tabs.~\ref{tab:ucf} and~\ref{tab:msrvtt} uses the latest SD 2.1-based model; see discussion below).
Unless indicated otherwise, all shown samples in the different figures throughout the paper are from our SD 1.4-based model without upsampler. Samples from the SD 2.0-based model with upsampler (see below) are presented in \Cref{fig:text2image_samples,fig:sd20_samples_supp_1,fig:sd20_samples_supp_2,fig:sd20_samples_supp_3,fig:sd20_samples_supp_4} and in the second example in \Cref{fig:teaser}. Many samples from the SD 2.1-based Video LDM are shown on our project page (\url{https://research.nvidia.com/labs/toronto-ai/VideoLDM/}).

Since SD is trained on images at resolution $512\times512$, naively applying it to the smaller-sized videos of the WebVid-10M dataset would lead to severe degradations in image quality.
We therefore first fine-tune the Stable Diffusion image backbone (spatial layers) on the WebVid-10M data. Specifically, we resize and center-crop the WebVid-10M videos to $320\times512$ resolution and then fine-tune the SD latent space diffusion model on independent encoded frames from the videos. We rely on standard SD training hyperparameters, with a learning rate of $10^{-4}$. Note that while this (spatial layer) fine-tuning of SD on the WebVid-10M data is necessary to prevent out-of-distribution problems when modeling videos in the next step, it also slightly hurts the overall image quality compared to the original SD checkpoint, because the WebVid data is arguably of lower visual quality than the images used to train the original SD model. We assume that training with more and higher-quality video data will solve this.

Next, as described in \Cref{sec:video_ldm} we video fine-tuned both SD's latent space diffusion model and its decoder (\Cref{sec:dec_finetuna}) using the WebVid-10M videos.
Note that our temporal layers also consume the text conditioning (\Cref{{fig:architecture}}).
We do not train a prediction model here, but only train for text-to-video generation without any additional context frames. 
Overall, we train our pipeline for generation of videos consisting of 113 frames (which we can render, for instance, into clips of 4.7 seconds length at 24 fps or into clips of 3.8 seconds length at 30 fps.).
To synthesize longer videos, we can optionally apply our temporal layers ``convolutionally in time''; similarly, to generate spatially extended higher resolution videos we can apply the model ``convolutionally in space'' (see \Cref{app:convolutional_ldm}). 

To reach high frame rates and enable smooth video generation, we again train an interpolation model that can temporally upsample videos from 1.875 fps to 7.5 fps as well as 7.5 fps to 30 fps. Note that in contrast to our experiments on driving scenes and mountain biking, for the text-to-video interpolation models we trained all model parameters, including the ones in the spatial layers of the image LDM backbone. For the SD 1.4/2.0-based Video LDMs, we initialized the spatial layers from the fine-tuned SD checkpoint and trained the temporal layers from scratch, whereas for SD 2.1 all layers (except for the first convolutional layer) are initialized from the trained keyframe model (including the temporal layers).
For the SD 2.0-based Video LDM, we did not use text-conditioning in the interpolation model and feed empty text into SD's text inputs (the SD 1.4- and SD 2.1-based Video LDMs did use text conditioning also in their interpolation models).
Since we do two rounds of interpolation, we condition the models on the fps rate to which we interpolate (by cross attention to a corresponding embedding).
Moreover, for these text-to-video interpolation models we did not use the learned downsampling approach (\Cref{fig:architecture}), but instead concatenated the context via partially masked out frames (and the mask itself) to the channel dimension $C$ of the U-Net input.
Furthermore, we use conditioning augmentation~\cite{ho2021cascaded} for this additional input during training (we randomly sample $t\in\{0, \dots, 250\}$ and perturb the conditioning inputs following our variance-preserving diffusion process, using the same linear noise schedule as for the main diffusion model). At test time, the conditioning augmentation is turned off, i.e., the noise level is set to zero.
Finally, the interpolation models of the SD 1.4-based and SD 2.1-based Video LDMs use one conditioning frame on either side, whereas the SD 2.0-based model uses two conditioning frames on either side.
Except for the training differences discussed above, the SD 1.4-based, SD 2.0-based and SD 2.1-based Video LDMs are trained with the same hyperparameters (see \Cref{table:hyperparameters_gen}).

\textbf{Upsampler Training:} As described in \Cref{sec:text-to-video}, we also video fine-tune the publicly available text-guided Stable Diffusion $4\times$-upscaler\footnote{\url{https://huggingface.co/stabilityai/stable-diffusion-x4-upscaler}}, which is itself a latent diffusion model. We train the upsampler for temporal alignment in a patch-wise manner on $320\times320$ cropped videos (WebVid-10M), which are embedded into a $80\times80$ latent space. The $80\times80$ low resolution conditioning videos are concatenated to the $80\times80$ latents. The learnt temporal alignment layers are text-conditioned, like for our base text-to-video LDMs. Like for the driving models, the upsampler is trained with noise augmentation and conditioning on the noise level, following previous work~\cite{ho2021cascaded,saharia2022imagen}. During training we randomly sample $t\in\{0, \dots, 250\}$ and perturb the low-resolution conditioning following our variance-preserving diffusion process (using the same linear noise schedule as for the main upsampling diffusion model). At inference time, we use 30 steps for perturbation.
Moreover, we apply the model at extended resolution during inference. We provide $320\times512$ resolution videos as low resolution input, predict $320\times512$ resolution latents, and decode to $1280\times2048$ resolution videos. Note that we did not find it necessary to temporally fine-tune the decoder of this latent diffusion model upscaler.

\textbf{Results in Tabs.~\ref{tab:ucf} and~\ref{tab:msrvtt}:} The results reported in those tables are based on experiments with the SD 2.1-based Video LDM. In Tabs.~\ref{tab:ucf_ext} and~\ref{tab:msrvtt_ext}, we provide extended results including an earlier, preliminary evaluation of our Video LDM based on the SD 1.4 model. Note that the FVD score was calculated with only 2,048 samples from the model in that case. The FVD score for the SD 2.1-based Video LDM was calculated with 10k generated videos, following Make-A-Video~\cite{singer2022make}, for fair comparison with this work.

Overall, we find that the SD 2.1-based Video LDM performs better than the SD 1.4-based one on these benchmarks. For UCF-101, our SD 2.1-based Video LDM even slightly outperforms Make-A-Video~\cite{singer2022make} on Inception Score.

\begin{table}[t]
    \centering
        \caption{\small UCF-101 text-to-video generation performance.\vspace{-1em}}
    \label{tab:ucf_ext}
    
    \resizebox{ 0.45\linewidth}{!}{%
    \begin{tabular}{l  c c c}
        \toprule
        \textbf{Method} &  Zero-Shot & IS ($\uparrow$) & FVD ($\downarrow$)  \\
        \midrule
        CogVideo (Chinese)~\cite{hong2022cogvideo} & Yes  & 23.55 & 751.34 \\
        CogVideo (English)~\cite{hong2022cogvideo} & Yes  & 25.27 & 701.59 \\
        MagicVideo~\cite{zhou2022magicvideo} & Yes  & - & 699.00 \\
        Make-A-Video~\cite{singer2022make} & Yes  & 33.00 & 367.23 \\
        \midrule
        Video LDM (SD 1.4) \emph{(Ours)} & Yes & 29.49 & 656.49 \\
        Video LDM (SD 2.1) \emph{(Ours)} & Yes & 33.45 & 550.61 \\
        \bottomrule
    \end{tabular}
    }
\end{table}
\begin{table}[t]
    \centering
        \caption{\small MSR-VTT text-to-video generation performance.\vspace{-1em}}
    \label{tab:msrvtt_ext}
    
    \resizebox{ 0.45\linewidth}{!}{%
    \begin{tabular}{l  c c}
        \toprule
        \textbf{Method} &  Zero-Shot & CLIPSIM ($\uparrow$)  \\
        \midrule
        GODIVA~\cite{wu2021godiva} & No  & 0.2402 \\
        N\"{U}WA~\cite{wu2022nuwa} & No & 0.2439 \\
        CogVideo (Chinese)~\cite{hong2022cogvideo} & Yes  & 0.2614 \\
        CogVideo (English)~\cite{hong2022cogvideo} & Yes  & 0.2631 \\
        Make-A-Video~\cite{singer2022make} & Yes  & 0.3049\\
        \midrule
        Video LDM (SD 1.4) \emph{(Ours)} & Yes & 0.2848 \\
        Video LDM (SD 2.1) \emph{(Ours)} & Yes & 0.2929 \\
        \bottomrule
    \end{tabular}
    }
\end{table}

\subsubsection{Number of Model Parameters}\label{app:model_params}
Our text-to-video LDMs that are based on Stable Diffusion have
\begin{itemize}
    \item 84 million parameters in the autoencoder (decoder is fine-tuned).
    \item 860 (SD 1.4) / 865 (SD 2.0/2.1) million parameters in the image backbone LDM, this is, in the spatial layers not including the CLIP text encoder (not trained).
    \item 649 (SD 1.4) / 656 (SD 2.0/2.1) million parameters in the temporal layers (trained).
    \item 123 (SD 1.4 uses CLIP ViT-L/14) / 354 (SD 2.0/2.1 uses OpenCLIP-ViT/H) million parameters in the text encoder (not trained).
    \item 1,509 million parameters in the interpolation latent diffusion model (trained). We use the same interpolation model for our SD 1.4- and SD 2.0/2.1-based models.
\end{itemize}
Combined, for the SD-2.0/2.1-based Video LDMs this sums to around 3.1B parameters in the autoencoder and diffusion model components (excluding the CLIP text embedders) in the low resolution text-to-video LDM. Out of this, only around 2.2B parameters are actually trained.

Moreover, our fine-tuned text-to-video latent upsampler that is based on the public Stable Diffusion $4\times$ upscaler has
\begin{itemize}
    \item 55 million parameters in the autoencoder (not trained).
    \item 473 million parameters in the image backbone LDM  (not trained).
    \item 449 million parameters in the temporal layers (trained).
    \item 354 million parameters in the OpenCLIP-ViT/H text encoder (not trained).
\end{itemize}
This sums to a total of 977 parameters in the autoencoder and diffusion model components of the upsampler, out of which only 449 were trained.

We see that compared, for instance, to Imagen Video~\cite{ho2022imagenvideo}, which has 11.6B parameters, our model is much smaller. Yet, it can produce high quality videos, which we attribute to the efficient LDM framework. CogVideo~\cite{hong2022cogvideo} also has much more parameters, around 9B, than our Video LDM.
We suspect that Make-A-Video~\cite{singer2022make} similarly is a much larger model than ours.

\subsection{Details: Personalized Text-to-Video with Dreambooth---\Cref{sec:dreambooth}}
Using DreamBooth~\cite{ruiz2022dreambooth},
we fine-tune our Stable Diffusion spatial backbone (after fine-tuning on WebVid-10M, as described above) on small sets of images of certain objects (using SD 1.4). We use 256 regularization images and train for 800 iterations using a learning rate $10^{-6}$. We found it greatly beneficial to train both the U-Net as well as the CLIP text encoder. Our DreamBooth code is based on the following public codebase: \url{https://github.com/XavierXiao/Dreambooth-Stable-Diffusion}. After training, we insert the temporal layers
from the previously video-tuned Stable Diffusion (without DreamBooth) into the new DreamBooth version of the original Stable Diffusion model. Importantly, for video generation, the spatial layers use the DreamBooth-fine-tuned CLIP text encoder whereas the temporal layers use the standard CLIP text encoder they were trained on.
\section{Additional Results} \label{app:additional_results}
Here we are showing further qualitative and quantitative results, including sampled videos from all our models.

In \Cref{app:extended_text2video}, we show more samples from our Stable Diffusion-based text-to-video LDM. This includes samples that were generated by using the model convolutional-in-time as well as convolutional-in-space (see~\Cref{app:convolutional_ldm}). We also discuss video fine-tuning of the decoder for this text-to-video LDM.

In \Cref{app:mountain_bike}, we show additional results from a Video LDM model we trained on a dataset of Mountain Bike videos. This includes quantitative comparisons to the previous state-of-the-art Long Video GAN baseline.

In \Cref{app:extended_driving}, we present more qualitative and quantitative results from our main Video LDM that was trained on real-world driving data.

\begin{table}[]
    \centering
    \caption{Effects of video fine-tuning of the decoders of the compression framework encompassing Video LDM. Here we compute reconstruction FVD and FID based on 2,048 examples from the respective datasets (WebVid data is used to fine-tune the decoder of the text-to-video LDM based on Stable Diffusion). \emph{fine-tuned} denotes our video fine-tuned decoders.}
    \label{tab:vid-ft-supp}
    \begin{tabular}{l c c | c c}
        \toprule
        Dataset & \multicolumn{2}{c|}{WebVid~\cite{bain21frozen}} & \multicolumn{2}{c}{Mountain Biking~\cite{brooks2022generating}} \\
        Method & \footnotesize\emph{image-only} & \footnotesize\emph{fine-tuned} & \footnotesize\emph{image-only} & \footnotesize\emph{fine-tuned} \\
        \midrule
        FVD & 35.82 & \textbf{18.66} & 73.78 &  \textbf{25.55} \\
        FID & 13.89 & \textbf{11.68} & 20.76 & \textbf{18.65} \\
        \bottomrule
    \end{tabular}
\end{table}

\subsection{Text-to-Video} \label{app:extended_text2video}
\subsubsection{Video-Finetuning of our Decoders}
We perform a small ablation experiment over the video fine-tuning of our decoder (as described in~\Cref{sec:dec_finetuna}). As can be seen in~\Cref{tab:vid-ft-supp}, video fine-tuning the decoder allows for a significant performance boost for out text-to-video model, similar to what we have observed for the driving model in the main paper.
\subsubsection{More Samples}
\begin{figure*}[t!]
    \centering
    \includegraphics[width=0.75\textwidth]{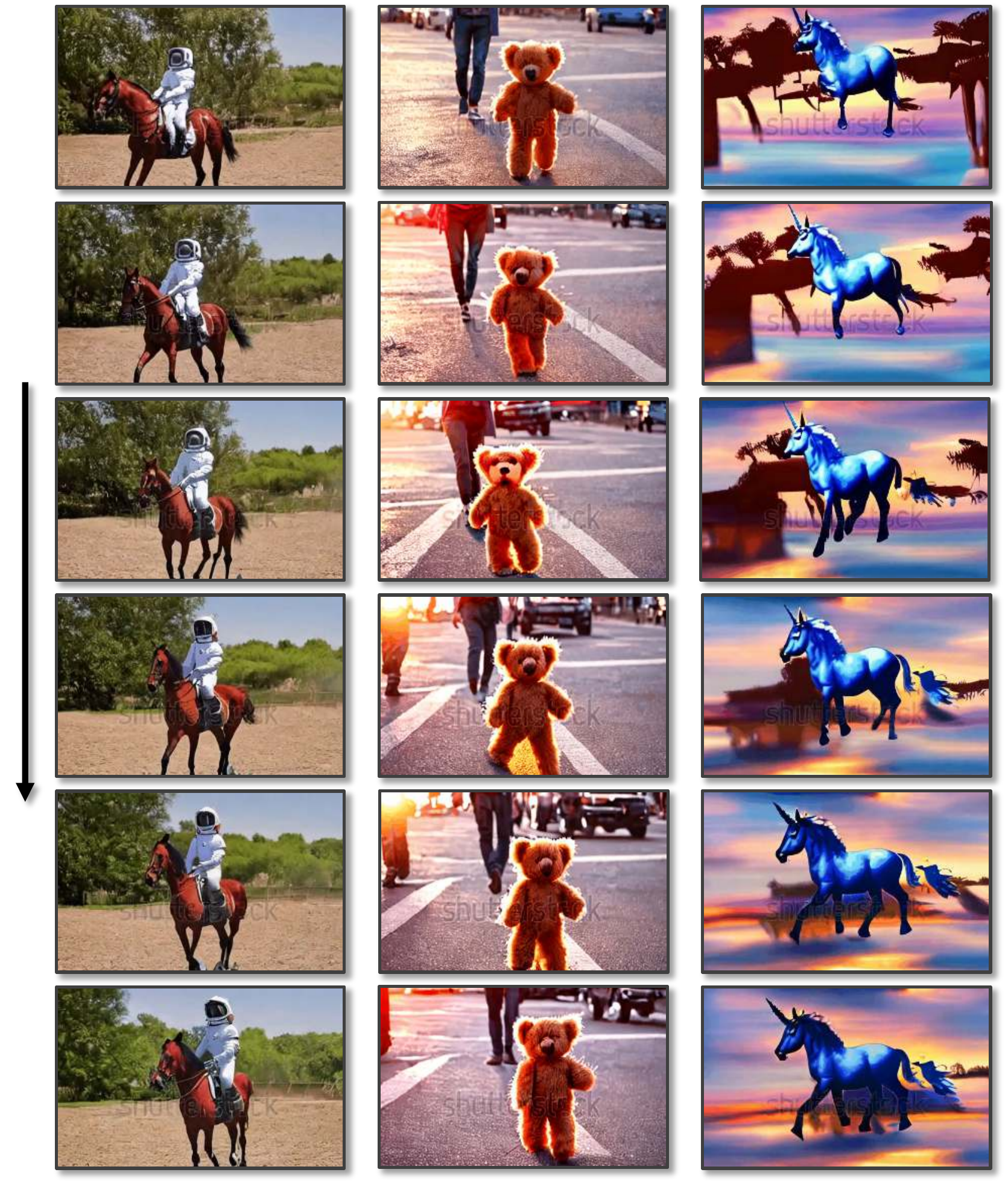}
    \caption{\small Generated videos at resolution $320 \times 512$. Captions from left to right are: ``An astronaut riding a horse, high definition, 4k'', ``Teddy bear walking down 5th Avenue, front view, beautiful sunset, close up, high definition, 4k'', and ``A blue unicorn flying over a fantasy landscape, animated oil on canvas''.
    Frames are shown at 2 fps.} 
    \vspace{-0.5em}
    \label{fig:common}
\end{figure*}

\begin{figure*}[t!]
    \centering
    \includegraphics[width=0.8\textwidth]{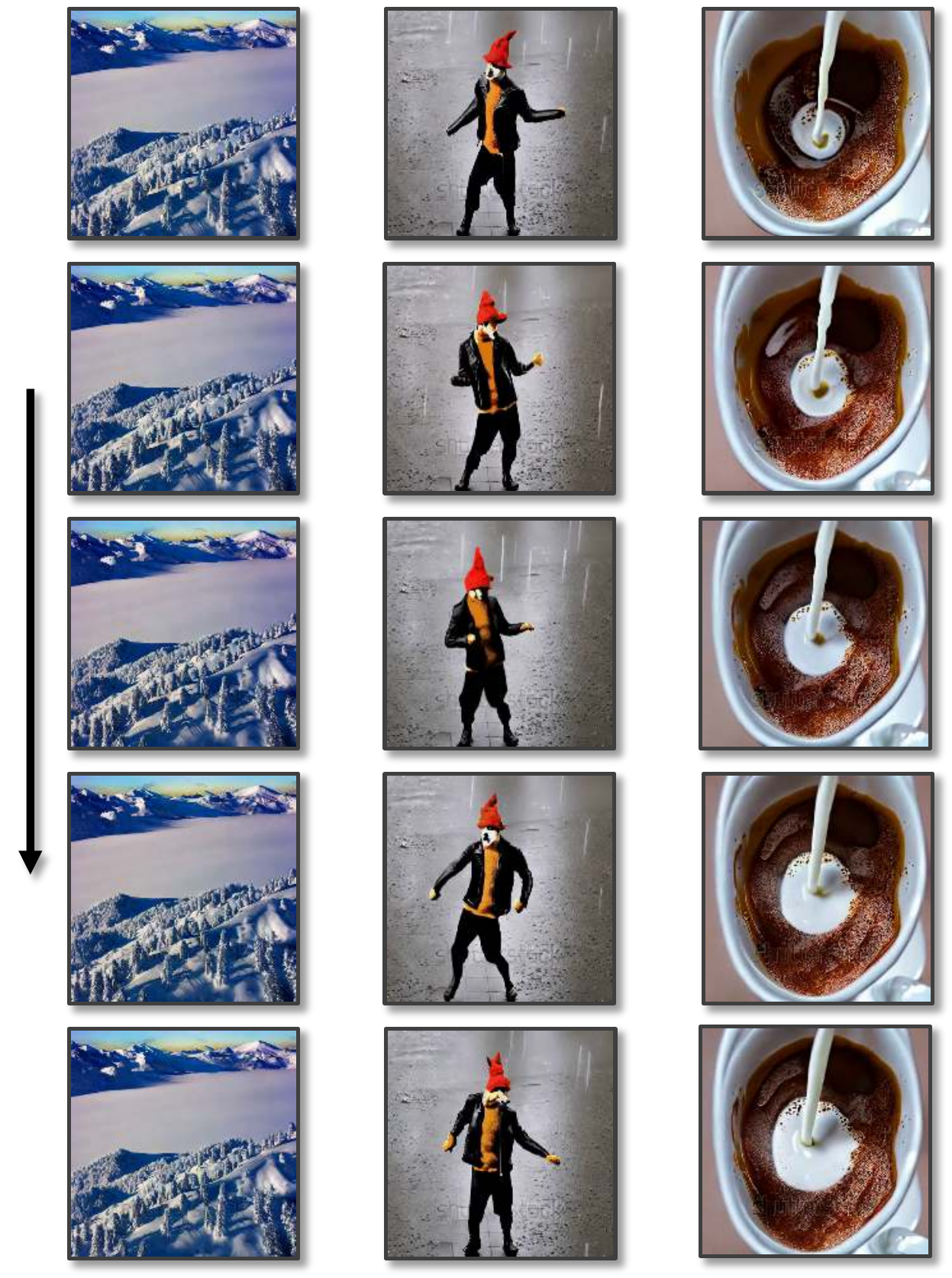}
    \caption{\small Generated videos at resolution $512\times512$ (extended ``convolutional in space''; see~\Cref{app:convolutional_ldm}).
    Captions from left to right are: ``Aerial view over snow covered mountains'', ``A fox wearing a red hat and a leather jacket dancing in the rain, high definition, 4k'', and ``Milk dripping into a cup of coffee, high definition, 4k''.
    Frames are shown at 2 fps.
    } \vspace{-0.5em}
    \label{fig:space_conv1}
\end{figure*}

\begin{figure*}[t!]
    \centering
    \includegraphics[width=0.8\textwidth]{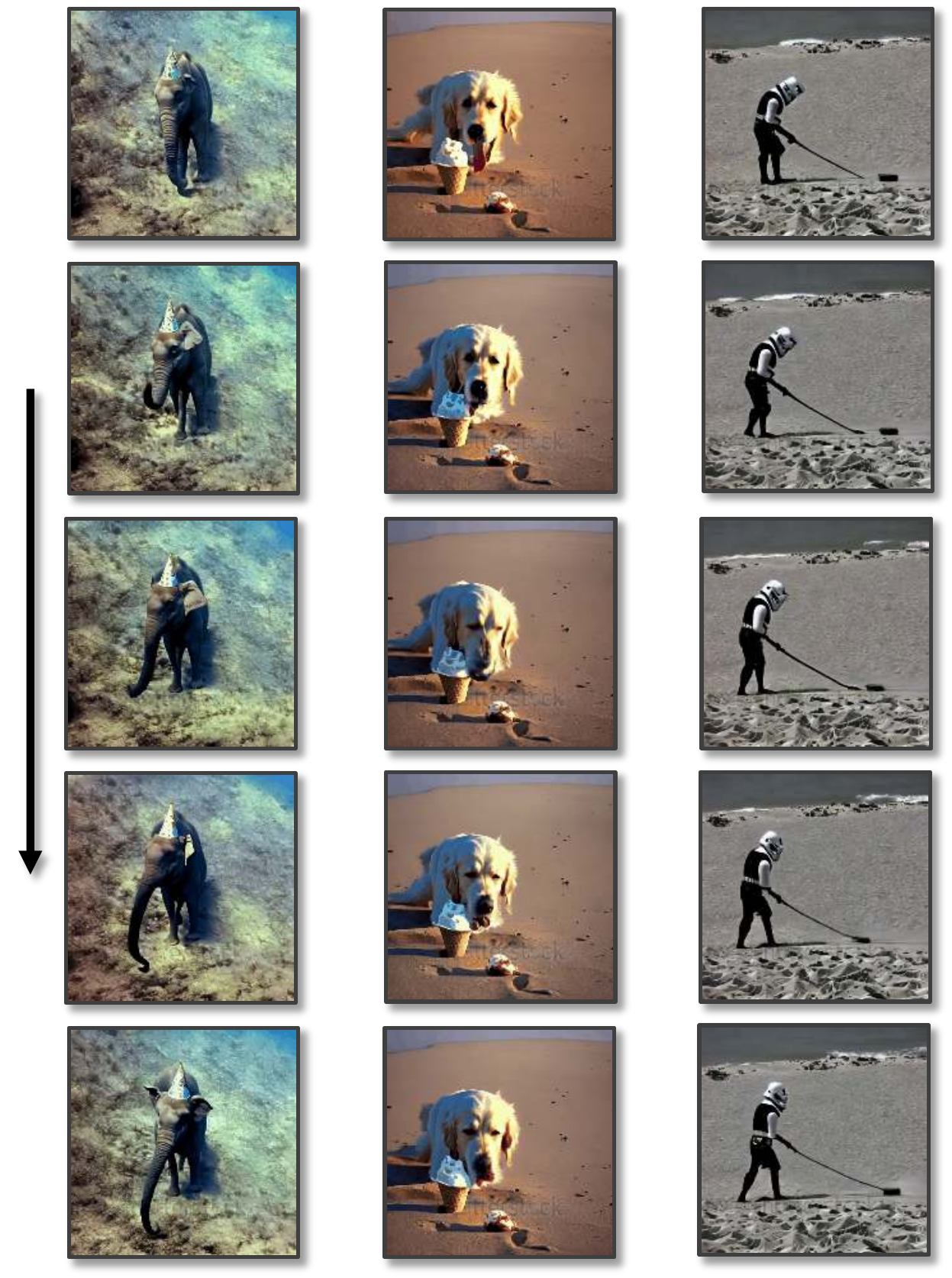}
    \caption{\small Generated videos at resolution $512\times512$ (extended ``convolutional in space''; see~\Cref{app:convolutional_ldm}).
    Captions from left to right are: ``An elephant wearing a birthday hat walking under the sea'', ``A golden retriever eating ice cream on a beautiful tropical beach at sunset, high resolution'', and ``A storm trooper vacuuming the beach''.
    Frames are shown at 2 fps.} \vspace{-0.5em}
    \label{fig:space_conv2}
\end{figure*}

\begin{figure*}[t!]
    \centering
    \includegraphics[width=0.8\textwidth]{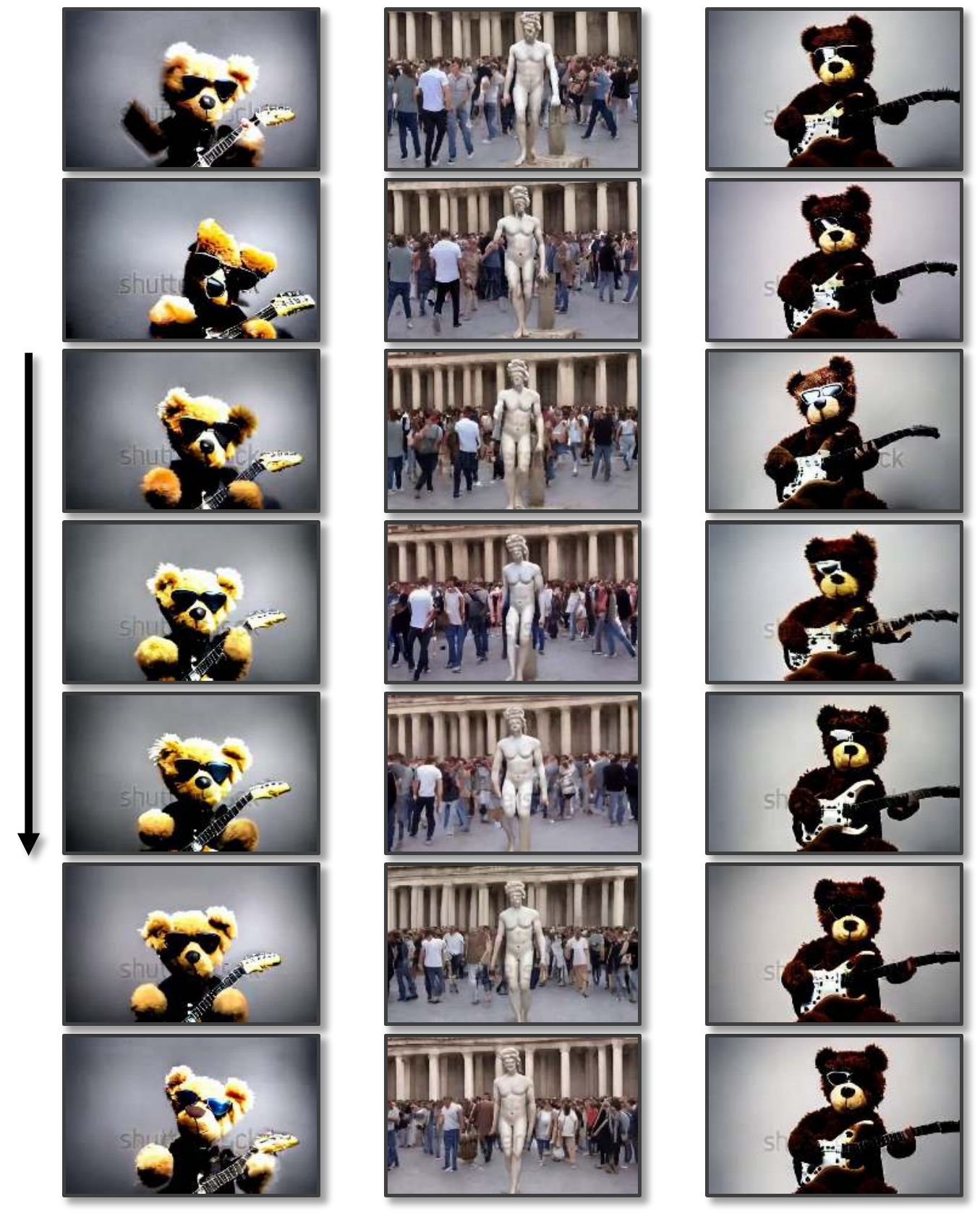}
    \caption{\small \small Generated videos at resolution $320\times512$ (extended ``convolutional in time'' to 8 seconds each; see~\Cref{app:convolutional_ldm}). 
    Captions from left to right are: ``A teddy bear wearing sunglasses and a leather jacket is headbanging while playing the electric guitar, high definition, 4k'', ``An ancient greek statue on a crowded square suddenly becomes alive and starts to walk, high definition, 4k'', and ``A teddy bear wearing sunglasses playing the electric guitar, high definition, 4k''.
    Frames are shown at 1 fps.} \vspace{-0.5em}
    \label{fig:time_conv1}
\end{figure*}

\begin{figure*}[t!]
    \centering
    \includegraphics[width=0.65\textwidth]{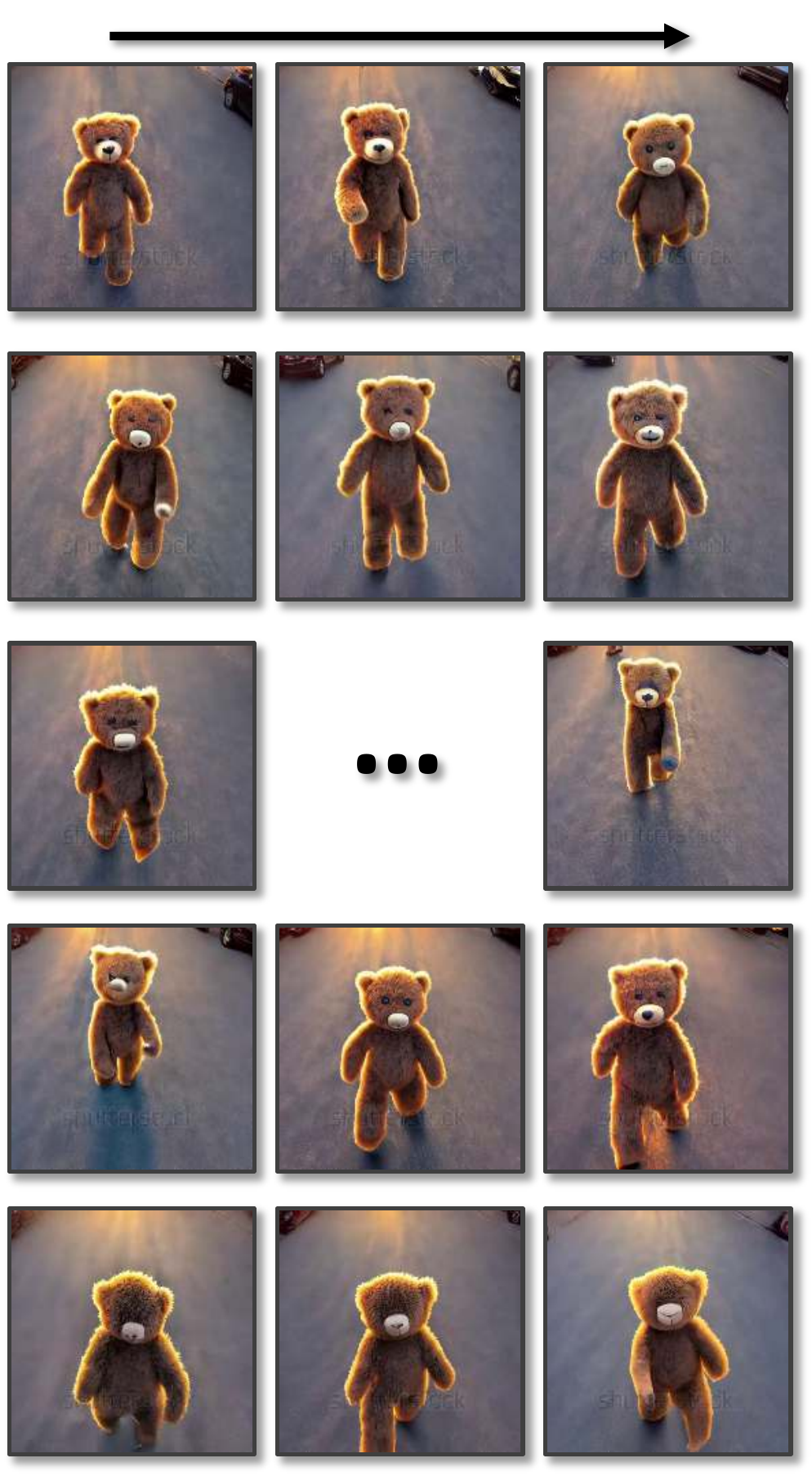}
    \caption{\small Generated 30 second video of ``a teddy bear walking down the road in the sunset, high definition, 4k'' at resolution $512\times512$ (extended ``convolutional in space'' and also ``convolutional in time''; see~\Cref{app:convolutional_ldm}). Frames are shown at 1 fps.} \vspace{-0.5em}
    \label{fig:longteddy}
\end{figure*}

\begin{figure*}[t!]
    \centering
    \includegraphics[width=0.65\textwidth]{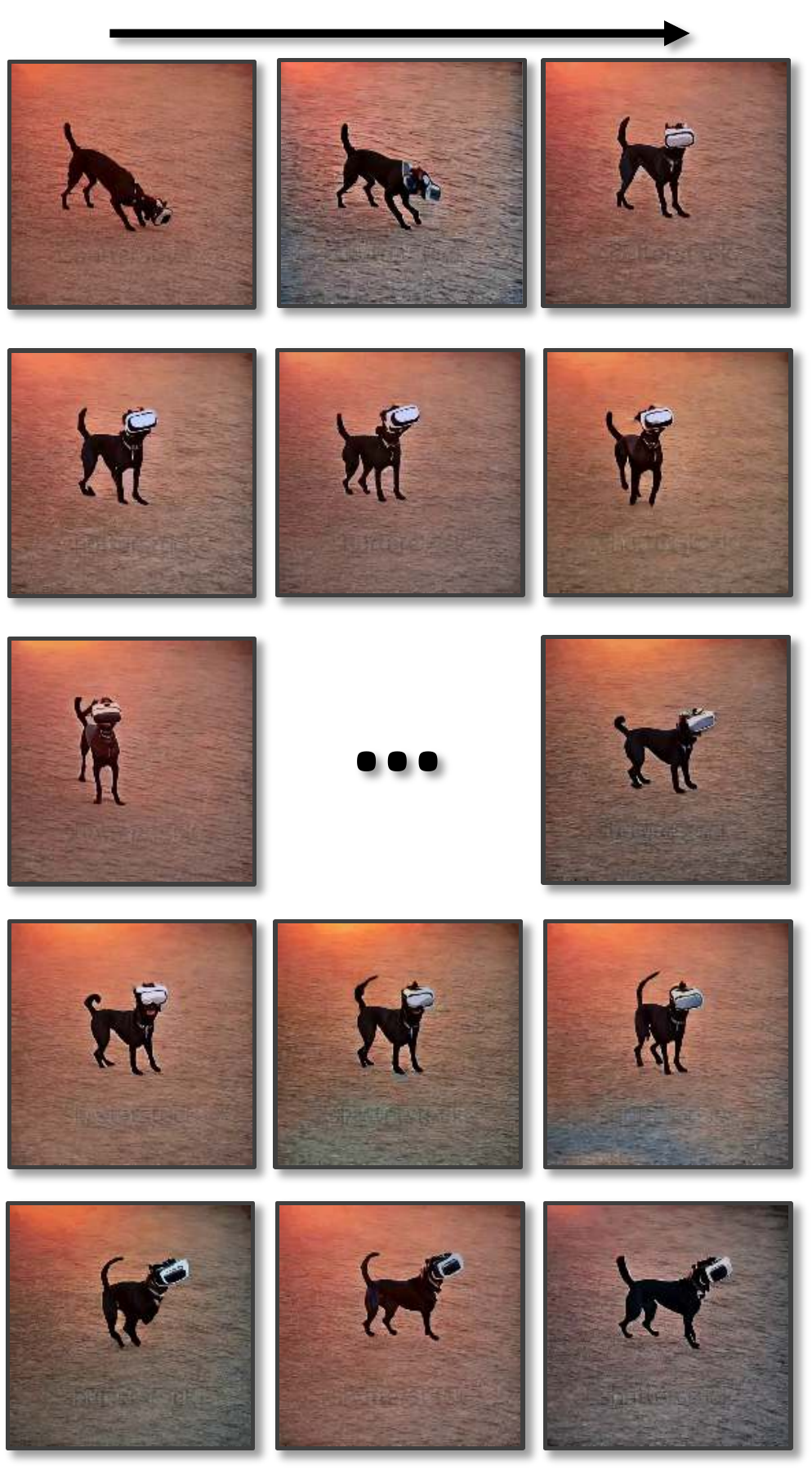}
    \caption{\small Generated 8 second video of ``a dog wearing virtual reality goggles playing in the sun, high definition, 4k'' at resolution $512\times512$ (extended ``convolutional in space'' and ``convolutional in time''; see~\Cref{app:convolutional_ldm}). Frames are shown at 4 fps.} \vspace{-0.5em}
    \label{fig:hotdog}
\end{figure*}

\begin{figure*}[t!]
    \centering
    \includegraphics[width=0.9\textwidth]{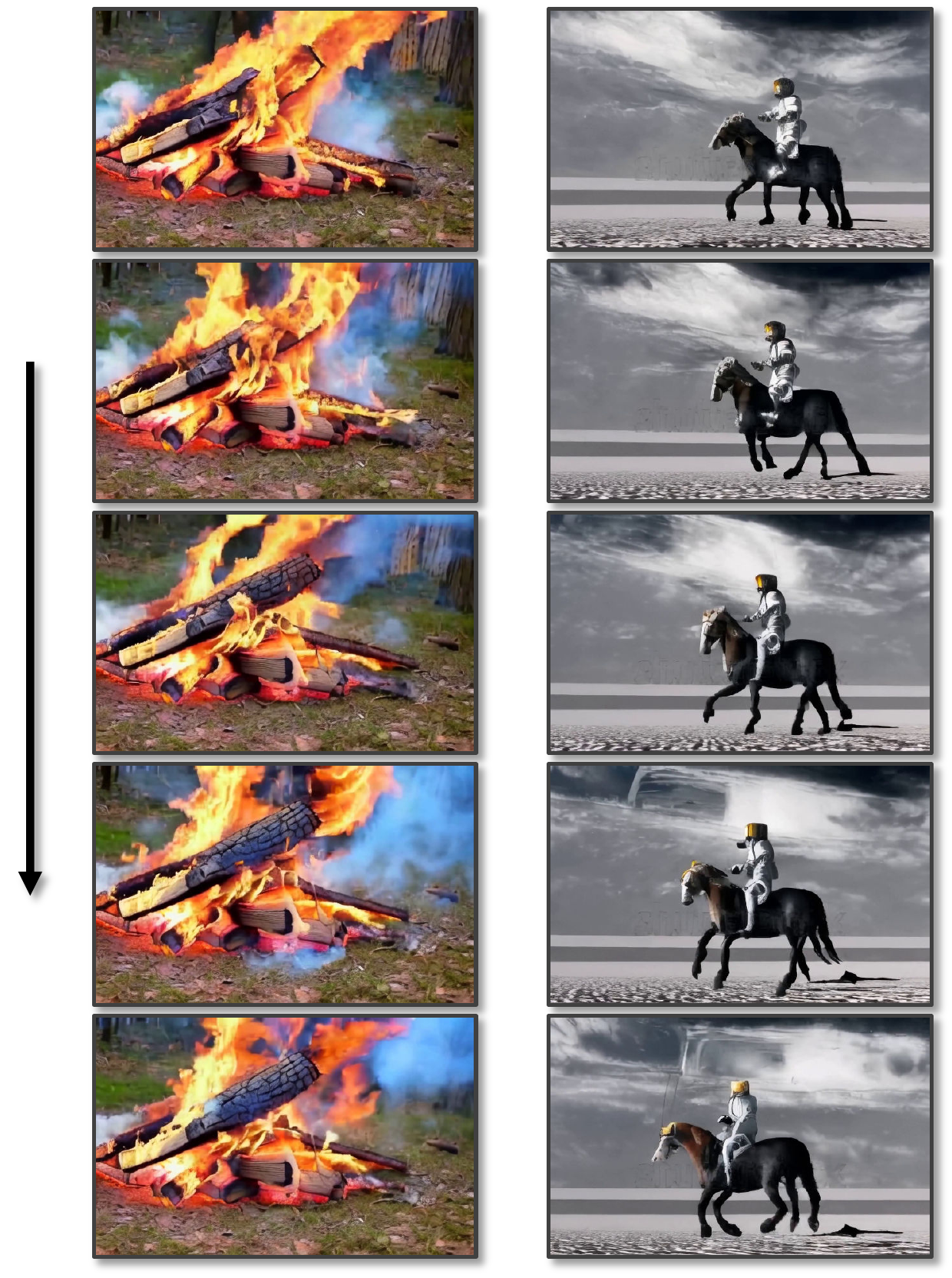}
    \caption{\small Generated videos at resolution $1280 \times 2048$ using our Stable Diffusion 2.0-based model and including our video fine-tuned text-to-video latent upsampler. Captions from left to right are: ``Burning firewood'' and ``An astronaut riding a horse, 4k, high definition''.
    Frames are shown at 2 fps.} 
    \vspace{-0.5em}
    \label{fig:sd20_samples_supp_1}
\end{figure*}

\begin{figure*}[t!]
    \centering
    \includegraphics[width=0.9\textwidth]{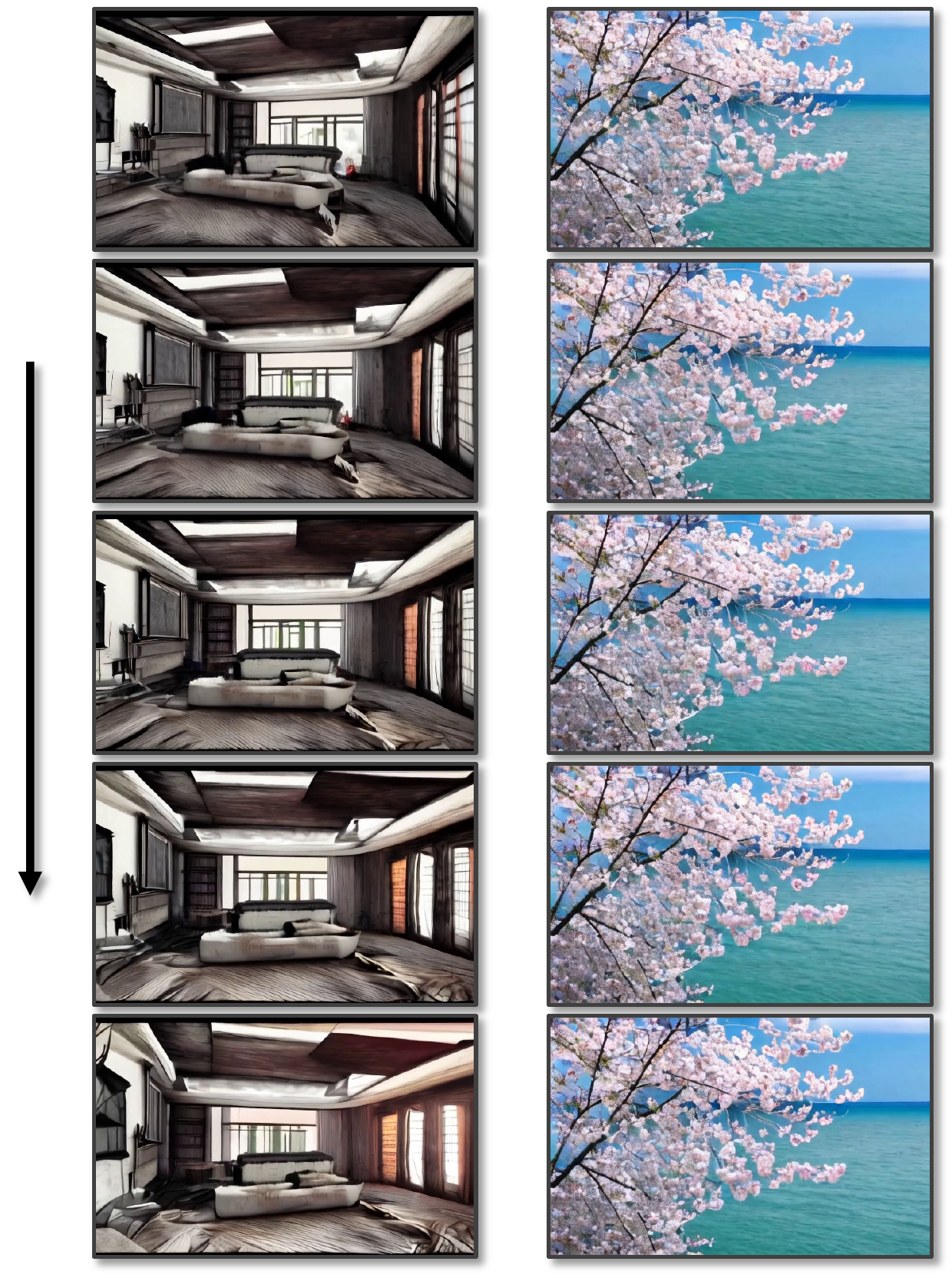}
    \caption{\small Generated videos at resolution $1280 \times 2048$ using our Stable Diffusion 2.0-based model and including our video fine-tuned text-to-video latent upsampler. Captions from left to right are: ``horror house living room interior overview design, Moebius, Greg Rutkowski, Zabrocki, Karlkka, Jayison Devadas, Phuoc Quan, trending on Artstation, 8K, ultra wide angle, pincushion lens effect.'' and ``Cherry blossom swing in front of ocean view, 4k, high resolution''.
    Frames are shown at 2 fps.} 
    \vspace{-0.5em}
    \label{fig:sd20_samples_supp_2}
\end{figure*}

\begin{figure*}[t!]
    \centering
    \includegraphics[width=0.9\textwidth]{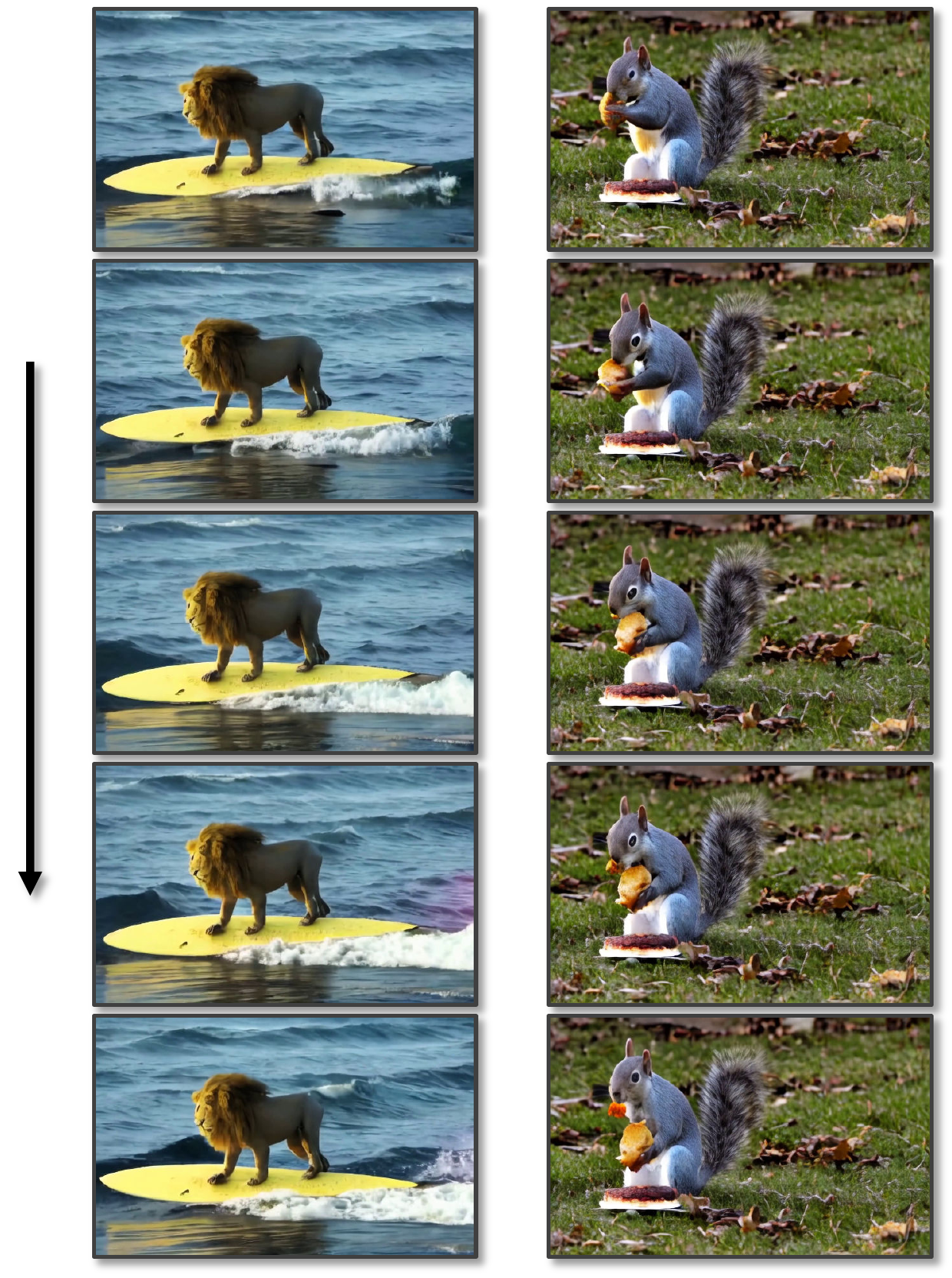}
    \caption{\small Generated videos at resolution $1280 \times 2048$ using our Stable Diffusion 2.0-based model and including our video fine-tuned text-to-video latent upsampler. Captions from left to right are: ``A lion standing on a surfboard in the ocean in sunset, 4k, high resolution'' and ``A squirrel eating a burger''.
    Frames are shown at 2 fps.} 
    \vspace{-0.5em}
    \label{fig:sd20_samples_supp_3}
\end{figure*}

\begin{figure*}[t!]
    \centering
    \includegraphics[width=0.9\textwidth]{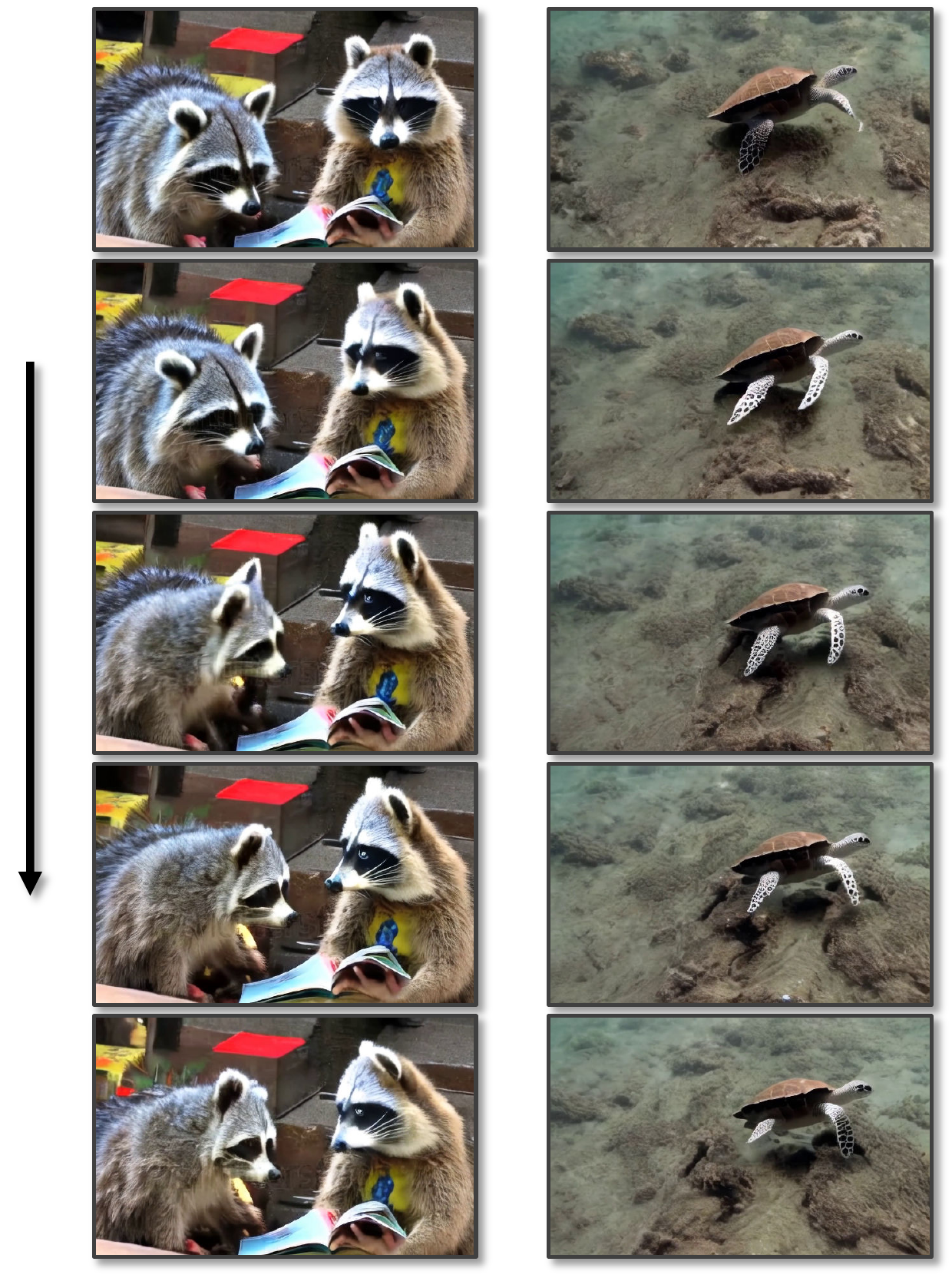}
    \caption{\small Generated videos at resolution $1280 \times 2048$ using our Stable Diffusion 2.0-based model and including our video fine-tuned text-to-video latent upsampler. Captions from left to right are: ``Two raccoons reading books in NYC Times Square'' and ``Turtle swims in ocean''.
    Frames are shown at 2 fps.} 
    \vspace{-0.5em}
    \label{fig:sd20_samples_supp_4}
\end{figure*}

In this section, we provide additional generated samples from our text-to-video LDMs. We show generated videos at resolution $320 \times 512$ (SD 1.4-based models) and at resolution $1280\times2048$ resolution (SD 2.0-based models with additional video fine-tuned $4\times$ upscaler). Moreover, we present generated videos that are extended ``convolutional in space'' and/or ``convolutional in time''; see~\Cref{app:convolutional_ldm}.
We are able to generate long, high resolution and high frame rate, expressive and artistic videos.
\begin{itemize}
    \item \textit{Regular video samples from SD 1.4-based Video LDM}: \Cref{fig:common}.
    \item \textit{``Convolutional in space'' (SD 1.4-based Video LDM)}: \Cref{fig:space_conv1,fig:space_conv2}.
    \item \textit{``Convolutional in time'' (SD 1.4-based Video LDM)}: \Cref{fig:time_conv1}.
    \item \textit{``Convolutional in space'' and ``convolutional in time'' (SD 1.4-based Video LDM)}: \Cref{fig:hotdog,fig:longteddy}.
    \item \textit{Regular video samples at $1280\times2048$ resolution from SD 2.0-based Video LDM with $4\times$ upscaler}: \Cref{fig:sd20_samples_supp_1,fig:sd20_samples_supp_2,fig:sd20_samples_supp_3,fig:sd20_samples_supp_4}.
    \item \textit{Regular video samples at $1280\times2048$ resolution from both SD 2.0-based and SD 2.1-based Video LDMs with $4\times$ upscaler}: All samples shown on our project page \url{https://research.nvidia.com/labs/toronto-ai/VideoLDM/}.
\end{itemize}

\subsubsection{Personalized Text-to-Video with DreamBooth}\label{app:more_dreambooth}
\begin{figure*}[t!]
    \centering
    \includegraphics[width=1.02\textwidth]{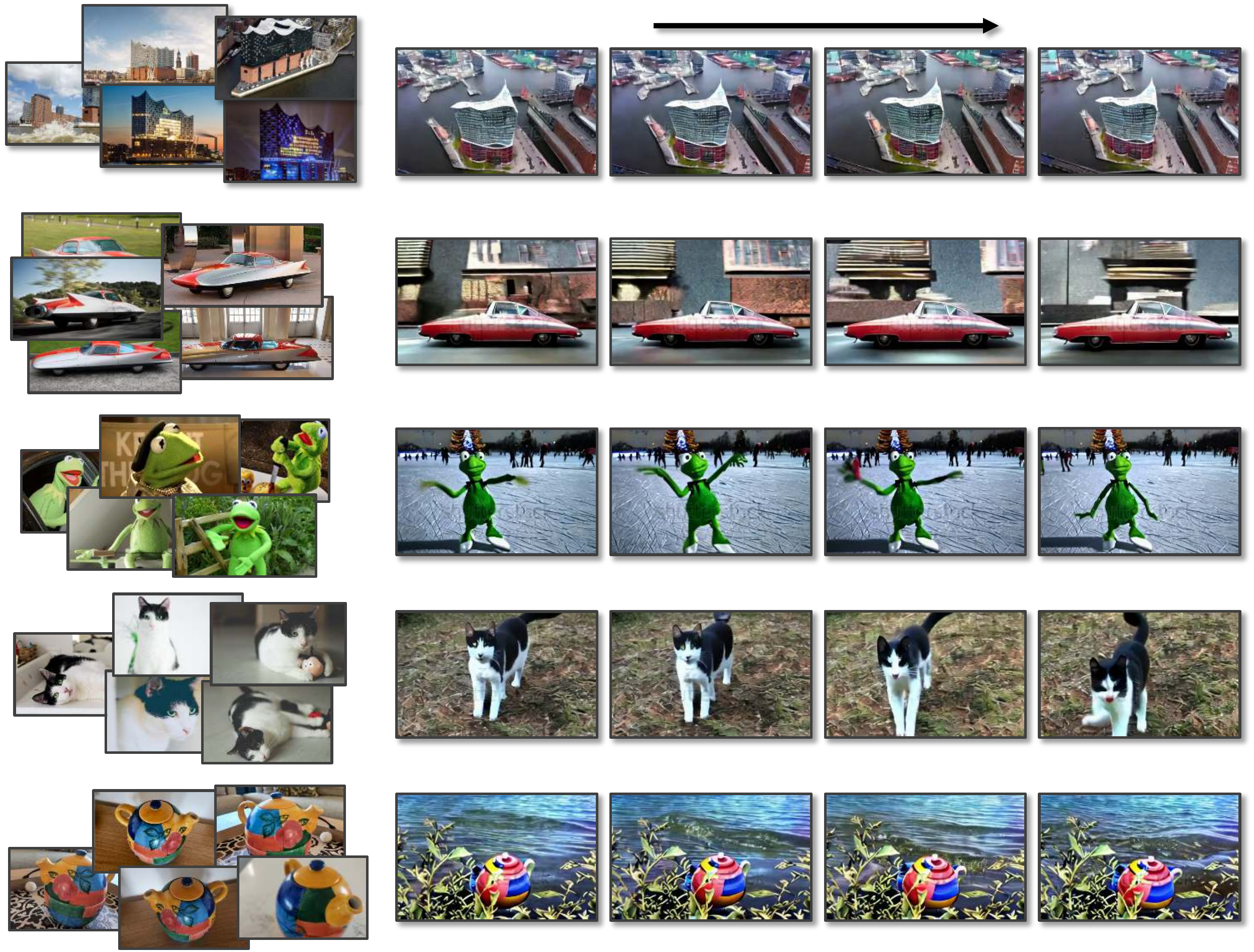}
    \caption{\small Generated DreamBooth-personalized videos at resolution $320 \times 512$. Text prompts of the videos can be found in~\Cref{tab:dreambooth}. Frames are shown at 1 fps.} 
    \vspace{-0.5em}
    \label{fig:five_dreambooth}
\end{figure*}
We provide additional generated personalized text-to-video samples. The generated videos can be found in~\Cref{fig:five_dreambooth}. The number of training images and the prompts used for the spatial and temporal layers can be found in~\Cref{tab:dreambooth}. We see that we are able to successfully generate videos that faithfully include the learnt objects and capture their identity well.
\begin{table}
    \centering
    \caption{Details for personalized text-to-Video with DreamBooth. Text in parentheses is given to the spatial layers but omitted for the temporal layers (as they are not part of the DreamBooth training). Generated samples can be found in~\Cref{fig:five_dreambooth}.}
    \label{tab:dreambooth}
    \begin{tabular}{l c l l}
        \toprule 
        \textbf{Model} & \# of training images & Prompts \\
        \midrule
        Building & 13 & ``(\emph{sks} building), 4k drone flight, high definition'' \\
        Car & 10 & ``A (\emph{sks}) car driving in Manhattan'' \\
        Frog & 23 & ``A (\emph{sks}) frog ice skating in Central Park on Christmas Eve'' \\
        Cat & 11 & ``A (\emph{sks}) cat walking, front view, high definition'' \\
        Tea pot & 8 & ``A (\emph{sks}) tea pot in the ocean'' \\
        \bottomrule
    \end{tabular}
\end{table}
\subsection{Mountain Biking Video Synthesis} \label{app:mountain_bike}
We conducted additional experiments on the Mountain Biking dataset~\cite{brooks2022generating} (see~\Cref{app:mountain_bike_dataset}) downsampled and center-cropped to resolution $256 \times 128$. We initially train our model for sparse key frame prediction at 1.875 fps. The temporal interpolation model is trained using 30 fps video data. We train
the temporal interpolation model to first scale from 1.875 fps to 7.5 fps, and then to scale from 7.5 fps to 30 fps. We are using one interpolation model with shared parameters for that, providing a conditioning label to indicate to the model which of the two temporal upsampling operations is desired.

We then compare our model with the publicly available model from Long Video GAN (LVG)~\cite{brooks2022generating}. We report FID \& FVD metrics as well as a human evaluation study in~\Cref{tab:biking}. We outperform LVG both in FID and human evaluation, but slightly underperform on FVD.

The first-person mountain biking videos have very rapidly changing background details (trees, branches, etc.). LVG cannot create these single-frame realistic details, ``smoothening out'' the background and therefore resulting in worse FID and also performing worse in the human evaluation study. Our method, on the other hand, has more realistic single frames; however, it slightly struggles to keep the temporal consistency of these details. The FVD metric favors short-term ``smoothness'' over photorealism, which explains the underperformance of our method in this metric. Generally, FVD is a metric with downsides and should be taken with a grain of salt, as discussed in detail in the Long Video GAN paper itself~\cite{brooks2022generating} (their Section 5.3), Overall quality and realism is best judged by human evaluators, where we outperform LVG.

\begin{figure*}[t!]
    \centering
    \includegraphics[width=0.8\textwidth]{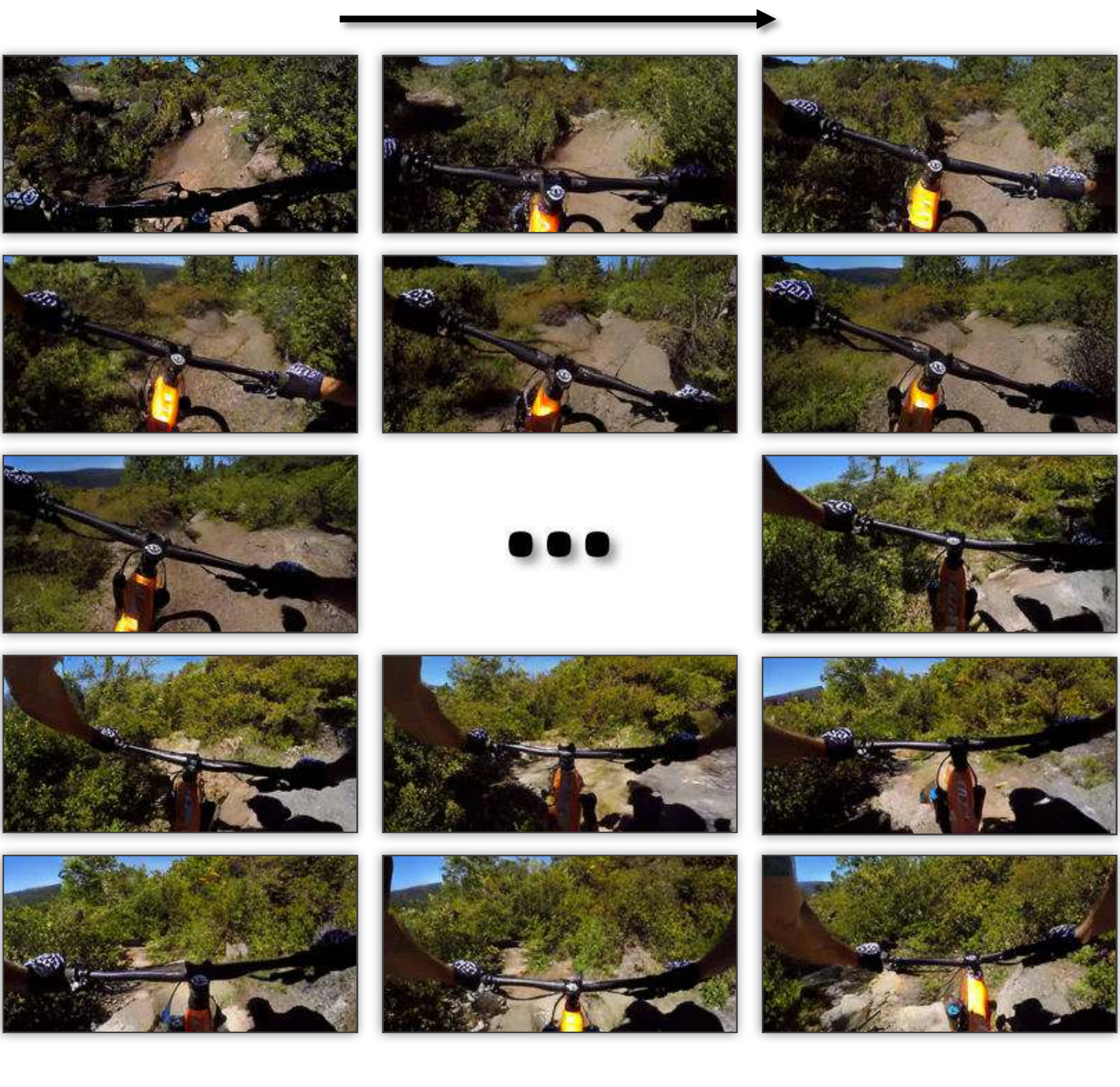}
    \caption{\small Generated 10 second (30 fps) Mountain Biking video at resolution $128 \times 256$. Frames are shown at 6 fps.} \vspace{-0.5em}
    \label{fig:bike1}
\end{figure*}

\begin{figure*}[t!]
    \centering
    \includegraphics[width=0.8\textwidth]{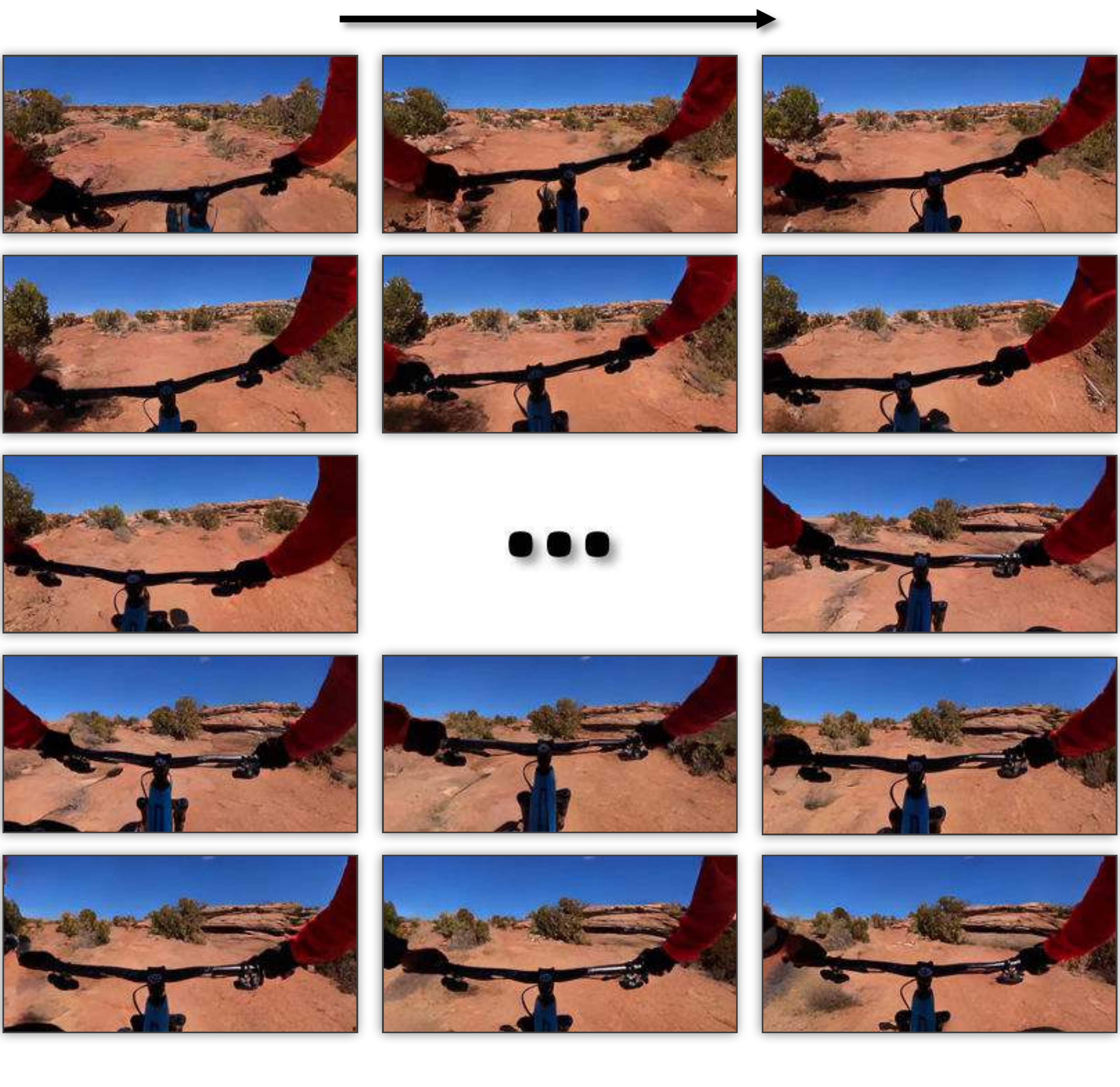}
    \caption{\small Generated 10 second (30 fps) Mountain Biking video at resolution $128 \times 256$. Frames are shown at 6 fps.} \vspace{-0.5em}
    \label{fig:bike2}
\end{figure*}

\begin{figure*}[t!]
    \centering
    \includegraphics[width=0.8\textwidth]{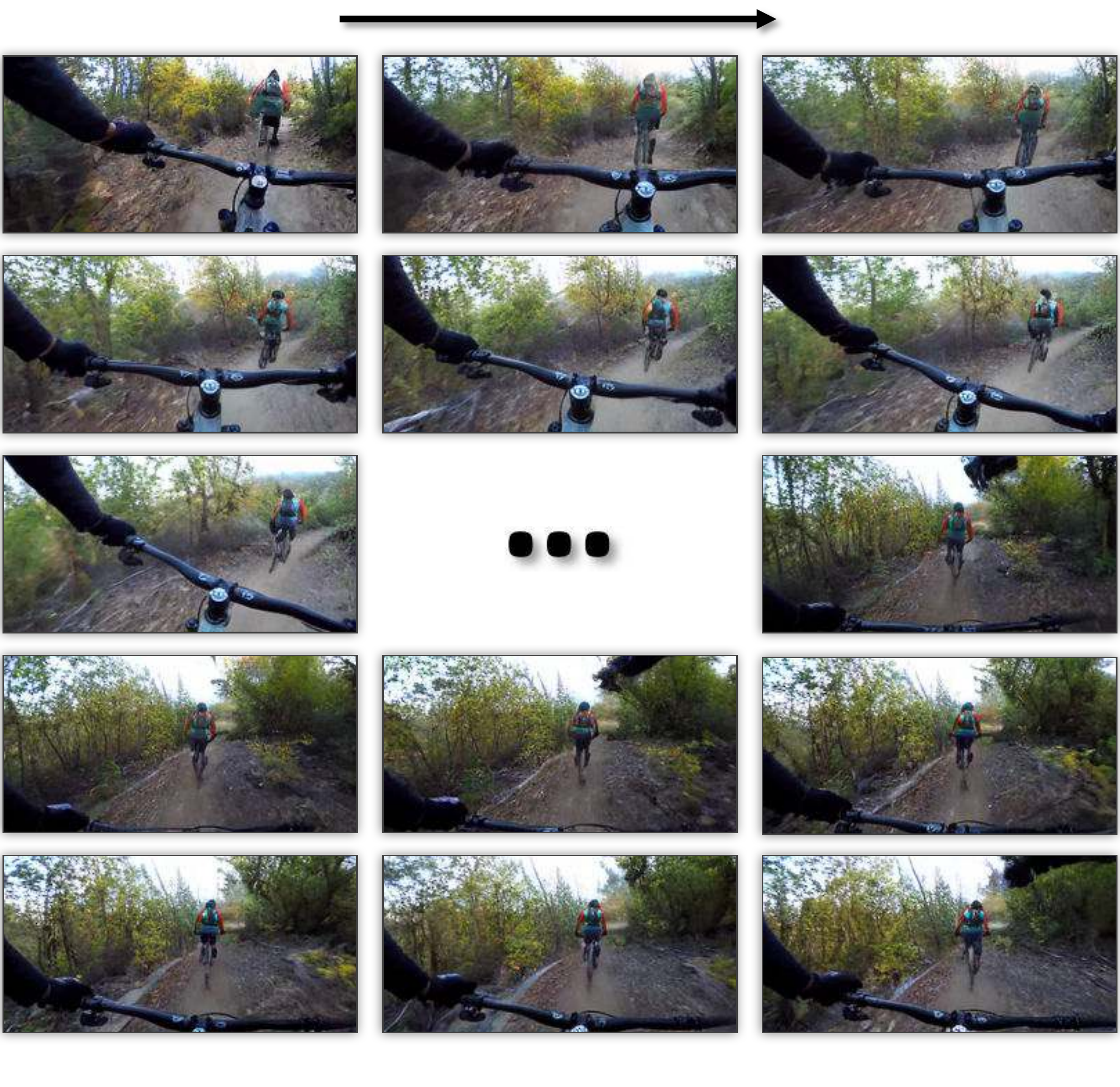}
    \caption{\small Generated 10 second (30 fps) Mountain Biking video at resolution $128 \times 256$. Frames are shown at 6 fps.} \vspace{-0.5em}
    \label{fig:bike3}
\end{figure*}

We show generated 10 second (30 fps) mountain biking videos in~\Cref{fig:bike1,fig:bike2,fig:bike3}.
\begin{table}[]
    \centering
    \caption{Comparison with Long Video GAN (LVG) on Mountain Biking videos (human evaluation on the right).}
    \label{tab:biking}
    \begin{tabular}{l c c}
        \toprule
        \textbf{Method} & FVD & FID \\
        \midrule
        LVG \cite{brooks2022generating} & 85.3 & 21.1 \\
        Video LDM \textit{(ours)} & 118 & 7.73 \\
        \bottomrule
    \end{tabular}
    \quad
    \begin{tabular}{l c c c}
        \toprule
        \textbf{Method} & Pref. A & Pref. B & Equal \\
        \midrule
        Video LDM \textit{(ours)} vs. LVG \cite{brooks2022generating} & 54.2 & 42.2 & 3.6 \\
        \bottomrule
    \end{tabular}
\end{table}
\subsubsection{Video-Finetuning of our Decoders}
We also perform a small ablation experiment over the video fine-tuning of the decoders (as described in~\Cref{sec:dec_finetuna}) for the mountain biking Video LDM. As can be seen in~\Cref{tab:vid-ft-supp}, video fine-tuning the decoder allows for a significant performance boost on mountain biking.
\subsection{Driving Video Synthesis} \label{app:extended_driving}
\begin{figure*}[t!]
    \centering
    \includegraphics[width=1.0\textwidth]{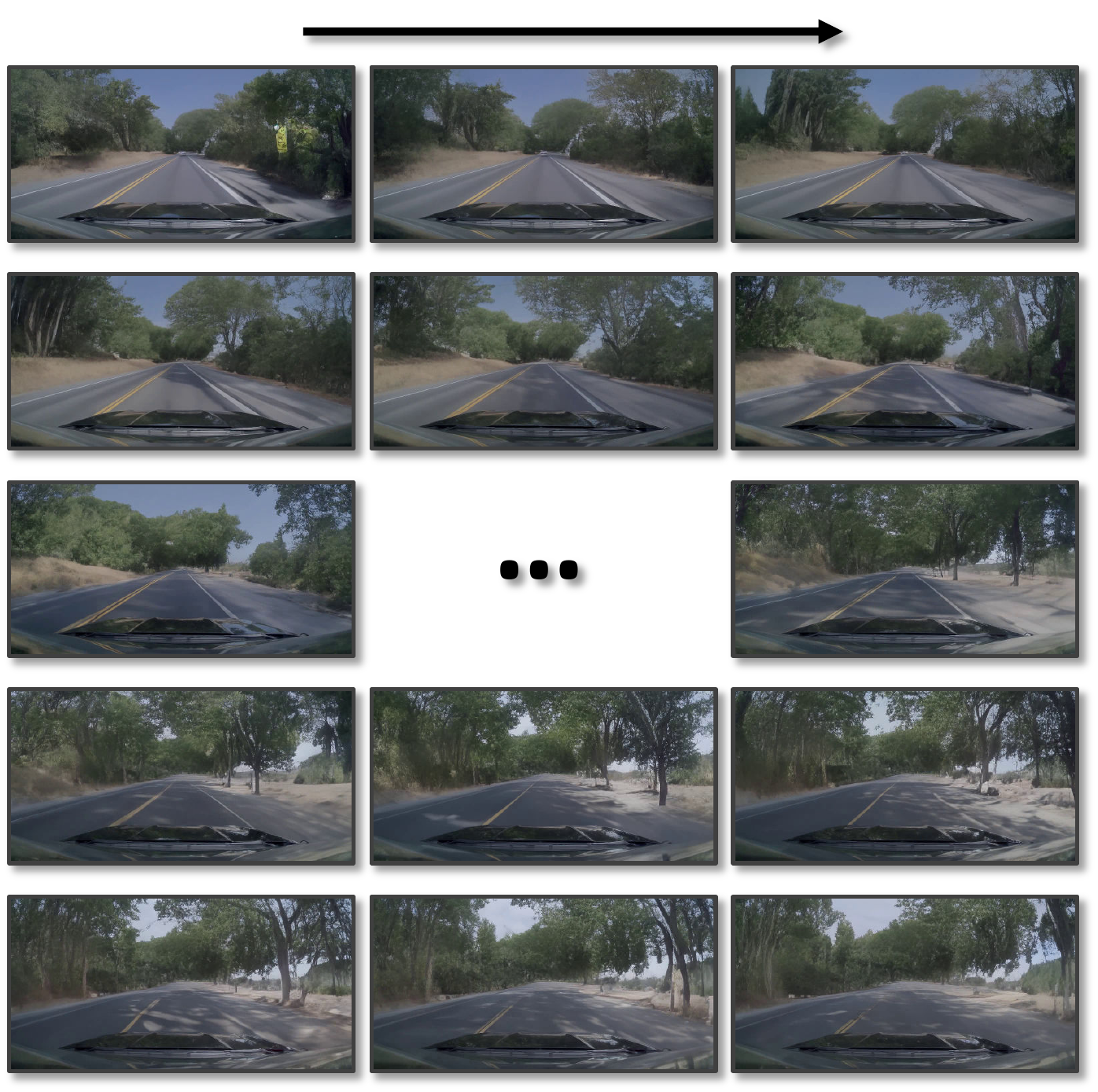}
    \caption{\small Generated Driving video upsampled to resolution $512 \times 1024$. Frames are shown at 2 fps.} \vspace{-0.5em}
    \label{fig:drive1}
\end{figure*}

\begin{figure*}[t!]
    \centering
    \includegraphics[width=1.0\textwidth]{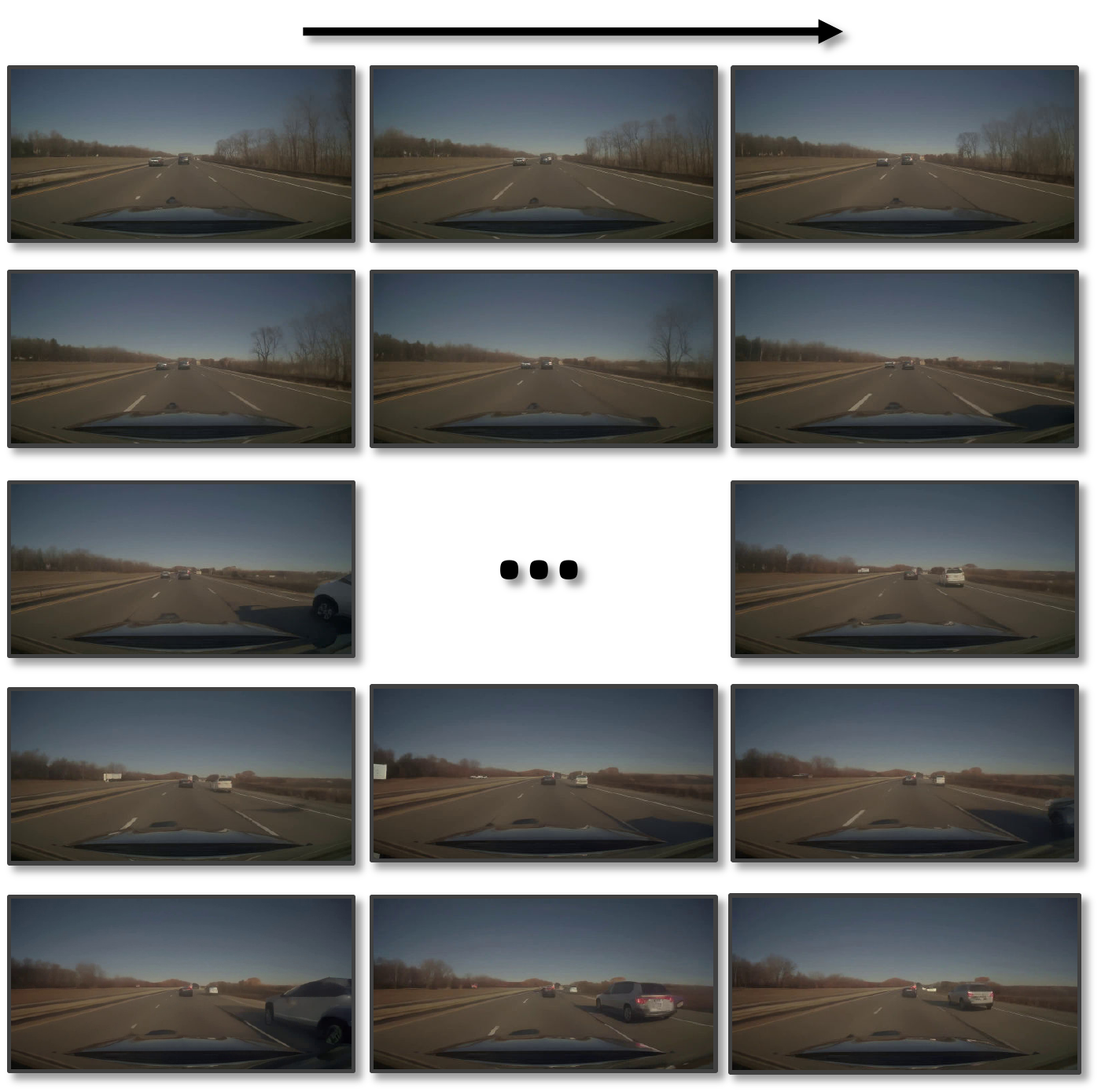}
    \caption{\small Generated Driving video upsampled to resolution $512 \times 1024$. Frames are shown at 2 fps.} \vspace{-0.5em}
    \label{fig:drive2}
\end{figure*}

\begin{figure*}[t!]
    \centering
    \includegraphics[width=1.0\textwidth]{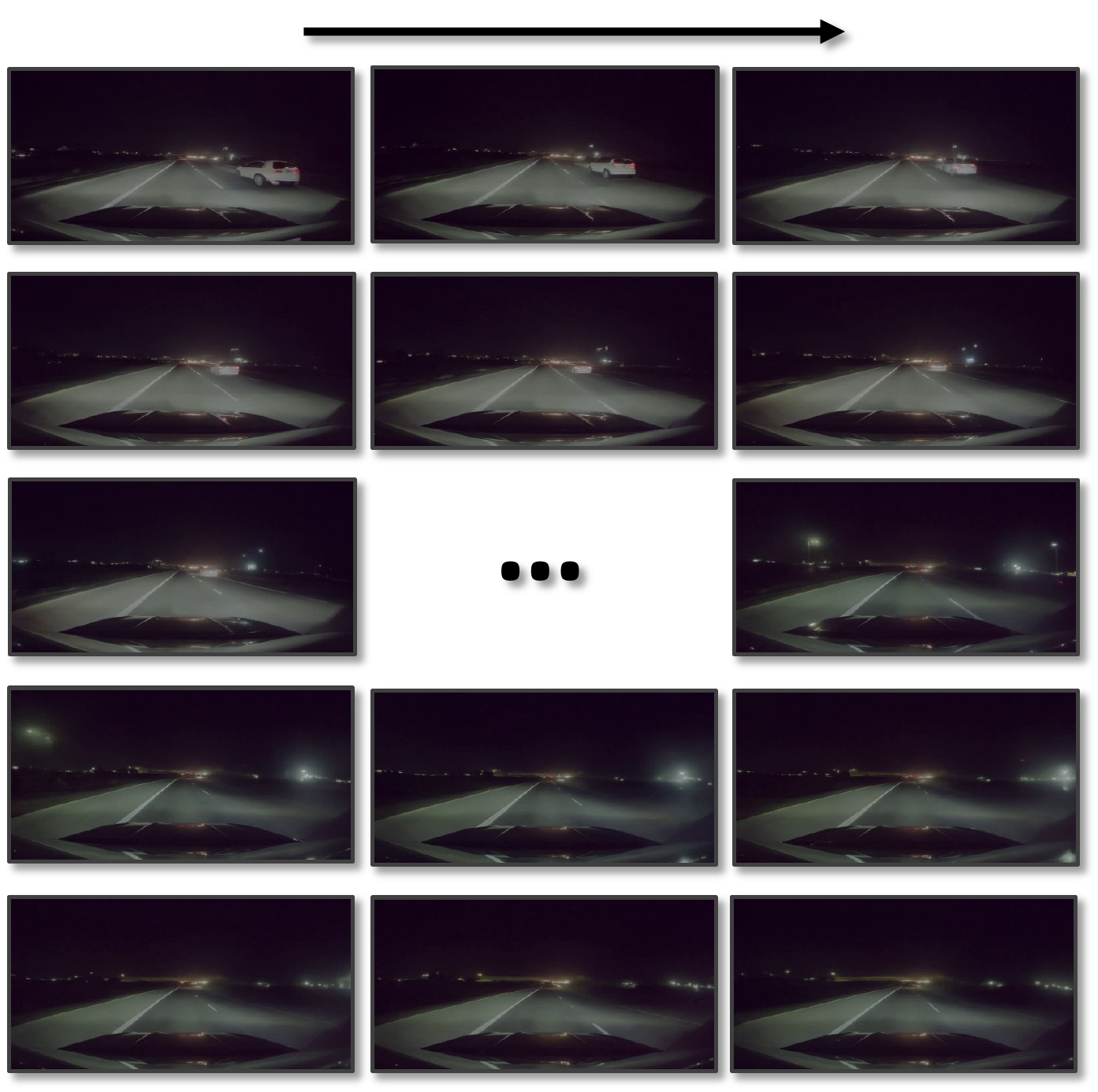}
    \caption{\small Generated Driving video upsampled to resolution $512 \times 1024$. Frames are shown at 2 fps.} \vspace{-0.5em}
    \label{fig:drive3}
\end{figure*}

In this section, we provide additional generated samples from our Video LDM trained on real-world driving data. The samples are upsampled to resolution $512 \times 1024$ using our temporally aligned video upsampler; see~\Cref{fig:drive1,fig:drive2,fig:drive3}.

\subsubsection{Ablation on Additional Image Discriminator for Decoder Fine-Tuning}
To fine-tune our decoder (see \Cref{sec:dec_finetuna} and \Cref{sec:driving_exp}), we tested using not only a 3D-convolutional video discriminator, but to also use an additional image discriminator to maintain and possibly enhance image-level quality. Using our main driving model Video LDM, we evaluated reconstruction FVD and FID scores after decoder fine-tuning using only the video discriminator vs. with an additional image discriminator. The results are shown in Tab.~\ref{tab:image_discriminator}. We found that image-level quality, as measured by FID, barely changed, while video quality, as measured by FVD, suffered considerably when an additional image discriminator was used. Consequently, we resorted to using only the video discriminator.

\begin{table}[t]
    \centering
        \caption{\small Decoder fine-tuning with and without additional image discriminator. We are showing reconstruction FVD and FID scores after decoder fine-tuning using our main Video LDM model for driving scenario video generation.\vspace{-0.5em}}
    \label{tab:image_discriminator}
    
    \resizebox{.6\linewidth}{!}{%
    \begin{tabular}{l c c}
        \toprule
        \textbf{Method} & Reconstruction FVD & Reconstruction FID \\
        \midrule
        Video discriminator only & 32.94 & 9.17 \\
        Additional image discriminator & 51.01 & 9.04 \\
        \bottomrule
    \end{tabular}
    }
\end{table}

\subsubsection{Ablation on Image-level Quality Degradation after Temporal Video Fine-Tuning}
Does the image-level quality of the generated outputs of the LDM degrade when the model is fine-tuned for video synthesis? To test this, we measured image-level FID scores using independent frames generated by the image backbone model (setting $\alpha_\phi^i{=}1$) and compared to FID scores based on frames generated by the full Video LDM after learning the $\alpha_\phi^i$ parameters and the temporal alignment layers on videos. For this experiment, we used the smaller version of the Video LDM for driving video generation that was used in our ablation experiments (\Cref{sec:ablations}). With $\alpha_\phi{=}1$, we obtain 47.00 FID; with the learnt parameters, we get 48.26 FID. We observe only a tiny degradation and conclude that image-level quality is affected only slightly when training the temporal layers for video generation.





\end{document}